\documentclass[10pt, a4paper]{article}
\usepackage[utf8]{inputenc} 
\usepackage[T1]{fontenc} 
\usepackage{amsmath, amssymb} 
\usepackage[margin=1in]{geometry} 
\usepackage{graphicx} 
\usepackage{hyperref} 
\usepackage{xurl} 
\usepackage{booktabs} 
\usepackage{listings}
\usepackage{ifluatex}
\ifluatex
  \usepackage{fontspec}
  \usepackage{unicode-math}
  \usepackage{xparse}
  \NewDocumentCommand{\citeay}{m}{%
    \citeauthor{#1} (\citeyear{#1})\cite{#1}%
  }
\else
  \usepackage{mathptmx} 
  \newcommand{\citeay}[1]{%
    \citeauthor{#1} (\citeyear{#1})\cite{#1}
  }
\fi

\usepackage{newunicodechar}
\newunicodechar{，}{,}    
\newunicodechar{—}{---}  
\newunicodechar{–}{--}   
\newunicodechar{’}{'}
\newunicodechar{“}{"}
\newunicodechar{”}{"}
\newunicodechar{→}{\ensuremath{\rightarrow}}
\newunicodechar{μ}{\ensuremath{\mu}}
\newunicodechar{⁺}{\ensuremath{^+}}
\newunicodechar{⁻}{\ensuremath{^-}}
\newunicodechar{−}{-}    

\usepackage{tcolorbox}
\tcbuselibrary{skins} 

\usepackage{longtable}

\usepackage{diagbox}
\usepackage{subcaption}
\usepackage{float}
\usepackage{placeins}
\usepackage{array}
\usepackage{listings}
\usepackage{seqsplit}
\usepackage{xcolor}
\usepackage{amsthm} 
\usepackage{tabularx} 
\usepackage[backend=bibtex, style=numeric]{biblatex}
\DeclareFieldFormat{title}{#1}
\title{BigBang-Proton Technical Report \\
    {\large Next-Word-Prediction Is Scientific Multitask Learner}
}
\author{SuperSymmetry Technologies}
\date{}

\newcounter{hypothesis}
\newtcolorbox[use counter=hypothesis]{hypothesis}{
    enhanced,
    colback=blue!3!white,
    colframe=blue!75!black,
    title=Hypothesis~\thetcbcounter,
    fonttitle=\bfseries,
    arc=4mm,
    boxrule=1pt,
    drop shadow,
    attach boxed title to top left={yshift=-2mm,xshift=4mm},
    boxed title style={sharp corners, colback=blue!75!black, fonttitle=\color{white}\bfseries}
}
\usepackage{hyperref}

\begin{document}
	
\maketitle
\footnotetext[1]{Correspondence to: Hengkui Wu, hkwu@ssymmetry.com}

\begin{abstract} 
We introduce BigBang-Proton, a unified sequence-based architecture for auto-regressive language modeling pretrained on cross-scale, cross-structure, cross-discipline real-world scientific tasks to construct a scientific multi-task learner. BigBang-Proton incorporates three fundamental innovations compared to mainstream general-purpose LLMs: Theory-Experiment Learning paradigm aligns large-scale numerical experimental data with theoretical text corpora; Binary Patch Encoding replaces byte pair encoding(BPE) tokenization; Monte Carlo Attention substitutes traditional transformer architectures. Through next-word-prediction pretraining on cross-discipline scientific datasets of real-world problems mixed with general textual corpus, followed by fine-tuning and inference on downstream tasks, BigBang-Proton demonstrates 100\% accuracy in up to 50-digit arithmetic addition operations, performance on par with leading specialized models in particle physics jet tagging, matching MAE of specialized models in inter-atomic potential simulation, performance comparable to traditional spatiotemporal models in water quality prediction, and benchmark-exceeding performance in genome modeling. These results prove that language-guided scientific computing can match or exceed the performance of task-specific scientific models while maintaining multitask learning capabilities. We further hypothesize to scale the pretraining to the universe scale as a fundamental step toward developing material world foundational model. 

\begin{figure}[ht]
    \centering
    \includegraphics[width=1\textwidth]{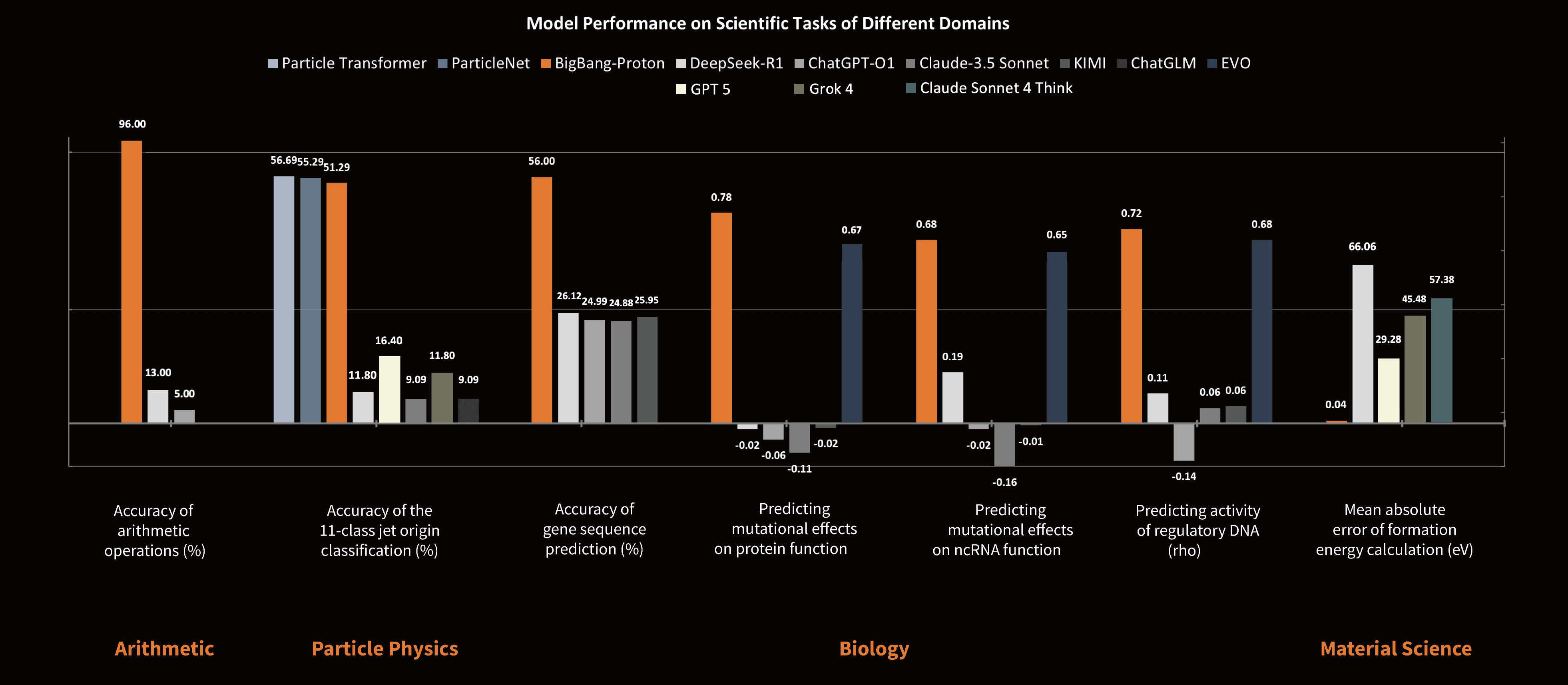} 
    \caption{Comparison of BigBang-Proton performance against specialized models and general-purpose LLMs across multiple scientific domains. BigBang-Proton achieves 96\% accuracy in up to 50-digit arithmetic, outperforming DeepSeek R1 (13\%) and ChatGPT o1 (5\%). In particle physics jet tagging, it attains 51.29\% accuracy in 11-class classification, comparable to specialized models ParticleNet (55.29\%) and Particle Transformer (56.69\%), while GPT 5 (16.4\%) and Grok 4 (11.8\%) show near-random performance. For genome modeling, BigBang-Proton demonstrates higher correlations than biology foundation model Evo, while general LLMs yield negative correlations. In materials science inter-atomic potential simulation, it achieves MAE of 0.043 eV in formation energy predictions, outperforming GPT 5 (29.28 eV), Grok 4 (45.48 eV), Claude 4 (57.38 eV), and DeepSeek R1 (66.06 eV).}
    \label{fig:task_performance_comparison}
\end{figure}

\end{abstract}
	
\section{Introduction}
	
Large language models (LLMs) have demonstrated remarkable general-purpose capabilities across a broad spectrum of tasks traditionally performed by humans, achieved solely through next-token prediction pretraining on sequence-based architectures\cite{qwen2025qwen25technicalreport}\cite{gpt2}\cite{brown2020language}\cite{gpt4}\cite{grattafiori2024llama3}\cite{deepseekv3}\cite{deepseekr1}. These models exhibit exceptional proficiency in domains such as text generation, programming, and standardized examinations, particularly in scenarios that align with prompt-response frameworks, as evidenced by performance on established language-based benchmarks, e.g., LMSYS Chatbot Arena\cite{lmsys}, MMLU\cite{mmlu}, GPQA\cite{gpqa},Humanity's last exam\cite{hle}.
    
However, LLMs have thus far exhibited limited success in addressing real-world scientific challenges or generating novel scientific insights, as these tasks typically demand solutions that extend beyond conventional prompt-answer paradigms. Prevailing evaluation benchmarks designed to simulate human cognitive processes, are predominantly constrained to multiple-choice or text-completion formats, which restricts their applications in assessing LLMs' potential for scientific applications. Existing applications of LLMs in scientific discovery can be categorized into the following types. First, LLMs are employed to search, summarize, and organize information and knowledge from scientific papers and documents, serving as an enhanced alternative to traditional search engine. The use of reasoning model post-trained with reinforcement learning, such as ChatGP o1\cite{chatgpto1} and DeepSeek-R1\cite{deepseekr1} to address Ph.D.-level questions with long-horizon reasoning steps also falls into this category. Second, general-purpose LLMs are fine-tuned with scientific data for domain-specific applications, as exemplified by Darwin\cite{xie2023darwin, xie2025darwin15largelanguage}, and Matterchat\cite{tang2025matterchatmultimodalllmmaterial}. Third, LLMs are trained either upon established models or from scratch using scientific data, as demonstrated by NatureLM\cite{xia2025naturelanguagemodeldeciphering} and Galatica\cite{Taylor2022GalacticaAL}. Four, general-purpose LLMs are utilized as tools for generating code or other intermediate outputs rather than definitive solutions, as seen in Funsearch\cite{romera2024mathematical} and Deepmind's AI Co-scientist\cite{coscientist}. Notably, general-purpose LLMs have been applied in symbol- and language-intensive domains, particularly in mathematical theorem proving, which benefits from formalized datasets, including LeanDojo\cite{leandojo} and other LLMs of theorem prover\cite{theoremllama}\cite{goedelprover}. Additionally, LLMs have been applied in computer science research, which is grounded in theoretical insights and programming experiments, such as algorithms developed by Sakana AI\cite{sakana}. Although LLMs have shown promises in theory-driven tasks within scientific research, progress in solving experimental challenges, particularly in the realm of natural sciences, remains limited.

A critical category of experimental science involves numerical analysis in data-driven computational problems. Modern scientific research is founded upon the interplay between theories and experimental observations, the latter of which frequently yield quantitative datasets across diverse disciplines. Despite their strength in language reasoning, mainstream LLMs exhibit negligible competence in numerical problem-solving, a limitation rooted in the inefficacy of byte-pair encoding (BPE) tokenization for numerical data representation. For instance, in arithmetic operations, the accuracy of LLMs degrades precipitously to below 50\% for problems exceeding five digits\cite{yuan2023largelanguagemodelsperform}. Circumventing direct numerical computation by calling external code interpreters or task-specific algorithms as seen in agentic operations further proves that LLMs lack intrinsic capabilities in numerical analysis in experimental science. 

Numerical computation capability is indispensable in vast scientific domains, including time series analysis, tabular data processing, and spatio-temporal modeling. Recent research has increasingly focused on developing world models to address the limitations of language-centric pretraining, exploring multimodal integration, such as Feifei Li's spatial intelligence\cite{yang2025thinkingspacemultimodallarge}, or specialized architectures for physical attribute learning, such as JEPA\cite{jepa} and NVIDIA's Cosmos\cite{cosmos}. However, these efforts often overlook the fundamental principle that the intrinsic properties of the physical world including space and time are quantitatively defined. Future advancements in model architectures and pretraining methodologies will necessitate breakthroughs in numerical reasoning to enable LLMs to achieve real understanding of the world. 

Over the past several decades, the scientific community has developed a multitude of specialized computational models to address task-specific challenges. Notable examples include AlphaFold\cite{AlphaFold}\cite{Abramson2024} for protein structure prediction in computational biology, inter-atomic potential models\cite{dpa2}\cite{omat24}for molecular dynamics in materials science, jet-tagging algorithms in high-energy particle physics\cite{Qu:2019gqs}\cite{Qu:2022mxj}, plasma-control systems in nuclear fusion research\cite{plasmadm}\cite{plasmatearing}, computational fluid dynamics (CFD) simulations\cite{cfdsurvey}, numerical weather prediction (NWP) models\cite{graphcast}\cite{bi2023accurate}, and electronic design automation (EDA) tools\cite{huang2021machinelearningelectronicdesign} for semiconductor development. These models employ highly specialized architectures tailored to their respective domains, incorporating unique data structures and modalities to optimize performance.

Scientific researches have been organized along disciplinary lines historically, a paradigm shaped by the cognitive and logistical constraints of human researchers. However, this compartmentalization stands in stark contrast to the interconnected nature of physical reality, where phenomena span multiple scales and structures. The artificial segregation of disciplines has led to incomplete and fragmented models of the material world, hindering the discovery of cross-disciplines principles. Recent efforts on unifying the learning of multi-scale domains have mostly been put on material science and biomedicine\cite{xia2025naturelanguagemodeldeciphering} \cite{dpa2}as both material crystals and biological molecules can be described through molecular dynamics, which are derived from first-principles calculations of many-body Schrödinger equations. Multi-scale modeling\cite{vlachas2022multiscale,lyu2023multifidelity,mi2023cascade,fang2023highfrequency,zhang2024sinenet,serrano2023operator} and multi-physics transfer learning\cite{lozano2023machine,yang2023prompting,yang2024pde,liu2023prose,sun2024foundation,song2024fminit,hang2024unisolver,subramanian2024foundation,mccabe2023multiple,chen2024building,hao2024dpot} have also been explored in computational fluid dynamics and geological modeling of soil and earth system\cite{inbook}, solving PDEs by numerical analysis techniques such as finite element method\cite{szabo2021finite}, finite volume method, finite difference method, discrete element method.  for the purpose of building domain-specific foundational models. However, these models tend to be confined to their respective field and usually fail to generalize to other research areas. Consequently, there is a pressing need for a foundational model capable of generalizing across all scientific disciplines. 

To compensate for the inherent lack of domain-specific inductive biases in generic neural network architectures, researchers have introduced physics-informed neural networks (PINNs)\cite{raissi2017physics1,raissi2017physics2,bai2020applying,chen2021physics,kim2021knowledge,jin2021nsfnets,gao2021phygeonet,mowlavi2021optimal,cheng2021deep,zeng2022competitive,gnanasambandam2022self,huang2022meta,raissi2019physics}, which integrate physics principles from thermodynamics, quantum mechanics to PDEs,ODEs,SDEs and symmetry and intuitive physics \cite{yu2022gradient,liu2022multi,sirignano2018dgm,zang2020weak,wang2021eigenvector,kharazmi2021hp,jagtap2020extended,wang2021physics} describing certain physics law as physics prior or constraints into model architecture,regularization term of loss function, or optimizer design, inference methods\cite{hao2022physics,karniadakis2021physics,cuomo2022scientific,cai2021physics,dissanayake1994neural,moseley2021finite,jagtap2020xpinns,wang2022causal,lu2021learning,li2020fourier,xue2020amortized,cai2021deepmmnet}. While these approaches have demonstrate better convergence in narrowly defined applications, they suffer from a critical limitation: their architectural specialization inherently isolates learning processes, restricting their applicability and impeding cross-disciplinary knowledge transfer. Unlike LLMs, which achieve multitask generalization and emergent reasoning capabilities through pretraining on vast, diverse corpora\cite{achiam2023gpt,wei2022emergentabilitieslargelanguage}, scientific models remain confined to specific domains, lacking a unified mechanism for scalable transferable learning. 

A task-agnostic generalist architecture\cite{gpt2} for scientific discovery is essential for several reasons. Current task-specific models are limited by their narrow scope, and a unified framework could transcend disciplinary boundaries, enabling a single model to tackle diverse scientific tasks, from molecular dynamics to climate simulations, without requiring architectural redesign. A generalist model would also facilitate cross scale, cross structure, and cross disciplinary knowledge transfer\cite{pan2009survey, han2021pretrainedmodelspastpresent, BengioCourvilleVincent2013} by identifying latent patterns across disparate domains, thus accelerating discoveries that rely on analogies between seemingly unrelated systems, such as turbulence in fluids and galactic structure formation. Additionally, scientific knowledge exists in both linguistic, such as theories, hypotheses, and publications, and quantitative, such as experimental and observational datasets, forms, and a unified model must integrate symbolic reasoning with data driven learning, enabling direct inference over both modalities within a single pretrained framework. While scaling laws in language based pretraining have started to plateau\cite{hardywhite2024scalinglaws, altinuc2025gptscalingwall, ilyas2024pretraining, marcus2025breakingopenai}, their potential in science remains largely unexplored. Given the complexity of physical structures, increased model scale, coupled with multi-modal scientific pretraining, could lead to breakthroughs in predictive accuracy and generalization, similar to the emergent capabilities observed in large language models. The future of computational science lies in developing universal, pretraining-compatible architectures that integrates knowledge across disciplines while respecting domain specific constraints. By unifying language, numerical data, and physical principles, such models could overcome the limitations of human defined disciplinary boundaries, fostering a more holistic understanding of nature. 

In this study, we introduce BigBang-Proton, a generalist architecture designed for multitask learning across diverse scientific domains. As an advanced iteration of its predecessor, BigBang-Neutron\cite{wu2024scalingparticlecollisiondata}, BigBang-Proton retains the binary patch framework while incorporating significant architectural enhancements. The proposed system utilizes a sequence-based architecture\cite{sutskever2014sequence} that transforms multimodal inputs, including textual, numerical, and symbolic scientific data, into byte sequences via a Binary Patch Encoding representation. Multidisciplinary datasets, encompassing quark jet tagging from particle collision experiments, inter-atomic potential simulations from materials science, genome and protein sequence structure prediction, and lake water quality prediction from environmental engineering, are curated under a Theory-Experiment learning paradigm and concatenated without customized preprocessing into a unified sequence for next-word-prediction pretraining. These patched representations are processed by the Monte Carlo Attention layer, which computes attention matrices while exponentially expanding the effective context length relative to layer depth. During inference, traditional classification and regression tasks in scientific computing typically require specialized model architectures, such as graph neural networks (GNNs), commonly used in jet tagging\cite{Zhu2023ParticleNetAI, Qu:2022mxj} and inter-atomic potential modeling\cite{deng2023chgnetpretraineduniversalneural, dpa2}, which are generally incompatible with next-token-prediction models. BigBang-Neutron\cite{wu2024scalingparticlecollisiondata}, the predecessor of BigBang-Proton, employs a classical large language model (LLM) architecture, including transformer layers and MLP feedforward networks, with an adjusted LM-head layer to perform classification or regression tasks, thereby supporting language generation, classification, and regression tasks. In contrast, BigBang-Proton uniformly addresses classification and regression using next-patch-prediction, converting all scientific tasks into auto-regressive inference problems. 

BigBang-Proton operates as a unified multitask learner, enabling diverse scientific modeling tasks to be reformulated as sequence-to-x (where x represents language, classification, regression, material structures, DNA, sensor signals, etc.) learning problems within a single architectural framework. The model was pretrained using our Theory-Experiment learning paradigm, which integrates textual theoretical knowledge  with large-scale experimental numerical datasets. The core contributions of our work include: 
\begin{enumerate}
    \item This study establishes for the first time that an auto-regressive model can perform multitask learning across diverse domains in real-world scientific research. BigBang-Proton is designed as a task-agnostic architecture that enables cross-scale, cross-structure, and cross-discipline pretraining by integrating human language with fundamental physical elements including space, time, energy, and matter. The use of highly heterogeneous datasets from multiple scientific disciplines constructs capabilities beyond language reasoning and chain-of-thought (COT) inference. This work presents the first attempt to treat the material world as a unified entity in the development of a foundational model for scientific discovery, engineering, manufacturing, robotics, and space intelligence.
    \item Our results indicate that Binary Patch Encoding is markedly superior to byte-pair encoding (BPE) tokenization in handling large digits and large-scale numerical computations, demonstrating the ability to simulate arithmetic logic unit (ALU) operating principle. Given that the majority of experimental results in science and engineering are derived from observations and measurements stored as numerical data, an effective representation of numbers in the hidden space is critical for constructing a foundational model of the material world.
    \item We have demonstrated that through language-guided scientific computing, BigBang-Proton has achieved performance surpassing or comparable to the state-of-the-art specialized models in tasks including particle jet tagging, inter-atomic surface potential simulation, DNA and protein structure prediction, and lake water quality forecasting.
    \item We have validated that the Theory-Experiment learning framework, which synthesizes linguistic knowledge and experimental data, provides an effective approach to address the corpus discrepancies across scientific disciplines, domains, and the gap between theory and experimentation.
    \item Our analysis reveals that BigBang-Proton is capable of capturing the complete structure of particle jets, material crystals, and DNA sequences, and can generate corresponding pseudo-structures, thereby paving the way for the model to learn increasingly complex structures such as those in QCD, cellular systems, and Earth system. In contrast to the widely adopted long-horizon chain-of-thought approaches, structure-learning is essential for understanding the material world and represents an important pathway toward artificial general intelligence (AGI) or artificial superintelligence (ASI). Our results show that long-horizon chain-of-thought methods exhibit complete failure in handling real-world scientific tasks.
    \item We introduce Monte Carlo Attention as an alternative for the transformer architecture used in mainstream LLMs. Monte Carlo Attention is designed to address the computational complexity inherent in binary patch attention calculations, while retaining the advantages of both sparse attention and state space models, which are considered major alternatives to transformers. With Monte Carlo Attention, the context length of language models can grow exponentially with the number of attention layers. In this study, BigBang-Proton employs 20 layers of Monte Carlo Attention, achieving a context capacity of $10^{30}$bytes. To reach the estimated number of baryon particles in the observable universe $10^{80}$, the number of Monte Carlo Attention layers can be set to 60. Such high context lengths are essential for the model to effectively learn complex material structures, ranging from microscopic systems such as cells and QCD phenomena, to macroscopic structures such as Earth system, aircraft, cars, and the universe.
    \item Our work illustrates that pretraining has not yet reached its fundamental limits. The ultimate limit of pretraining is the boundary of the universe. This work naturally drives the pursuit of scaling model to the magnitude of the universe.
\end{enumerate}

The multitask learning performance of BigBang-Proton provides compelling evidence that specialized scientific tasks across diverse domains can be effectively unified through sequence-based auto-regressive end-to-end learning despite significant differences in their data characteristics and solution approaches. This finding establishes a critical foundation for the development of comprehensive material world foundational models. The demonstrated capabilities suggest that a single, properly designed architecture can simultaneously achieve both breadth of application across scientific disciplines and depth of performance competitive with specialized solutions.

\section{Approach}

\subsection{Theory-Experiment Learning for Language-Guided Scientific Computing}

The Theory-Experiment Learning paradigm addresses critical limitations in current LLMs on large-scale experimental data computing by creating a unified framework that aligns experimental data with theoretical knowledge through sequence-based learning. The core innovation lies in establishing a hybrid representation that directly aligns numerical experimental data with textual descriptions, borrowing the alignment principle from vision-language models\cite{radford2021learningtransferablevisualmodels, li2023blip2bootstrappinglanguageimagepretraining, ramesh2021zero, yu2022scaling, betker2023improving, videoworldsimulators2024} where visual data are paired with captions. As shown in Figure~\ref{fig:chat5_jet}, in particle physics jet tagging, for example, each final-state particle's numerical measurements (charge, energy, momentum components, impact parameters, etc.) are paired with a textual annotation such as "charged pion" or "neutral hadron", forming an experimental-data-to-text alignment analogous to two modalities image-caption pairs\cite{radford2021learningtransferablevisualmodels}. Similarly, as shown in Figure~\ref{fig:chat_material}, in materials science, large-scale experimental or simulated datasets in numerical format can be systematically converted into natural language descriptions and embedded within theoretical context. For instance, in the case of the Ag\textsubscript{2}SnYb crystal structure, the raw numerical data from original MPtrj format~\cite{deng2023chgnetpretraineduniversalneural} including lattice matrix values [4.9526721, 0.0, 0.0], [-2.47633605, 4.27679244, 0.00653309], [0.0, 2.85731761, 4.0453514], atomic fractional coordinates, and the chemical formula (Ag\textsubscript{2}SnYb), atomic composition (Yb: 1 atom, Sn: 1 atom, Ag: 2 atoms), lattice parameters ($a$ = 4.9527 \AA, $b$ = 4.942 \AA, $c$ = 4.9527 \AA, $\alpha$ = 59.977$^\circ$, $\beta$ = 90.0$^\circ$, $\gamma$ = 120.0715$^\circ$), and unit cell volume (85.59 \AA\textsuperscript{3}) are decomposed and transformed into natural language descriptions. This creates a unified sequence where raw numerical values coexist with their semantic interpretations, enabling the model to learn correspondences between empirical observations and physical meanings. Beyond these immediate annotations, the framework incorporates deeper theoretical explanations such as QCD principles, quark-gluon dynamics in particle physics and density functional theory\cite{dft}, electronic structure\cite{munro2020improved} in condensed matter physics from general scientific corpora like Wikipedia and research literature. During pretraining, these theoretical concepts are placed in the same context as experimental data sequences, creating a dual alignment structure with immediate data-caption pairs at the local level and comprehensive theoretical explanations at the global level. The sequence-based auto-regressive language model learns patterns in experimental data, which are traditionally captured by specialized models like GNNs\cite{Zhu2023ParticleNetAI, deng2023chgnetpretraineduniversalneural} or numerical analysis models\cite{graphcast}, and aligns numerical observations with theoretical concepts in a unified context, enabling language-guided scientific computing through integrated pattern recognition and language reasoning. Overall, the paradigm comprises several key components. First, embed large-scale numerical, tabular, or time series experimental data within theoretical contexts through textual representation. Second, controlling the ratio between theoretical and experimental inputs during pre-processing according to domain-specific characteristics. For example, in particle collision studies direct experimental results for analysis are typically compressed from vast amounts of raw binary data. Third, employ the Binary Patch Encoding method(see Section~\ref{tokenization}) to handle both language and numerical inputs without information distortion, as previously introduced in BigBang-Neutron (named as BBT-Neutron in the paper) \cite{wu2024scalingparticlecollisiondata}. Fourth, we transform a wide range of scientific problems, primarily classification and regression tasks such as particle jet tagging, electronic structure calculations, genome structure prediction, and PDE solving, into language-guided computing tasks during both training and inference, framing them as next-word prediction within a natural language context; and fifth, utilize Monte Carlo Attention(see Section~\ref{Monte Carlo Sampling Block}) to achieve ultra-large context lengths sufficiently to accommodate material structures across scales. By aligning theoretical constructs and experimental data as sequences, Theory-Experiment Learning can achieve task-agnostic and discipline-agnostic pretraining and inference across different scales and structures in physical world, and provides a framework for building foundational model particularly for Big Science experiments which handle large-scale experimental data, including high energy physics\cite{lhc}, nuclear physics\cite{nif}, cosmology and astronomy\cite{Gardner_2006}, computational fluid dynamics\cite{cfdsurvey} and biomedicine\cite{venter2001sequence}.

\begin{table}[H]
    \centering
    \small
    \caption{Overview of the Training Dataset}
    \label{tab:data_sizes}
   \begin{tabularx}{\textwidth}{>{\ttfamily}l>{\raggedright\arraybackslash}X>{\raggedright\arraybackslash}X>{\raggedright\arraybackslash}X}
        \toprule
        \textbf{Data Subset} & \textbf{Fields/Elements} & \textbf{Dataset Sample} & \textbf{Size (Bytes)} \\
        \midrule
        SlimPajama & General text corpus & In hadronization quarks and gluons combine to form hadrons & 230B \\
        \midrule
        Arithmetic Operations & Addition, Subtraction, Multiplication & 
        123123457457352354 + 7467458472832 = 4+2,5+3,3+8,2+2,... 
        result=123130924915825186 & 1.7B \\
        \midrule
        Particle Physics & $\Delta\eta$, $\Delta\phi$, $\log P_t$, $\log E$ & $\log P_t$ (5.525968, 23.885775, 10.332325),$\Delta\phi$ 3.281097, $\log E$ 26.604952 & 20B \\
        \midrule
        Material Structures & Number of Atoms, atomic mass, Atomic charge, Atomic Positions & NumberOfAtoms:16; Position:[0.94437093 1.87348563 3.77456225], Mass:158.92535, InitialCharge:0.0 & 641M \\
        \midrule
        Genomics & Gene Sequences & TGATTTTT······ACCA TCATTATATTTT TCACCAGCG & 27.2B \\
        \midrule
        Sensors & Light Intensity, Voltage, Temperature, Speed & 3.84322, 350.84631, 25.55369, 0.27954 & 262M \\
        \midrule
        Stock Prices & Date, Open, High, Low, Close, Volume & 2023-10-01, 150.25, 152.30, 149.80, 151.90, 2500000 & 60B \\
        \midrule
        Python Code & Code snippets & def add(a, b): return a + b
        Function to add two numbers  & 1.1B \\
        \midrule
        Post-Training Data & Text-book level and Instruction data & Explain quantum mechanics Quantum mechanics is... & 38B \\
        \bottomrule
    \end{tabularx}
\end{table}
	\subsection{Dataset}

Our training corpus for pre-training and post-training stages is carefully curated to support the development of a scientific multi-task learner capable of reasoning across diverse scales and structures of physical world, including natural language, mathematics, physics, materials science, genomics, sensor dynamics, finance, and programming. The corpus consists of nine major data subsets, each designed to instill domain inductive biases and task-specific knowledge, as summarized in Table~\ref{tab:data_sizes}: 

\begin{enumerate}
    \item \textbf{General Text Corpus (SlimPajama)} \\
    Sourced from the SlimPajama dataset\cite{cerebras2023slimpajama}, this component provides broad linguistic and conceptual coverage, including web text, books, scientific articles, and technical Q\&A. It consists of 26.7\% C4, 5.2\% GitHub, 4.2\% Books, 4.6\% ArXiv, 3.8\% Wikipedia, and 3.3\% StackExchange. This diverse mix ensures exposure to both general knowledge and domain-specific scientific literature, forming a robust cross-discipline context of theory for Theory-Experiment Learning.

    \item \textbf{Arithmetic Operations} \\
    We generate 300 million synthetic arithmetic examples (100 million each for addition, subtraction, and multiplication) involving integers up to 50 digits. These operations are computed exactly using Python’s arbitrary-precision arithmetic. Training model on arithmetic operations involving up to 50-digit numbers is unusual in real-world applications. It is designed to stress-test model's numerical reasoning, probing the hypothesis that poor numerical performance stems from byte-pair-encoding(BPE) tokenization, which distorts numbers' representation. By replacing BPE with a Binary Patch Encoding method that preserves numerical semantics, we test whether robust arithmetic computation can be recovered, identifying root causes highlighted by failures like ChatGPT’s 9.11>9.9 error~\cite{lingme2023chatgpt}.

\item \textbf{Particle Physics (Jet Tagging Data)} \\
This dataset simulates high-energy particle collisions in which quarks and gluons fragment into detectable final-state particles, forming particle jets. It has been used in Jet Origin Identification (JOI)~\cite{Zhu2023ParticleNetAI} and our previous work, \texttt{BigBang-Neutron}~\cite{wu2024scalingparticlecollisiondata}. For each jet, the input features include the electric charge (\texttt{charge}), energy (\texttt{energy}) and its logarithm (\texttt{log10(energy)}), the three components of momentum (\texttt{Px, Py, Pz}) and the logarithm of transverse momentum (\texttt{log10(pt)}), the pseudorapidity and azimuthal angle differences relative to the jet axis ($\Delta\eta$, $\Delta\phi$), the logarithmic ratios of particle-to-jet transverse momentum and energy ($\log p_T^\text{rel}$, $\log E^\text{rel}$), the angular separation ($\Delta R$), and the track impact parameters with their uncertainties ($d_0$, $d_{0\text{err}}$, $z_0$, $z_{0\text{err}}$). The data covers 11 jet types corresponding to $b$, $\bar{b}$, $c$, $\bar{c}$, $d$, $\bar{d}$, $g$, $s$, $\bar{s}$, $u$, $\bar{u}$, $gluon$, with 10 million jets for each type.

    \item \textbf{Material Structures} \\
    We use MPtrj data\cite{deng2023chgnetpretraineduniversalneural} for material structure-property prediction and new material discovery tasks. MPtrj comprises 1.5 million crystalline and molecular structures, including the chemical formula, elemental composition (atom counts), lattice matrix (3$\times$3 vectors), lattice parameters ($a$, $b$, $c$, $\alpha$, $\beta$, $\gamma$), unit cell volume, and atomic site coordinates, total energy, formation energy, etc. These data enable the model to learn 3D spatial relationships, symmetry, and structure-property mappings critical for materials discovery and quantum chemistry.

    \item \textbf{Genomics Sequences} \\
    We use the OpenGenome dataset~\cite{parks2015checkm, camargo2023imgvr, camargo2024imgpr}, a large-scale genomic corpus aligned with the Evo model's pretraining data~\cite{dEric2024evo}, comprising over 80,000 bacterial and archaeal genomes and millions of predicted prokaryotic phage and plasmid sequences, totaling approximately 300 billion nucleotide tokens. The dataset is compiled from three primary sources: (1) representative genomes from the Genome Taxonomy Database (GTDB) v214.1~\cite{parks2015checkm}, with one genome per species selected to minimize redundancy; (2) curated prokaryotic viruses from IMG/VR v4~\cite{camargo2023imgvr}; and (3) plasmid sequences from IMG/PR~\cite{camargo2024imgpr}. It includes DNA, RNA, and protein sequences from diverse organisms, enabling the model to capture evolutionary patterns, gene structure, and biochemical constraints. For \texttt{BigBang-Proton} pretraining, we use a subsample of 27.2 billion tokens, supporting biological sequence modeling and downstream applications of molecular and system scales predictions. 

    \item \textbf{Spatiotemporal Sensor Data} \\
    Real-world sensor logs record time-series measurements of light intensity, voltage, temperature, and speed. These continuous, high-frequency signals form a spatiotemporal learning substrate, enabling the model to infer dynamic patterns, detect anomalies, and extrapolate physical behaviors in real-world environments. In downstream applications, BigBang-Proton is further fine-tuned on lake water quality spatiotemporal data and successfully predicts future water quality fluctuations (see~\ref{subsec:lake_water_prediction}).

    \item \textbf{Stock Prices} \\
    This subset contains fifteen years of historical price data from both A-shares (China) and U.S. equities, including daily open, high, low, close prices, volume, and timestamps. By exposing the model to economic time series with stochastic trends, volatility clustering, and market regimes, we strengthen its temporal reasoning capabilities.

    \item \textbf{Python Code} \\
    Extracted from public GitHub repositories, this subset includes short code snippets, function definitions, and inline comments. It enables the model to learn computational syntax, logic flow, and programming semantics, facilitating code generation, debugging, and algorithmic problem solving.

    \item \textbf{Post-Training Data} \\
    We curate a high quality text book level dataset for annealing phases of post-training, including data sources from Open Phi\cite{openphi}, alpaca-gpt4\cite{peng2023instruction}, and synthetic dataset\cite{sciphi2024textbooks, jtatman2024textbooks} following Phi\cite{gunasekar2023textbooksneed}. We use the Infinity-Instruct dataset\cite{BAAI_2024_Infinity-Instruct} in the supervised finetuning. It provides 7.4M foundational and 1.5M chat instructions covering explanation, reasoning, and task execution, enabling effective alignment with human intent while maintaining broad scientific and technical coverage.
    
\end{enumerate}

Incorporating general human languages into the pre-training and post-training corpus is essential for model to develop capabilities of language-guided scientific computing. These capabilities enable the model to interpret complex instructions and apply them effectively to guide scientific tasks. Through human language-guided classifications and regression, the model can be instructed to classify different particle interactions or predict material properties based on given parameters using natural language commands. This approach bridges the gap between human understanding and computational analysis, making the model more versatile and applicable across various scientific domains. For more details on the dataset, please refer to \ref{subsec:pretraining_dataset}.

During the pre-training phase of BigBang-Proton, data from multiple disciplines is concatenated to form a unified input sequence. A sample input might look like `General Text: [Contextual paragraph] <|im$\_$end|> DNA Sequence: ATCGGCTA... <|im$\_$end|>Particle Interaction: $\Delta\eta$: 0.5, $\Delta\phi$: 1.2,  $\log P_t$: 3.0, $\log E$: 4.5<|im$\_$end|>'. All these inputs are concatenated together into a single sequence for training. This approach ensures that both textual and numerical data are properly aligned and processed.
    
To illustrate how language and experimental data are aligned and processed, we will consider examples from particle physics. Consider an experiment where particle interactions are recorded with specific parameters $\Delta\eta$ 0.5, $\Delta\phi$ 1.2,  $\log P_t$ 3.0, $\log E$ 4.5, This numerical data can be aligned with its physical meaning through linguistic descriptions: Energy of the particle ($\log E$),  Logarithm of the particle's energy (log10(energy)), Logarithm of the particle's transverse momentum ($\log P_t$), Difference in pseudorapidity between the particle and the jet axis ($\Delta\eta$), Difference in azimuthal angle between the particle and the jet axis ($\Delta\phi$).  Integrate these descriptions directly into the input sequence, using special tokens (`<|im$\_$end|>`) to separate different types of data while maintaining coherence,which allow the model to learn cross-disciplinary relationships and patterns.
    
\subsection{Binary Patch Encoding} \label{tokenization}
\begin{figure}[H]
    \centering
    \includegraphics[width=1\textwidth]{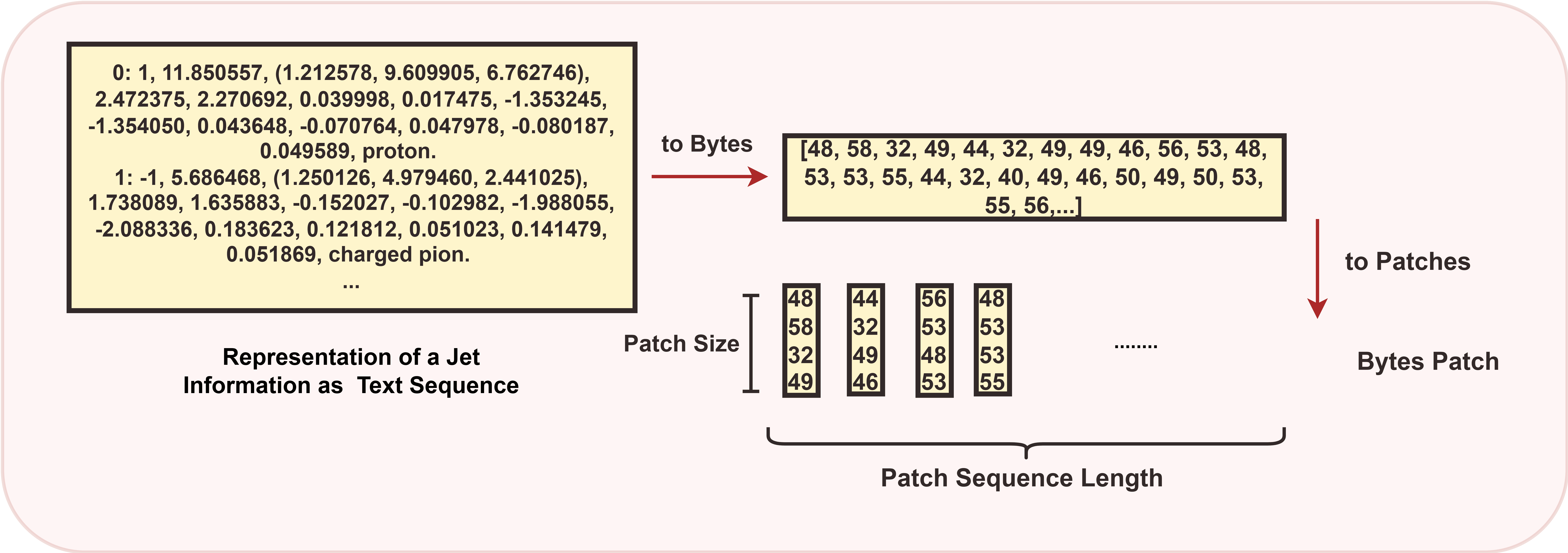}
    \caption{Binary Patch Encoding: how a particle jet is converted into a series of byte tokens. This multi-modality native method converts all dataset formats originally stored in bits to byte sequences using UTF-8. Patching is introduced to reduce computational complexity. Binary Patch Encoding eliminates the tokenization process and vocabulary requirements.} 
\label{fig:tokenization_scheme}
\end{figure}

Traditional tokenizer BPE\cite{sennrich2016neuralmachinetranslationrare}, sentence piece\cite{kudo-richardson-2018-sentencepiece} and word piece\cite{Schuster2012JapaneseAK} are replaced by Binary Patch Encoding, which is designed to handle diverse data modalities, including text, scientific symbols, numerical data, and images, ensuring uniform processing across all types. This approach builds upon our previous work~\cite{wu2024scalingparticlecollisiondata},  as well as other significant contributions in the field like BGPT \cite{wu2024languagemodelsbytemodels}, Megabyte \cite{yu2023megabytepredictingmillionbytesequences}, SpaceByte \cite{slagle2024spacebytedeletingtokenizationlarge}, and BLT \cite{pagnoni2024bytelatenttransformerpatches}. These works are pioneering byte-level models that explore the effectiveness of byte-based encoding for various data types. This method leverages the fact that all data are ultimately stored in binary format in computers, allowing us to process different data types consistently. Figure~\ref{fig:tokenization_scheme} illustrates the input encoding scheme used in Particle Physics, originally presented in our previous work\texttt{BigBang-Neutron} \cite{wu2024scalingparticlecollisiondata}. The figure provides a visual representation of how different types of data are encoded into a consistent byte-based format.

Byte Pair Encoding (BPE)~\cite{sennrich2016neuralmachinetranslationrare} and its variants\cite{kudo-richardson-2018-sentencepiece, Schuster2012JapaneseAK} have been widely used for tokenizing textual data. However, BPE has notable limitations when it comes to handling numerical data.  Byte Pair Encoding(BPE) can introduce ambiguity and inconsistency in tokenizing numbers, leading to different segmentations of the same number based on context. For instance, the same number 12345 might be tokenized as `12', `34', and `5' in one context, or as `1', `23', and `45' in another due to the mechanism of adjacent pair compression, losing the inherent meaning of the original numeric value. Additionally, BPE results in fragmented token IDs for numerical entities, such as `7' and `8' being assigned token IDs 4779 and 5014, respectively. This discontinuity in token IDs complicates the management and processing of numerical data, especially when sequential or patterned token IDs are beneficial. Similarly, single-digit tokenization 
 \cite{dubey2024llama}, while straightforward, also leads to discontinuous token IDs for multi-digit numbers, such as 15 being broken down into separate tokens `1' and `5',  each of which is then mapped to an independent token ID. This fragmentation can disrupt the continuity and authenticity of numerical information, potentially making it more challenging for the model to capture the inherent structure and relationships within multi-digit numbers.
These methods are not designed to effectively encode numerical values, which often results in suboptimal performance when processing datasets rich in numerical information(see more discussions in ~\ref{subsec:arithmetic}). This limitation is particularly pertinent to our cross-disciplinary scientific model, where the ability to accurately process and understand numerical data, such as those found in financial documents, particle physics, material lattice, and genomic sequences, is crucial for achieving robust and versatile predictive capabilities. Therefore, while BPE remains a powerful tool for text tokenization, its inadequacy with numerical data underscores the necessity for developing more advanced input encoding strategies tailored to the diverse needs of scientific datasets.
In contrast, our Binary Patch Encoding addresses the fundamental limitations of traditional tokenization by providing a unified, integrity-preserving, and flexible approach to input representation. By treating all data including text, code, numbers, and scientific symbols as raw byte sequences, it ensures uniformity across modalities, eliminating the need for modality-specific tokenization schemes and streamlining the preprocessing pipeline. Operating directly on binary representations guarantees preservation of data integrity, as no information is lost or distorted through segmentation. This is particularly critical for numerical reasoning, where conventional subword tokenizers (e.g., SentencePiece~\cite{kudo-richardson-2018-sentencepiece}) can fragment numbers like \texttt{9.11} into \texttt{9}, \texttt{.}, \texttt{11}, leading to erroneous comparisons such as $9.11 > 9.9$ as observed in mainstream LLMs~\cite{lingme2023chatgpt}. Our method brings simplicity by preserving numbers in their native format without unnecessary fragmentation into sub-word tokens, enabling accurate arithmetic computation, while also offering flexibility to handle diverse and novel data types without complex adaptations. Together, these advantages establish Binary Patch Encoding as a robust foundation for language-guided scientific computing.

Many scientific datasets are stored in native binary formats (e.g., \texttt{.bin}, \texttt{.dat}). We have experimented with representing input data as sequences of bits (i.e., $0$s and $1$s) instead of bytes (values from $0$ to $255$) and therefore named this approach Binary Patch Encoding, but deferred due to the increase in computational complexity. Currently, \texttt{BigBang-Proton} uses UTF-8 byte sequences. A direct binary encoding remains a key goal for enabling models to learn from the building block of information.
        
\subsection{Architecture}
The BigBang-Proton architecture consists of three core components. Input Embedding transforms discrete input tokens into continuous, dense vector representations, enabling the model to capture semantic relationships. Monte Carlo Attention enhances information flow through Inter-Patch Delegation mechanism to enable extremely long sequence length with controllable computational complexities. Temporal Convolutional Network (TCN) captures sequential dependencies and temporal patterns through hierarchical convolutional layers, getting rid of the need of positional encoding.

\subsubsection{Input Embedding}
        
\textbf{Input Representation} The input to the embedding module is a sequence of discrete tokens, where each token is represented by an integer value. 
        
Byte values range from 0 to 255, representing the raw input data.Special tokens include Padding token, Begining of sequence token and End of Sequence token. Padding token (pad id: 258): Used to pad sequences to a fixed length. Beginning of sequence token (bos id: 256): Marks the start of a sequence. End of sequence token (eos id: 257): Marks the end of a sequence. Thus, the input vocabulary size is 259, covering all possible byte values and special tokens.
        
\textbf{One-Hot Encoding} Each input token is first converted into a one-hot encoded vector of dimension 259. For a token with value $i$ (where $0 \leq i \leq 258$), the one-hot encoded vector is a sparse binary vector of size 259, where the $i$-th element is 1, and all other elements are 0. Mathematically, the one-hot encoding function can be expressed as:
$$
\text{one\_hot}(i) = [\underbrace{0, \dots, 0}_{i \text{ zeros}}, 1, \underbrace{0, \dots, 0}_{258-i \text{ zeros}}] \in \mathbb{R}^{259},
$$
where the value 1 is located at the $i$-th position.         
\textbf{Linear Projection} The one-hot encoded vectors are then projected into a dense, continuous vector space of dimension $D$, where $D$ is the embedding dimension of the model. This is achieved using a learnable linear transformation (fully connected layer) without bias. Mathematically, the transformation can be expressed as:
$$
\text{embedding}(i) = W \cdot \text{one\_hot}(i),
$$      
where:
\begin{itemize}
    \item $W \in \mathbb{R}^{D \times 259}$ is the weight matrix of the linear layer.
    \item $\text{embedding}(i) \in \mathbb{R}^D$ is the resulting dense embedding vector for token $i$.
\end{itemize}

\subsubsection{Monte Carlo Attention} \label{Monte Carlo Sampling Block}

Monte Carlo Attention is designed to achieve ultra-large context lengths for modeling physical structures with computationally feasible complexity in real-world settings. Monte Carlo Attention is a type of sparse attention that serves as an alternative to vanilla transformers with full attention computations. Physical structures are fundamentally different from natural languages, which typically require context lengths many orders of magnitude larger than those used in mainstream LLM pretraining. Therefore, a fundamental architectural redesign is necessary. 
\begin{figure}[tbp]
		\centering
		\includegraphics[width=1\textwidth]{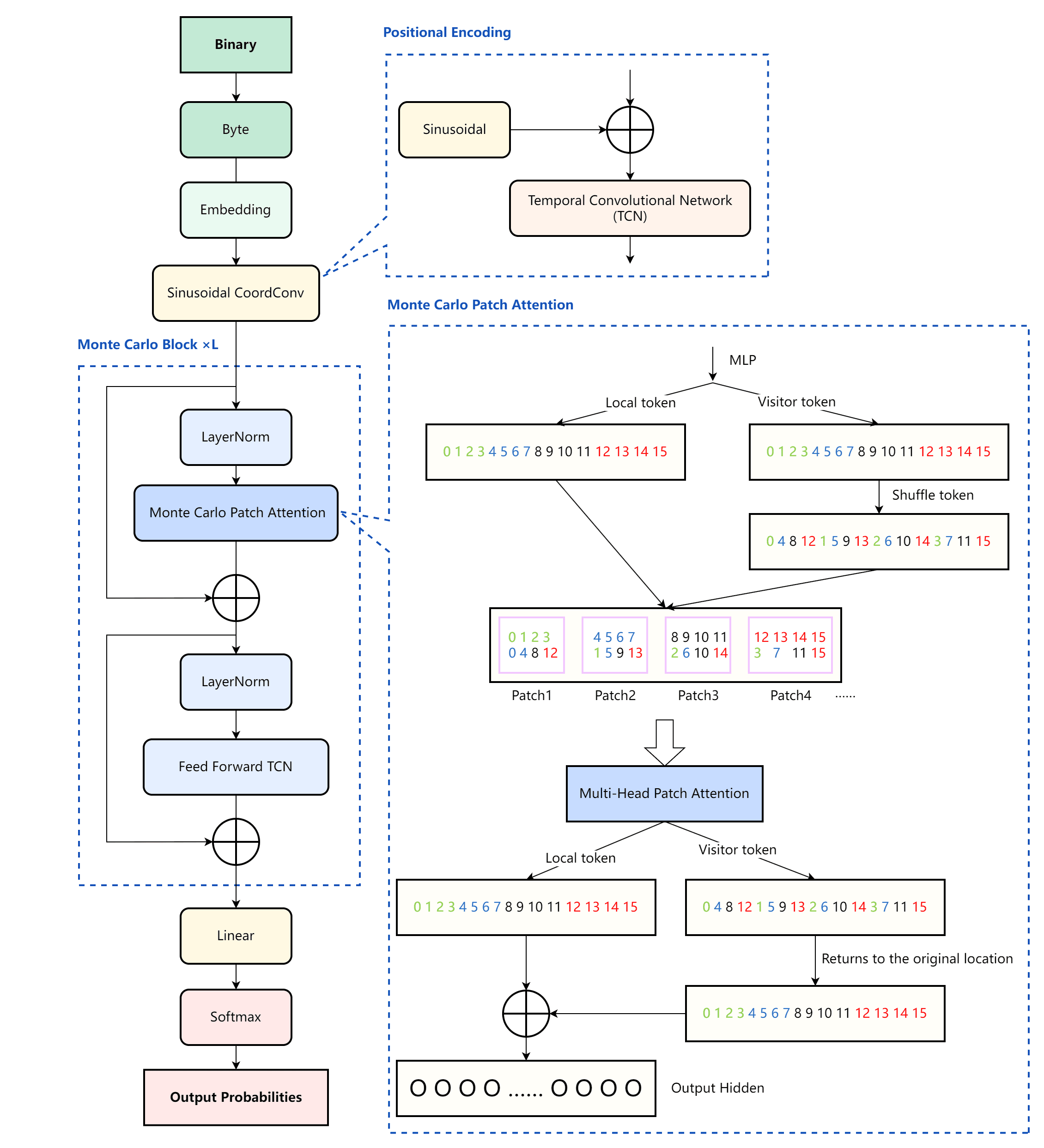} 
		\caption{Overview of Monte Carlo Attention architecture. The model consists of three main components: (1) input embedding, which converts discrete input tokens into dense vector representations; (2) Monte Carlo Attention, which utilizes an inter-patch-delegation mechanism to drive local and global information exchange, leading to context length growth proportional to the power of layer numbers while maintaining linear computational complexity; and (3) a Feed Forward temporal convolutional network (TCN), which replaces traditional Feed Forward fully connected networks in transformers and captures local spatial and temporal patterns. Since TCN learns positional information, positional embeddings used in transformers are eliminated.}
		\label{fig:proton_overview}
\end{figure}
\paragraph{Sparse Attention} The expansion of sequence length in large language models presents significant computational challenges due to the quadratic scaling behavior of standard attention mechanisms \cite{vaswani2017}. This computational burden has motivated extensive research efforts focused on enhancing efficiency while maintaining model performance. A key research direction exploits the naturally occurring sparsity in attention distributions, which manifests both mathematically through softmax operations \cite{Jiang2023} and biologically through neural connectivity patterns \cite{Watson2025}.
Current methodologies typically employ either fixed structural limitations, including sink-based \cite{Xiao2023} or sliding window approaches \cite{Beltagy2020}, or dynamic mechanisms that adaptively choose token subsets during runtime, such as Quest \cite{Tang2024}, Minference \cite{Jiang2024}, and RetrievalAttention \cite{Liu2024}. While these techniques reduce computational demands for extended sequences, they fail to address the substantial training overhead associated with long-context models, limiting scalable deployment to million-token contexts.
To address these limitations, sparse attention frameworks like MoBA \cite{lu2025mobamixtureblockattention} and NSA \cite{yuan2025nativesparseattentionhardwarealigned} introduce efficient solutions. MoBA applies the Mixture of Experts paradigm \cite{Shazeer2017} to partition context into blocks, using gating to route queries to relevant blocks, reducing computational costs. NSA employs compression and selection strategies for compact key-value representations, optimizing inference and training on GPUs. Both enhance long-context model efficiency.
Concurrently, linear attention architectures, exemplified by Mamba \cite{Dao2024}, RWKV \cite{Peng2023, Peng2024}, and RetNet \cite{Sun2023}, substitute softmax computations with linear transformations to lower overhead. However, adapting these to existing Transformers requires costly conversion \cite{Mercat2024, Wang2024, Bick2025, Zhang2024} or retraining \cite{Li2025}, with limited validation in complex reasoning tasks.
The growing recognition of long-context capabilities as essential for next-generation models is driven by the need to model physical material structures, such as material crystals, cells, ecosystems, or galaxies, which require ultra-long sequences exceeding billions or trillions of tokens, far beyond current million-token context limits \cite{Zaheer2020}. Applications in complex reasoning \cite{deepseekr1, Zelikman2022}, code generation \cite{Zhang2023a}, and autonomous agents \cite{Park2023} further underscore this challenge. Empirical analysis shows standard attention consumes majority of inference latency for 64k-token sequences \cite{Zaheer2020}. Ultra-long sequence methods that achieve hundreds of trillions of tokens while reducing computational requirements are needed. 

\paragraph{Monte Carlo Attention} 

Context length, defined as the limit one layer of transformer can read in one full attention computation, is independent of the depth of transformer layers. Direct information flow in transformer are confined to tokens within context length. In pretraining, information flow between batches rely on shared weights, rather than attention computation. In contrast, receptive field of convolutional neural network (CNN) extends with the depth of network\cite{luo2017understandingeffectivereceptivefield}. We are inspired to employ layer-wise operation to enhance information flow among all of input embeddings even between batches. The key innovation of Monte Carlo attention lie in a delegation operation in each layer, leading to dynamic token reorganization. 

\begin{figure}[H]
    \centering
    \includegraphics[width=1\textwidth]{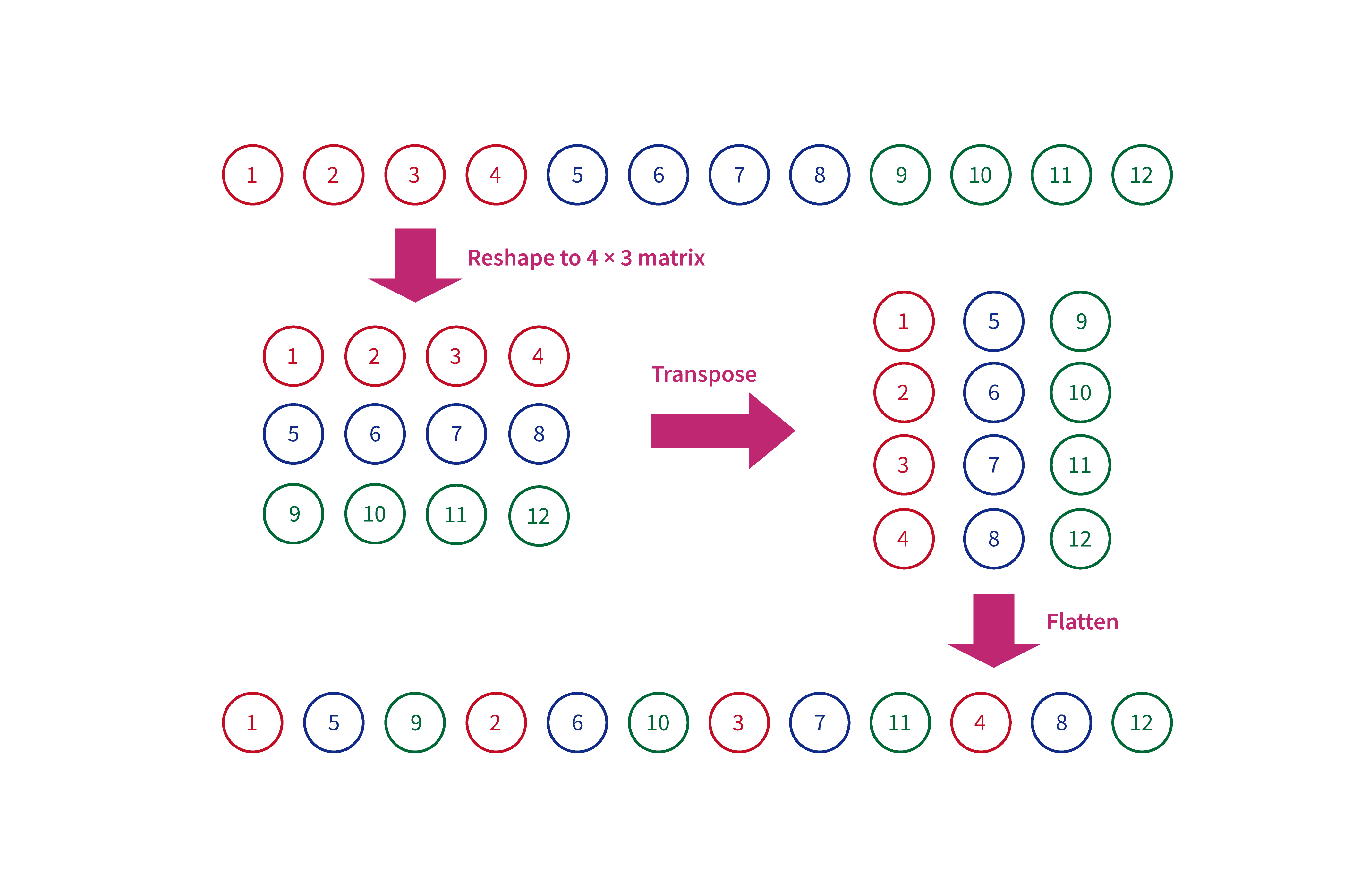} 
    \caption{Embedding vectors are reorganized between patches. Each patch sends delegates to and receives delegates from other patches for information exchange through attention computations.}
    \label{fig:shuffle}
\end{figure}

\paragraph{Inter-Patch Delegation Mechanism} The input embeddings are grouped into patches of a specific patch size, unlike the approach commonly employed in byte-level language models\cite{pagnoni2024bytelatenttransformerpatches, yu2023megabytepredictingmillionbytesequences, slagle2024spacebytedeletingtokenizationlarge, wu2024languagemodelsbytemodels} which implement patching before converting to embedding vectors. Drawing inspiration from representative political systems, during each layer-wise operation, each patch composed of multiple embeddings from $P$ bytes of input randomly (Monte Carlo) or selectively designates one byte as a representative to exchange information with other patches. Following the delegation process among all patches, each patch receives at most $P-1$ representatives from other patches, having sent the same number of delegates to other patches while maintaining a constant number of bytes within the patch. Attention computations are executed within each patch, resulting in $\mathcal{O}(P^2)$complexity. Each representative byte, containing contextual information from both its originating patch and the patches from which other delegates were received, returns to its original patch for subsequent attention computation. In the toy model illustrated in Figure~\ref{fig:shuffle}, we set $P=4$ for 3 patches (red, blue, and green). The sequence of 12 bytes is reshaped into a $4 \times 3$ matrix, which is then transposed to a $3 \times 4$ matrix and subsequently flattened back into a sequence of 12 bytes. This reorganization transforms the sequence $[1,2,3,4]$ to $[1,5,9,2]$ , where bytes 1 and 2 originate from the red patch, byte 5 from the blue patch, and byte 9 from the green patch. Through delegation and reorganization, information flows globally with computational complexity dependent only on the square of the patch size.

\begin{figure}[H]
    \centering
    \includegraphics[width=1\textwidth]{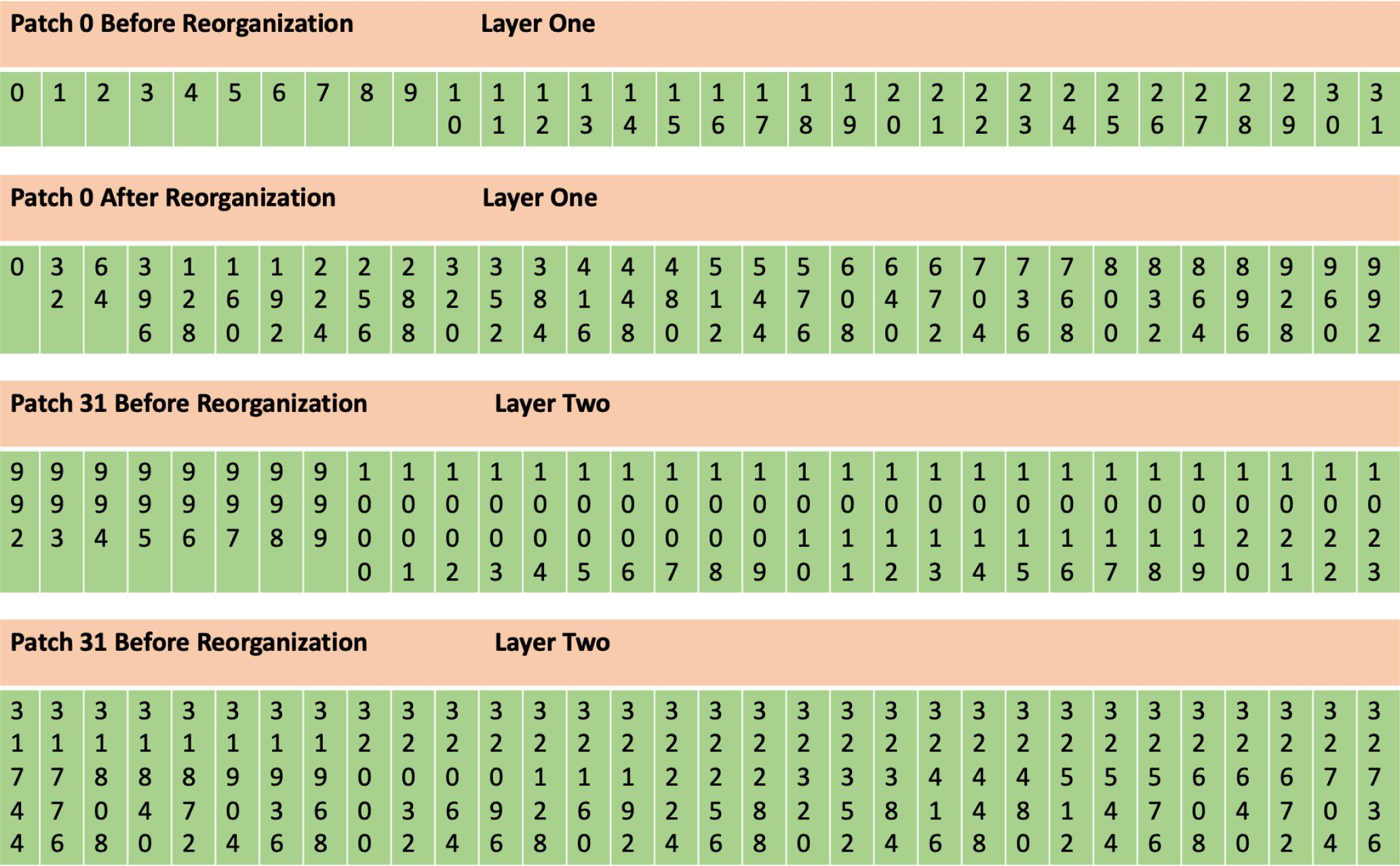} 
    \caption{Layer-wise inter-patch delegation operations drive the context length of information flow to increase by $P^{N+1}$, where $P$ is the patch size and $N$ is the number of layers. For patch size=32,  in layer one, information can reach 992, and in layer two, 32736.} 
    \label{fig:shuffle32}
\end{figure}

Figure~\ref{fig:shuffle32} demonstrates how the context length of information flow increases by layer-wise delegation operation inter-patch. Set patch size $P$ to 32 and length of sequence to $40906$, then each patch can hold $P-1$ delegated bytes for information exchanges, and each delegate contain contextual information of $P$ length. We can find local information in Patch $0$ and Patch $31$ in Figure~\ref{fig:shuffle32}. After delegation operation, Layer 1 gains context length $31 \times 32=992$, which increase sharply to $31 \times 32 \times 32+31 \times 32=32736$ in layer 2. When the depth of the layers becomes large, the recursive relationship for the effective context length after $N$ layers is given by:

\begin{equation}
    C(N) = (P-1) \times P^N + C(N-1)
    \label{eq:context_length}
\end{equation} 

with $P=32$ and $C(0)=0$, we can compute the context length for different numbers of layers. The results are shown in Table~\ref{tab:context_length}.

The delegation operation can be formally defined as a hierarchical process consisting of four key steps that mirror the actual computational flow. First, the input sequence  $\mathbf{X} \in \mathbb{R}^{B \times D \times L}$ is decomposed into $N_P$ patches where each patch contains $P$ tokens:
\begin{equation}
    \mathcal{N_P}: \mathbf{X}_{(L)} \rightarrow \{\mathbf{X}_i\}_{i=1}^{N_P}
\end{equation}
where $\mathbf{X}_{(L)}$ represents the input sequence of length $L$, $B$ is the batch size, $D$ is the hidden dimension, and $\{\mathbf{X}_i\}_{i=1}^{N_P}$ denotes the decomposition into $N_P$ patches each containing $P$ tokens such that $L = N_P \times P$.

Second, a $1 \times 1$ convolution operation is applied to each patch to generate delegate tokens that will be sent to other patches, and the delegation mapping selects and distributes representative tokens from each patch:
\begin{equation}
    \mathbf{d}_i = \text{Conv}_{1 \times 1}(\mathbf{P}_i)
    \label{eq:delegation_generation}
\end{equation}
\begin{equation}
    \mathcal{D}: \mathbf{P}_i \rightarrow \{r_{i,j}\}_{j=1}^{N_P}
    \label{eq:delegation}
\end{equation}
where $\mathbf{d} \in \mathbb{R}^{B \times D \times N_P}$ represents the group of delegate tokens, $\mathbf{d}_i$ denotes the delegate token generated from the $i$-th patch $\mathbf{P}_i$, and $r_{i,j}$ represents the representative byte selected from patch $\mathbf{P}_i$ for patch $\mathbf{P}_j$. These delegate tokens are permuted and reshaped to align with the original local patches.

Third, the delegate tokens are distributed to other patches and concatenated with the original local tokens to form an augmented representation that contains local and global information:
\begin{equation}
    \mathbf{C}_i = \text{Concat}(\mathbf{X}_i, \{\mathbf{d}_j | j \in \mathcal{N}(i)\})
    \label{eq:concatenation}
\end{equation}
where $\mathbf{C}_i$ represents the augmented context for the $i$-th patch, and $\{\mathbf{d}_j | j \in \mathcal{N}(i)\}$ denotes the set of delegate tokens received from neighboring patches $\mathcal{N}(i)$.

Finally, self-attention is computed over the augmented representation to facilitate information exchange between local and global contexts. The vanilla self-attention mechanism can be described as:
\begin{equation}
    \text{Attention}(\mathbf{Q}, \mathbf{K}, \mathbf{V}) = \text{Softmax}\left(\frac{\mathbf{Q} \mathbf{K}^T}{\sqrt{d_k}}\right) \mathbf{V}
    \label{eq:attention}
\end{equation}
where $\mathbf{Q}, \mathbf{K}, \mathbf{V}$ are the query, key, and value matrices derived from the augmented representation $\mathbf{c}$ through linear projections, and $d_k$ is the dimensionality of the keys.

Combining with the inter-patch delegation operation, the attention computation at each layer can be formally expressed as:
\begin{equation}
    \mathbf{A}^{(l)}_i = \text{Attention}(\mathbf{Q}^{(l)}_i, \mathbf{K}^{(l)}_i, \mathbf{V}^{(l)}_i)
    \label{eq:layer_attention}
\end{equation}
where $\mathbf{A}^{(l)}_i$ represents the attention output for the $i$-th patch at layer $l$, and $\mathbf{Q}^{(l)}_i, \mathbf{K}^{(l)}_i, \mathbf{V}^{(l)}_i$ are the query, key, and value matrices for the $i$-th patch after the delegation operation.

The final output hidden state is computed as the sum of the local patch representation and the reorganized patch
representation, using a residual connection, ensuring stable gradient propagation and information persistence:
\begin{equation}
    \mathbf{Y}_i = \mathbf{X}_i + Linear(\text{Attention}(\mathbf{C}_i))
    \label{eq:attention_computation}
\end{equation}
where $\mathbf{Y}_i$ represents the final output for the $i$-th patch after attention computation and residual connection\cite{he2015deepresiduallearningimage}, and Linear represents a linear transformation applied after the attention computation. 

The patch reorganization transformation can be expressed as a bijective mapping:
\begin{equation}
    \mathbf{T}: \mathbb{R}^{P \times N_P} \rightarrow \mathbb{R}^{N_P \times P}
    \label{eq:transformation}
\end{equation}

The attention computation complexity within each patch is:
\begin{equation}
    \mathcal{C}_{\text{patch}} = \mathcal{O}(4P^2)
    \label{eq:complexity}
\end{equation}

The global information flow complexity is:
\begin{equation}
    \mathcal{C}_{\text{global}} = \mathcal{O}(4N_P \cdot P^2)
    \label{eq:global_complexity}
\end{equation}
where $N_P$ is the number of patches.
\begin{table}[H]
\centering
\caption{Context length for different numbers of layers with $P=32$}
\label{tab:context_length}
\begin{tabular}{|c|c|c|}
\hline
\textbf{Layer Count (N)} & \textbf{Context Length C(N)} & \textbf{Comparison} \\
\hline
1 & 992 & - \\
2 & 32,736 & - \\
3 & 1,048,544 & - \\
4 & 33,554,400 & - \\
8 & 35,184,372,088,800 & GPT-4 corpus token number \\
10 & 36,028,797,018,963,936 & Atoms in human cell $\times$ 100 \\
20 & 3.741444191567e+30 & Atoms in Starship \\
40 & 4.056481920730e+60 & Atoms in Earth $\times$ 10 \\
60 & 4.398046511104e+90 & Bits in observable universe \\
80 & 4.767240170282e+120 & - \\
100 & 5.165034368751e+150 & - \\
\hline
\end{tabular}
\end{table}
\paragraph{Representation Degradation, Looped Attention and Sparsity} 

The vanilla transformer relies on full attention computation to let input embeddings gain representations from other tokens within a predefined context length. Monte Carlo attention employs an inter-patch delegation mechanism to achieve global representation exchange, achieving ultra-large effective context length. However, through Monte Carlo attention, representations can only exchange between patches indirectly through multiple attention computations. This indirect attention computation can lead to information degradation during the iterative representation propagation process. To address this limitation, we introduce looped Monte Carlo attention by repeating one layer operation several times. Looped transformers have been successfully applied to strengthen length generalization\cite{fan2025loopedtransformerslengthgeneralization} and latent reasoning\cite{saunshi2025reasoninglatentthoughtspower} through test-time compute optimization\cite{geiping2025scalingtesttimecomputelatent}. The underlying principle of looped attention is to leverage repeated representation exchanges to mitigate information loss and enhance representation quality during iterative attention computations.
\begin{align*}
\textbf{Vanilla Transformer:} &\quad \text{Layer}_1 \xrightarrow{} \text{Layer}_2 \xrightarrow{} \text{Layer}_3 \xrightarrow{} \text{Layer}_4 \xrightarrow{} \cdots \\
\textbf{Looped Monte Carlo Attention:} &\quad \text{Layer}_1^{\circlearrowleft N} \xrightarrow{} \text{Layer}_2^{\circlearrowleft N} \xrightarrow{} \text{Layer}_3^{\circlearrowleft N} \xrightarrow{} \text{Layer}_4^{\circlearrowleft N} \xrightarrow{} \cdots
\end{align*}

The computational complexity analysis in Table~\ref{tab:complexity_comparison} reveals a significant advantage of our Monte Carlo Attention architecture over traditional vanilla Transformers. The Monte Carlo Attention complexity is $O(N \times 4N_P \times P^2)$, where $N_P = \frac{\text{Context Length}}{P}$, which simplifies to $O(4N \times P \times \text{Context Length})$. In contrast, Vanilla Transformers exhibit quadratic scaling with $O(\text{Context Length}^2)$ complexity. This fundamental difference means that while vanilla Transformers become computationally prohibitive for very long sequences, Monte Carlo Attention maintains linear scaling with context length, making it particularly advantageous for processing ultra-large sequence tasks. The speedup becomes increasingly dramatic as context length grows, with Monte Carlo Attention achieving orders of magnitude reduction in computational requirements compared to standard attention mechanisms. As seen in ~\ref{Appendix:Computational Complexity Analysis}, Table~\ref{tab:complexity_comparison} shows that for a context length of 100 million with $P=32$, looping 32 times of Monte Carlo Attention can still achieve 99.9\% complexity reduction compared to full attention computation.

\subsubsection{Feed Forward TCN}

Complementing the attention module, we propose a temporal convolution block with enhanced pattern capture capabilities. This Temporal Convolutional Network (TCN) achieves multi-scale processing through stacked 1D convolutions with kernel size $K$, which extract hierarchical temporal features.

Let $\mathbf{X} \in \mathbb{R}^{B \times D \times L}$ denote the input tensor. The TCN applies multiple layers of 1D convolutions to this input. Each layer can be represented as:
$$
\mathbf{Y}^{(l)} = \text{ReLU}\left(\text{Conv}_{1 \times K}(\mathbf{Y}^{(l-1)}) + \mathbf{Y}^{(l-1)}\right)
$$
where $\mathbf{Y}^{(0)} = \mathbf{X}$, $\text{Conv}_{1 \times K}$ denotes the 1D convolution operation with kernel size $K$, and $\text{ReLU}$ is the activation function applied element-wise.

The stacking of these convolutional layers allows the network to capture hierarchical temporal features at different scales. Specifically, deeper layers in the network can capture longer-range dependencies due to the cumulative effect of successive convolutions.

Additionally, the proposed TCN maintains a local-global balance, while the attention mechanism handles long-range dependencies, the TCN focuses on fine-grained local pattern discovery. This is achieved by applying dilated convolutions in the TCN architecture, which allow the model to cover a wider receptive field without increasing the number of parameters significantly. For a dilation factor $d$, the convolution operation becomes:
$$
\mathbf{Y}^{(l)} = \text{ReLU}\left(\text{Conv}_{1 \times K}^d(\mathbf{Y}^{(l-1)}) + \mathbf{Y}^{(l-1)}\right)
$$
where $\text{Conv}_{1 \times K}^d$ represents the dilated convolution operation with dilation factor $d$.
Notably, due to the inherent convolutional nature of the TCN, our model possesses the capability to learn spatial and positional information directly from the input sequences, eliminating the need for explicit positional embeddings that are typically required in traditional Transformer architectures\cite{vaswani2017}.
Finally, the output of the TCN block is combined with the attended features from the attention module using a residual connection:
$$
\mathbf{Z} = \mathbf{Y}^{(L)} + \text{Attention}(\mathbf{C}_{(B, D, G, 2P)})
$$
where $\mathbf{Y}^{(L)}$ is the output of the last layer of the TCN, and $\text{Attention}(\mathbf{C}_{(B, D, G, 2P)})$ represents the attended features obtained from the attention mechanism.This design ensures that the model benefits from both the global context provided by the attention mechanism and the fine-grained local patterns captured by the TCN, leading to enhanced overall performance.

\subsubsection{Context Length for Physical World Modeling} 

Monte Carlo Attention was developed to address the theoretical requirements of the BigBang-Proton framework, although practical implementation faces hardware constraints. The underlying assumptions driving this development include several key considerations. First, for auto-regressive pretraining, Binary Patch Encoding as a native multi-modal approach can seamlessly convert all digital data formats into standard binary sequences, creating stringent demands for ultra-long context lengths. Second, the Theory-Experiment Learning paradigm offers potential for integrating experimental data from historical and ongoing scientific experiments across scales, structures, and disciplines during pretraining, requiring context lengths far exceeding those of pure natural language pretraining. Finally, in the ultimate scenario treating the universe as a single entity, if information from all atoms ($10^{80}$\cite{PhysRevLett.88.237901}) could be converted to a single sequence for pretraining, could context length achieve cosmos-scale?

The token length estimation for converting complex scientific structures to sequences reveals unprecedented scale requirements. For the comprehensive virtual cell integration, encompassing multi-omics data and cellular structures with approximately $10^{14}$ atoms, each requiring 10-20 tokens for complete representation including position, bonding, interactions and dynamic state information, the total sequence length reaches approximately $10^{15}$ tokens (1 quadrillion tokens)\cite{noutahi2025virtualcellspredictexplain, bunne2023build} . Similarly, for QCD modeling involving lattice QCD data with $\sim 10^9$ configurations, each containing $10^6$ sites and $\sim 100$ floating-point values per site, the total data of $\sim 10^{19}$ bytes translates to approximately $10^{20}$ tokens \cite{Gattringer2010} when accounting for complete parameter and theoretical description representation. Both scenarios demand context lengths orders of magnitude beyond current LLM capabilities ($10^{15}$-$10^{20}$ tokens versus typical $10^5$-$10^6$ token limits), necessitating novel attention mechanisms like Monte Carlo Attention to achieve the theoretical context length requirements for cosmos-scale scientific modeling and simulation.

To extend context window in pretraining, mainstream LLMs such as DeepSeek V3\cite{deepseekv3}, Qwen3\cite{yang2025qwen3technicalreport}, Llama3\cite{dubey2024llama} usually employ a long context pretraining stage to extend the context length from typically 4096 tokens to 128K tokens with a small fraction of total pretrained tokens. In contrast, Monte Carlo Attention enables theoretically infinite context length through its inter-patch delegation mechanism, fundamentally transforming pretraining methodologies. This breakthrough has profound implications for existing pretraining technologies and hardware designs. First, traditional batch processing constraints are alleviated\cite{he2025nondeterminism} as attention computations can be distributed across batches, enabling efficient processing of ultra-long sequences. Second, the reduced computational complexity from $\mathcal{O}(L^2)$ to $\mathcal{O}(L)$ significantly decreases the training steps required for convergence, potentially improving loss convergence rates and perplexity curves. Third, by decoupling context length from GPU memory limitations, Monte Carlo Attention enables training on sequences orders of magnitude longer than device memory capacity. Finally, this method facilitates the development of compute-in-memory architectures specifically designed for long-sequence processing. Without consideration of GPU memory constraints, this approach can achieve effective context lengths matching the full pretrained corpus sequence length. This paradigm shift necessitates the development of next-generation hardware architectures that can support truly cosmos-scale sequence processing.

\subsubsection{Architecture Advantages and Comparisons}

The fundamental distinction between Sparse Attention and Monte Carlo Attention lies in their core computational mechanisms. Sparse Attention approaches, including NSA\cite{yuan2025nativesparseattentionhardwarealigned} and MoBA\cite{lu2025mobamixtureblockattention}, employ selection-based mechanisms that filter key-value pairs to reduce computational complexity by choosing subsets of tokens for attention computation. NSA utilizes three sophisticated strategies: token compression through block-level aggregation, token selection via blockwise top-n identification, and sliding window mechanisms to preserve local context, dynamically constructing compact representations from N tokens by selecting top-K subsets. MoBA adopts a mixture-of-experts inspired approach by dividing context into blocks and applying top-k gating mechanisms for selective attention. In contrast, Monte Carlo Attention employs reorganization-based mechanisms through inter-patch delegation, structurally reorganizing token information by compressing global context into representative tokens that are exchanged between patches, enabling indirect information propagation through delegate token exchange rather than direct attention between selected tokens.

These core mechanism differences lead to critical disadvantages of sparse attention approaches. First, sparse attention suffers from selection bias and information loss where unselected tokens are discarded, leading to overlooked global dependencies and fragmented coordination, while Monte Carlo Attention preserves key information through controlled delegate mechanisms. Second, sparse attention encounters computational bottlenecks requiring $\mathcal{O}(K^2)$ complexity among selected tokens, while Monte Carlo Attention achieves efficient global exchange through local 2P attention with superior $\mathcal{O}(4N \times P \times \text{Context Length})$ complexity. Third, sparse attention has limited context modeling capability due to selection constraints, while Monte Carlo Attention enables exponential context length expansion.

Compared to other sparse attention mechanisms, our approach offers unique advantages. Random Sparse Attention \cite{huang2024advancingtransformerarchitecturelongcontext} selects elements randomly for dynamic adjustment, while our inter-patch token reorganization enables structured global-local interactions. BigBird Sparse Attention \cite{condevaux2022lsgattentionextrapolationpretrained} combines local attention with skipping mechanisms, but relies on predefined patterns unlike our dynamic delegate token communications. Bounded Sparse Attention \cite{a2025computationalefficientinformativenonignorable} restricts attention to local regions for short-range dependencies, whereas our method balances local and global contexts through patch concatenation. Fixed Sparse Attention \cite{chittyvenkata2023surveytechniquesoptimizingtransformer} uses predefined patterns lacking adaptability, while our dynamic reorganization adapts to input structure. Star-shaped Sparse Attention \cite{huang2024advancingtransformerarchitecturelongcontext} connects central nodes to peripheral nodes, focusing on specific regions, whereas our method distributes global context comprehensively via token shuffling. LogSparse Attention \cite{li2020enhancinglocalitybreakingmemory} adjusts weights using logarithmic functions, while our approach achieves smoothness through delegate tokens and residual connections enhancing gradient stability.

Structured State Space Sequence Models (S4)\cite{gu2022efficientlymodelinglongsequences, somvanshi2025surveystructuredstatespace} and their successors including RetNet~\cite{sun2023retentivenetworksuccessortransformer}, RWKV~\cite{peng2024eaglefinchrwkvmatrixvalued}, and Mamba~\cite{gu2024mambalineartimesequencemodeling} represent a class of sequence models that treat hidden tensors as state-spaces, with S4 employing structured linear dynamical systems through diagonal $\mathbf{A}$ matrices\cite{gu2022efficientlymodelinglongsequences} for efficient long-range dependency modeling, while Mamba introduces input-dependent parameter selection and RetNet incorporates gating mechanisms to enhance selective information propagation.The fundamental distinction between Monte Carlo Attention and State Space Models lies in their core information flow mechanisms. S4 employs sequential state propagation where information flows through linear recurrence relations $\mathbf{h}_t = \mathbf{A}\mathbf{h}_{t-1} + \mathbf{B}\mathbf{x}_t$, creating a Markovian dependency chain that limits each state to only directly access the previous state. In contrast, Monte Carlo Attention enables direct global information exchange through inter-patch delegation, allowing any patch to access information from any other patch through strategic token reorganization. S4 suffers from inherent limitations such as limited modeling flexibility due to linear time invariance, information bottlenecks caused by finite-dimensional state vectors, difficulties in capturing long-range dependencies through multi-step propagation, and inadequate expressiveness for modeling complex nonlinear relationships. While Mamba\cite{gu2024mambalineartimesequencemodeling} addresses some S4 limitations through input-dependent parameter selection, introducing selective state space mechanisms $\mathbf{h}_t = \mathbf{A}_t\mathbf{h}_{t-1} + \mathbf{B}_t\mathbf{x}_t$ where $\mathbf{A}_t, \mathbf{B}_t$ are dynamically computed, it still inherits fundamental approximation errors from low-rank representations and remains susceptible to numerical instabilities during state transition computations. Monte Carlo Attention transcends these limitations by maintaining full attention expressiveness through exact computations within augmented local-global contexts, avoiding both the linear time invariance constraints of S4 and the precision loss associated with low-rank approximations. This approach enables genuine global context propagation while preserving local precision, circumventing the information bottlenecks inherent in linear attention models and state space approximations that trade expressiveness for computational efficiency.

As one variant of S4 models, traditional linear attention models\cite{pmlr-v119-katharopoulos20a,peng2020random,qin2022cosformer} achieve computational efficiency through fundamental approximations that compromise information fidelity. These methods eliminate the softmax operation and map queries and keys to hidden representations through kernel functions, then compute attention through right products of keys and values to reduce complexity from $\mathcal{O}(L^2d)$ to $\mathcal{O}(Ld^2)$ where $N$ is the sequence length and $d$ is the matrix dimension. TransNormer further addresses unbounded gradient issues by replacing scaling with normalization operations\cite{qin2022devillineartransformer, qin2024lightningattention2freelunch}. However, these approaches suffer from critical limitations when compared to Monte Carlo Attention. Linear attention methods fundamentally compromise representation quality by mapping high-dimensional Q, K, V matrices to low-dimensional feature spaces through kernel functions, inevitably losing high-order statistical information and complex token interactions. In contrast, Monte Carlo Attention maintains full attention expressiveness without approximation, ensuring complete information preservation through exact self-attention computations within augmented local-global contexts. While linear attention achieves $\mathcal{O}(Ld^2)$ complexity, this efficiency comes at the cost of limited global context integration, as the low-rank approximations cannot adequately capture long-range dependencies when sequence length $L$ becomes extremely large. Monte Carlo Attention, however, enables near-unlimited global information flow through its inter-patch delegation mechanism, allowing effective context lengths to scale exponentially with layer depth while maintaining linear computational complexity. Furthermore, full attention computation circumvents the attention dilution problem\cite{qin2022devillineartransformer} that affects linear transformers by preserving local emphasis\cite{vaswani2017, qin2022cosformer} while facilitating comprehensive global context integration, ensuring that both short-range and long-range dependencies are adequately captured without the trivial distribution of attention scores over extended sequences. Finally, Monte Carlo Attention avoids the training instability issues commonly observed in linear attention models, particularly the unbounded gradient problems\cite{qin2022devillineartransformer} that TransNormer attempts to address, by maintaining well-conditioned attention computations within bounded local contexts.

Lastly, incorporating Temporal Convolutional Networks (TCNs)~\cite{bai2018empiricalevaluationgenericconvolutional} into the architecture enables capturing both instantaneous correlations and evolving temporal patterns without the need for positional encoding. TCNs are designed to handle sequences by stacking dilated convolutions, which allow the network to access long-range dependencies while maintaining causality. Due to the inherent nature of convolutions in TCNs, the model can naturally learn the relative positions of elements in the sequence, thus eliminating the necessity for explicit positional encodings.

As demonstrated in our results (Section \ref{results}), this structure achieves state-of-the-art performance while requiring significantly fewer computational resources compared to conventional approaches. Quantitative complexity analysis and detailed FLOPs comparisons are provided in Appendix \ref{Appendix:Computational Complexity Analysis}.

\subsection{Training}     \label{Training}                                          
This section details the training methods for BigBang-Proton, including both the pre-training and post-training phases.
  
\subsubsection{Pre-Training}

As shown in Table~\ref{tab:model_hyperparams}, BigBang-Proton model of $1.5$ billion parameters is configured as a 20-layer Monte Carlo Attention architecture with a hidden state dimension of 1024 and 4 attention heads per layer. It uses a vocabulary size of 259, encompassing 256 possible byte values and 3 special tokens for start, end, and padding. The feed-forward network employs an expansion ratio (\texttt{mlp\_ratio}) of 2 to increase channel capacity. To regularize training and improve generalization, the model incorporates dropout mechanisms with a maximum rate of 0.15 applied across various layers, an attention-specific dropout rate of 0.1, and a stochastic depth survival rate of 0.15. This setup enables the model to effectively process and generate sequences in the context of scientific data, leveraging the architecture's scalability for diverse modalities.

\begin{table}[htbp]
\centering
\begin{minipage}[t]{0.45\textwidth}
\centering
\caption{Model Hyperparameters}
\label{tab:model_hyperparams}
\begin{tabular}{|l|l|}
\hline
\textbf{Hyperparameter} & \textbf{Value} \\
\hline
embedding\_dim & 1024 \\
\hline
num\_classes & 259 \\
\hline
num\_layers & 20 \\
\hline
num\_heads & 4 \\
\hline
mlp\_ratio & 2 \\
\hline
dropout\_rate & 0.15 \\
\hline
attention\_dropout & 0.1 \\
\hline
stochastic\_depth\_rate & 0.15 \\
\hline
\end{tabular}
\end{minipage}
\hfill
\begin{minipage}[t]{0.45\textwidth}
\centering
\caption{Pretraining Hyperparameters}
\label{tab:pretraining_hyperparams}
\begin{tabular}{|l|l|}
\hline
\textbf{Hyperparameter} & \textbf{Value} \\
\hline
patch\_size & (16,32,1024) \\
\hline
max\_input\_size & $8192 \times 8$ \\
\hline
min\_input\_size & $8192 \times 8$ \\
\hline
vocab\_size & 259 \\
\hline
epochs & 20 \\
\hline
lr & 0.00004 \\
\hline
warmup & 100 \\
\hline
batch\_size & 1 \\
\hline
grad\_accum & 4 \\
\hline
optimizer & AdamW \\
\hline
weight\_decay & $3 \times 10^{-3}$ \\
\hline
workers & 8 \\
\hline
limit\_tokens & $8192 \times 8$ \\
\hline
iter\_val & 70 \\
\hline
iter\_print & 10 \\
\hline
iter\_save & 280 \\
\hline
iter\_break\_val & 300 \\
\hline
\end{tabular}
\end{minipage}
\end{table}

We utilized distributed framework Fully Sharded Data Parallel (FSDP) to implement the distributed pretraining. As demonstrated in Table~\ref{tab:pretraining_hyperparams}, the pretraining process is configured with the following settings. The model uses patch dimensions of (16,32,1024) for its Monte Carlo Attention layers, where patches are randomly divided from 16 to 32 tokens, with the final two Monte Carlo Attention layers using a larger patch size of 1024 for global context integration. The input sequence length is fixed at $8192 \times 8$ tokens, as both the maximum and minimum input sizes are set to this value. The vocabulary size is 259, comprising 256 values for byte-level data (0-255) and three special tokens for start, end, and padding. The training is set for 20 epochs, although the process is primarily managed by iteration count. A learning rate of 0.00004 is used for pretraining, with a warmup period of 100 steps to stabilize initial convergence. For fine-tuning, the learning rate is reduced to 0.00001. To optimize resource usage and prevent overfitting, an early stopping strategy is implemented. The batch size is set to 1, where multiple sequences are concatenated into a single long sequence, and gradient accumulation is applied over 4 steps to effectively simulate a larger batch size. The AdamW optimizer is used for parameter updates, with a weight decay of $3 \times 10^{-3}$ for regularization. Data loading is handled by 8 worker processes to ensure efficient I/O. A token limit of $8192 \times 8$ is enforced per batch. The training loop is monitored with evaluation performed every 70 steps, progress printed every 10 steps, and model checkpoints saved every 280 steps. To prevent excessively long validation runs, the validation process is capped at 300 steps.

\begin{figure}[H]
    \centering
    \begin{subfigure}[t]{0.48\textwidth}
        \centering
        \includegraphics[width=\textwidth, height=6cm, keepaspectratio]{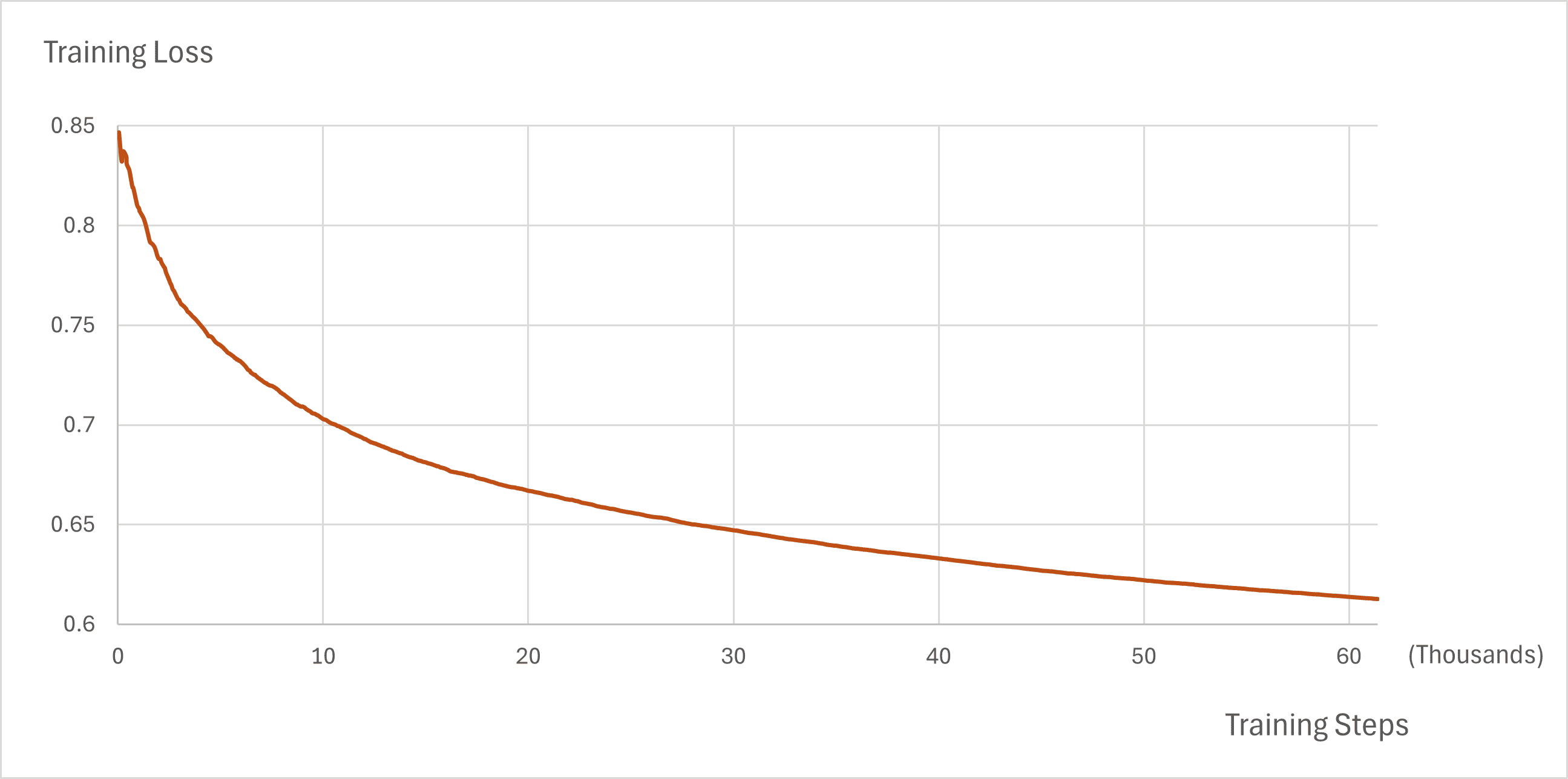}
        \caption{Training loss steadily decreases to 0.613 over 61,381 steps during pretraining on a mixed corpus of 9 distinct tasks. Despite diverse and highly heterogeneous data patterns, next-word prediction yields smooth and stable convergence.}
        \label{fig:training loss}
    \end{subfigure}
    \hfill
    \begin{subfigure}[t]{0.48\textwidth}
        \centering
        \includegraphics[width=\textwidth, height=6cm, keepaspectratio]{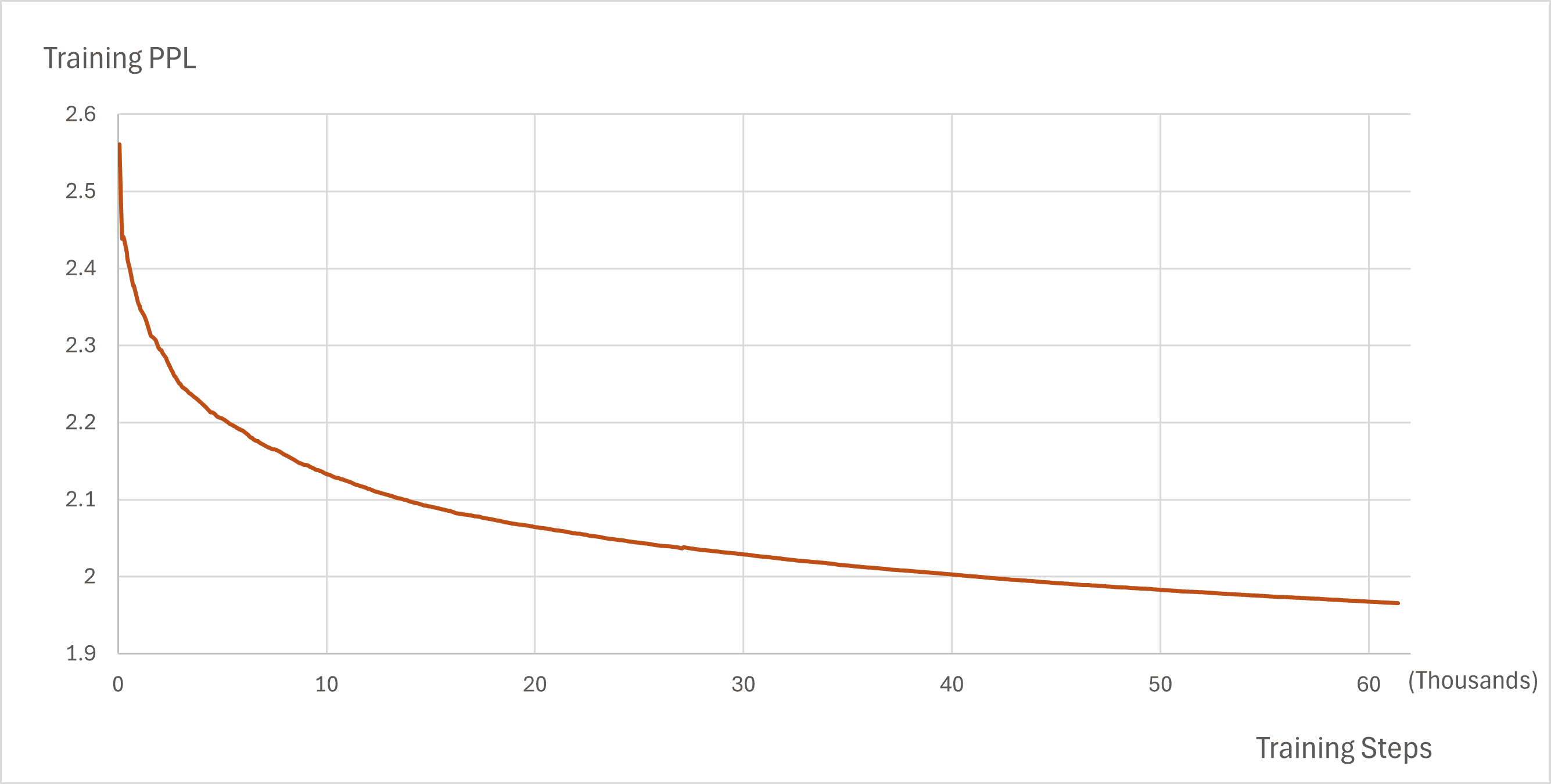}
        \caption{Training perplexity decreases monotonically to ~2.04, demonstrating stable convergence despite highly heterogeneous data structures.}
        \label{fig:training ppl}
    \end{subfigure}
    \caption{ Training loss and perplexity during pre-training on a heterogeneous corpus of 9 diverse datasets}
    \label{fig:training performance}
\end{figure}

As shown in Figure~\ref{fig:training performance}, The training loss and perplexity curves exhibit consistent, smooth, and monotonic convergence over 61,381 steps, demonstrating stable and effective learning throughout pre-training. The loss decreases steadily to 0.613, while perplexity drops to 2.04, reflecting a substantial improvement in the model’s ability to predict the next token accurately across all nine diverse tasks. This consistent improvement demonstrates that next-word prediction implemented with Binary Patch Encoding, overcomes the high data heterogeneity and enables robust model convergence effectively.

\subsubsection{Post-Training} \label{post training}

In the last phase of pretraining, we use text book level high quality data~\cite{gunasekar2023textbooksneed} for annealing, then use Infinity-Instruct dataset~\cite{BAAI_2024_Infinity-Instruct} for supervised instruction finetuning. After that, for scientific downstream tasks, the pretrained model is further finetuned on task datasets to perform language-guided scientific computing. 

\section{Results} \label{results}
\subsection{Language Generation}    

\begin{table}[H]
\centering
\caption{Performance of BigBang-Proton Byte-Level Model Across Seven Leading Benchmarks}
\label{tab:benchmarks}
\begin{tabular}{lc}
\toprule
\textbf{Benchmark} & \textbf{Accuracy (\%)} \\
\midrule
MMLU  & 25.44 \\
Arc-E & 25.29 \\
Arc-C & 26.19 \\
HellaSwag & 25.19 \\
PIQA & 51.14 \\
\bottomrule
\end{tabular}
\end{table}

The BigBang-Proton of 1.5 billion parameters, same model size as GPT-2\cite{gpt2}, has been tested across five widely recognized benchmarks to evaluate its performance in various aspects of language understanding and generation.
Building on the high-quality datasets detailed in Section \ref{post training}, we finetuned our model using instruct data designed to enhance its conversational capabilities. This process involved adapting datasets such as Infinity-Instruct~\cite{BAAI_2024_Infinity-Instruct} and Open Phi~\cite{gunasekar2023textbooksneed, openphi} into instruct format.

As shown in Table~\ref{tab:benchmarks}, trained on a general-purpose dataset comprising approximately 60B BPE tokens sourced from one-tenth of the SlimPajama\cite{cerebras2023slimpajama} dataset's 10 chunks, The performance of BigBang-Proton on the MMLU\cite{mmlu}, ARC, HellaSwag, and PIQA benchmarks, with scores of 25.44\%, 25.29\%, 26.19\%, 25.19\% and 51.14\% respectively, is significant. These results demonstrate that BigBang-Proton, which uses a Binary Patch Encoding method, achieves reasoning capabilities on par with GPT-2 which was trained on approximately 40B BPE tokens from WebText and obtained 25.83\% on MMLU\cite{mmlu}. This finding is particularly noteworthy as it indicates that the Binary Patch Encoding strategy is not only effective for processing scientific and numerical data but also holds substantial potential for general language reasoning tasks. This result suggests that the unique encoding method does not compromise its capacity for complex reasoning. This language ability is a critical prerequisite for the model's core function of language-guided scientific computing. 

\begin{figure}[H]
    \centering
    \includegraphics[width=1\textwidth]{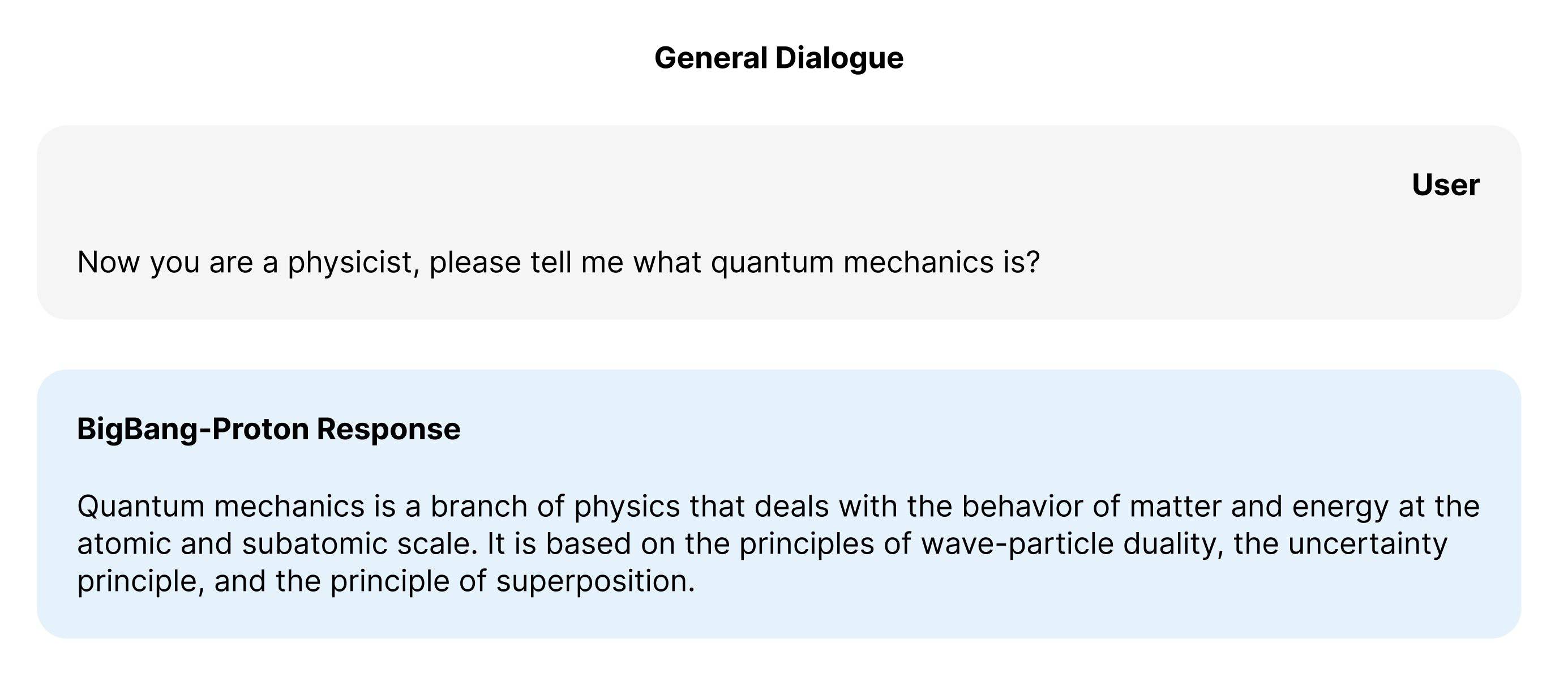} 
    \caption{Illustration of a conversation demonstrating the model's general language generation capabilities. The figure shows a dialogue example where the model responds to user prompt in a coherent manner.}
    \label{fig:chat9_dialogue}
\end{figure}
We present an example of general language generation capabilities in Figure~\ref{fig:chat9_dialogue}. This figure illustrates a conversation between a user and the model, highlighting BigBang-Proton have learned basic physics knowledge through training on general text.

\subsection{Arithmetic Operations}
\label{subsec:arithmetic}
It is widely acknowledged that LLMs encounter significant difficulties when performing intricate numerical computations. GPT-4 only achieve to 59\% accuracy in 3-digit by 3-digit multiplication by \cite{dziri2023faithfatelimitstransformers, yang2023gptsolvemathematicalproblems, yuan2023largelanguagemodelsperform}, performance further declining sharply for larger numbers \cite{li2024humanswritecodelarge,dziri2023faithfatelimitstransformers}. A systematic study on LLMs capabilities on arithmetic tasks by \citeay{yuan2023largelanguagemodelsperform} shows that GPT-4 fails in all cases in multiplications of two integers up to $100,000$. This fundamental flaw can be traced to factor that mainstream Byte Pair Encoding(BPE) tokenization doesn't work well for number representation\cite{lee2024digitstodecisions, beren2023integer}. BPE tends to split multi-digit numbers into irregular chunks (e.g., ``1009 + 8432'' might be segmented as ``100'', ``9'', ``+'', ``84'', ``32''), which can disrupt the model's ability to correctly align operands and perform arithmetic operations accurately \cite{10.1371/journal.pone.0310409}.  
    
In response to these limitations, recent research has focused on developing strategies to augment the arithmetic capabilities of LLMs. Early explorations by Nogueira et al. \cite{nogueira2021investigating} and Wang et al. \cite{wang2021exploring} laid the groundwork by assessing the ability of LLMs to execute basic arithmetic operations such as addition and subtraction. Subsequently, Muffo et al. \cite{muffo2023evaluating} conducted a focused evaluation on the performance of language models in two-digit multiplication, providing valuable insights into their proficiency in this specific domain. The BIG-bench dataset \cite{srivastava2022beyond}, which encompasses a diverse array of arithmetic tasks involving numbers up to five digits, has emerged as a pivotal resource for benchmarking the capabilities of these models.
 Nye et al. \cite{nye2021show} implemented scratchpad-based fine-tuning methods, which significantly enhanced the models' ability to perform eight-digit addition. Goat \cite{liu2023goat} employed supervised instruction fine-tuning to bolster the performance of LLMs on elementary arithmetic operations with large integers, covering addition, subtraction, multiplication, and division. Jelassi et al. \cite{jelassi2023length} explored the concept of length generalization in basic arithmetic tasks, utilizing methods such as relative position embeddings and training set priming to improve model adaptability.
Building on these advancements, Zhen et al. \cite{yang2023gptsolvemathematicalproblems} proposed MathGLM, which achieved an accuracy of 93.09\% in 5-12 digits arithmetic tasks without relying on external calculators. This model demonstrated superior performance compared to GPT-4 and ChatGPT, particularly in complex arithmetic operations involving large numbers, decimals, and fractions. The success of MathGLM can be attributed to its step-by-step strategy and curriculum learning approach, which enabled the model to learn the underlying rules of arithmetic operations effectively. 

In another significant contribution, McLeish et al. \cite{mcleish2024transformersarithmeticrightembeddings} addressed the positional encoding limitations of transformers by introducing Abacus Embeddings. Using character-level tokenizer, Abacus Embeddings encodes the position of each digit relative to the start of the number, enabling transformers to align the exact position of digits in the arithmetic operations. This innovation led to state-of-the-art performance in arithmetic tasks, with models achieving up to 99\% accuracy on 100-digit addition problems. The study also demonstrated the potential of these embeddings in other algorithmic reasoning tasks, such as multiplication and sorting, further expanding the scope of LLMs' numerical capabilities. While Abacus Embeddings have shown remarkable improvements in handling complex numerical computations, they introduce additional layers of complexity into the model architecture which is hard to adapt to current mainstream architectures for general-purpose pretraining. 

In contrast, BigBang-Proton employs Binary Patch Encoding, which differs fundamentally from BPE tokenization, and also from character-level position alignment of Abacus Embedding. We employ Binary Patch to preserve the integrity of number representation, and incorporate chain-of-thought (COT) reasoning during pretraining, to let the model learn positional alignment directly without bringing in extra modifications to model architectures.

\begin{figure}[tbp]
    \centering
    \begin{subfigure}[b]{1\textwidth}
        \includegraphics[width=\textwidth]{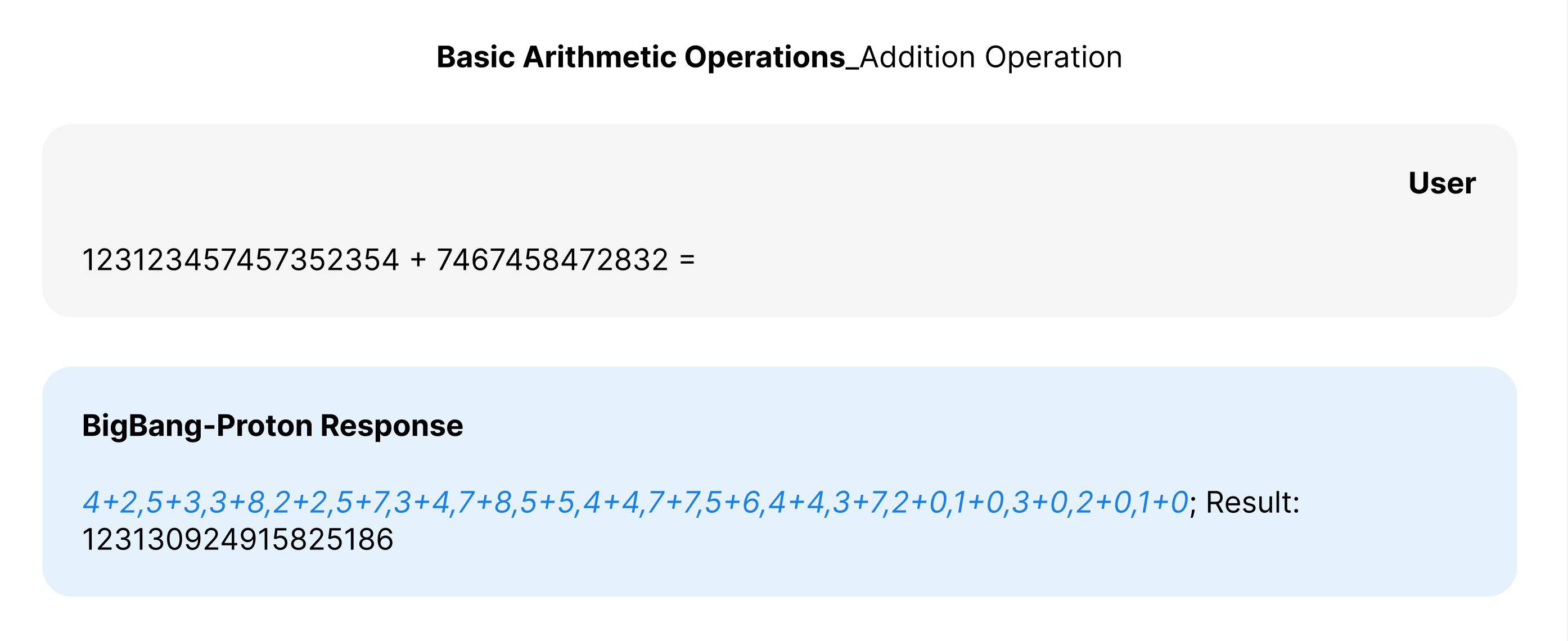}
        \caption{Addition example}
        \label{fig:chain_of_thought_addition}
    \end{subfigure}
    \hfill
    \begin{subfigure}[b]{1\textwidth}
        \includegraphics[width=\textwidth]{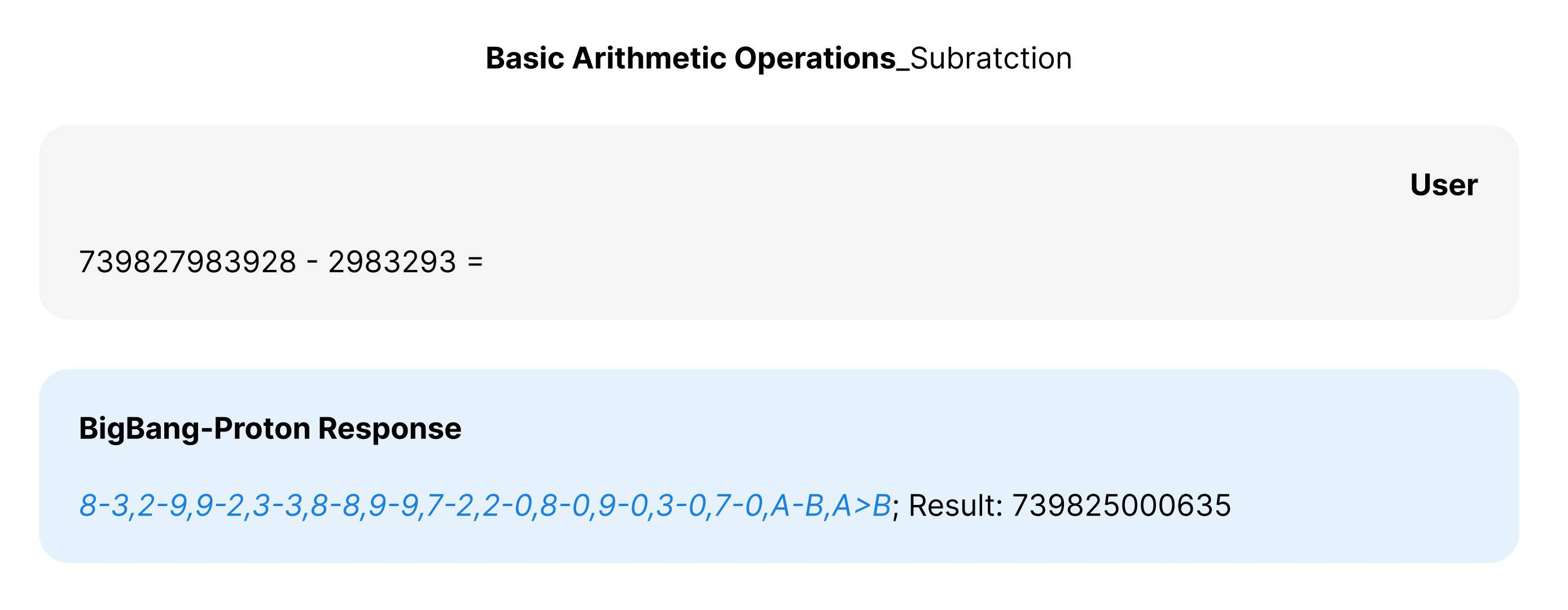}
        \caption{Subtraction example}
        \label{fig:chain_of_thought_subtraction}
    \end{subfigure}
    \hfill
    \begin{subfigure}[b]{1\textwidth}
        \includegraphics[width=\textwidth]{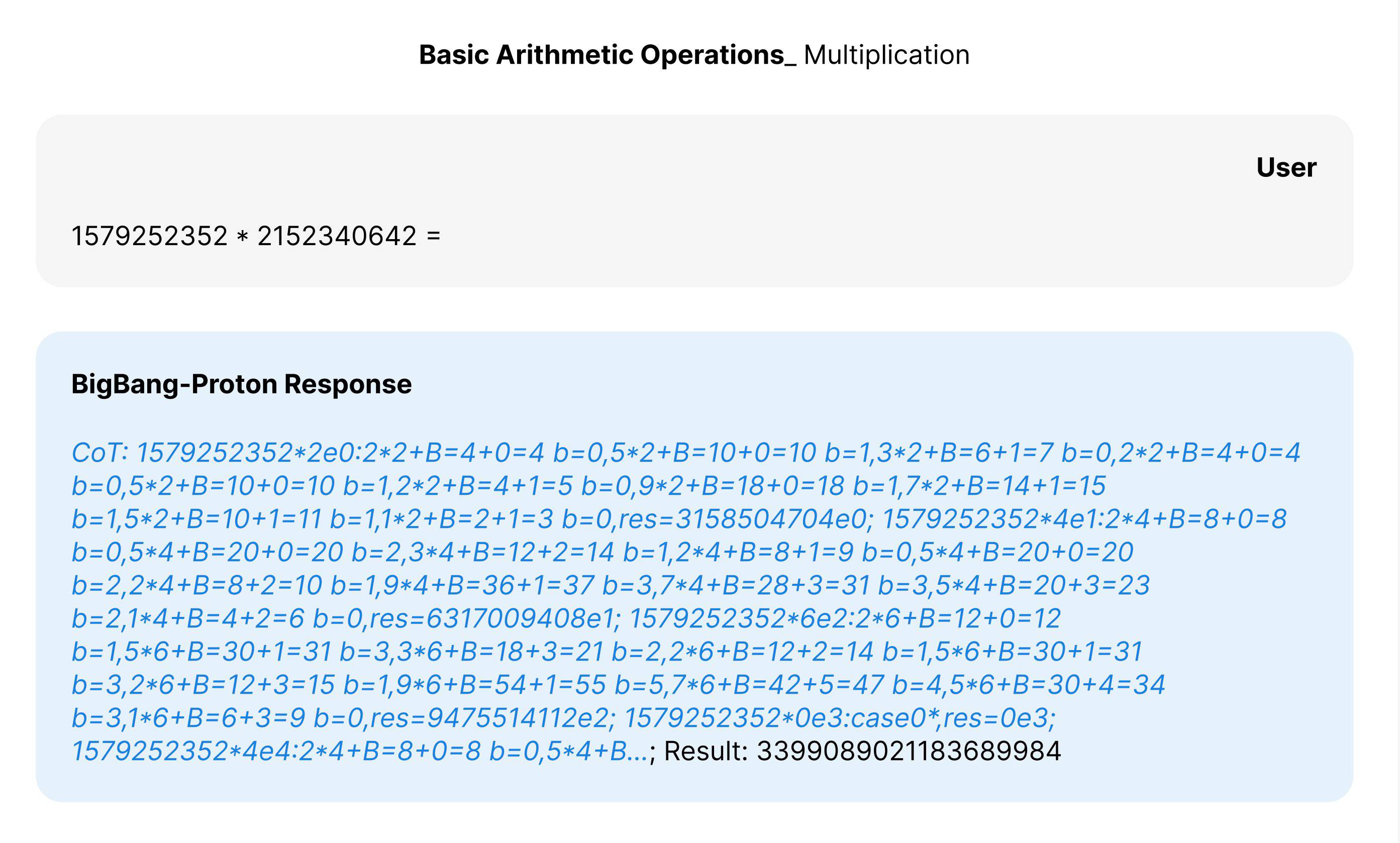}
        \caption{Multiplication example}
        \label{fig:chain_of_thought_multiplication}
    \end{subfigure}
    \caption{Illustration of prompt-answer examples for arithmetic operations using chain-of-thought reasoning. This design enables the model to learn the carry principle of ALU through position-aligned step-by-step computation, where the model understands the sequential propagation of carry bits during multi-digit arithmetic operations. }
    \label{fig:chain_of_thought_example}
\end{figure}

In this task we use the arithmetic COT datasets in pretraining corpus again for finetuning, including addition, subtraction of up to 50-digit numbers, and multiplication of up to 12-digit numbers. As illustrated in Figure~\ref{fig:chain_of_thought_example}, we construct an arithmetic CoT dataset that simulates the step-by-step computation processes used by human to solve arithmetic problems. For addition and subtraction, the CoT breaks down the problem into a sequence of position-aligned-double-digit operations (e.g., `4+2', `5+3'), and for subtraction, the inclusion of a comparison step (`A>B') is critical as it teaches the model the concept of borrowing in subtraction, transforming the task from a single complex problem into a series of simple, position-wise operations that the model can learn sequentially. The multiplication CoT is significantly more sophisticated, with a specific focus on the carry mechanism. The model is shown to take one digit from the first number A and multiply it by the entire second number B, with the notation `b=0', `b=1', `b=2' explicitly representing the carry-over value from one digit multiplication to the next. For example, in `5*2+B=10+0=10 b=1', the model learns that the result is 10, so the digit `0' is written down and a carry of `1' is passed to the next higher digit, while the notation `res=3158504704e0' and `res=6317009408e1' explicitly encodes the position value of the partial product, where `e0', `e1', `e2' represent multiplying by $10^0$, $10^1$, $10^2$, etc., which is essential for aligning the partial products of different positions correctly before their final summation. The CoT implicitly shows that the final answer is obtained by summing all the partial products (e.g., `res=...e0 + res=...e1 + res=...e2 + ...').

\begin{figure}[tbp]
\centering
\includegraphics[width=1\linewidth]{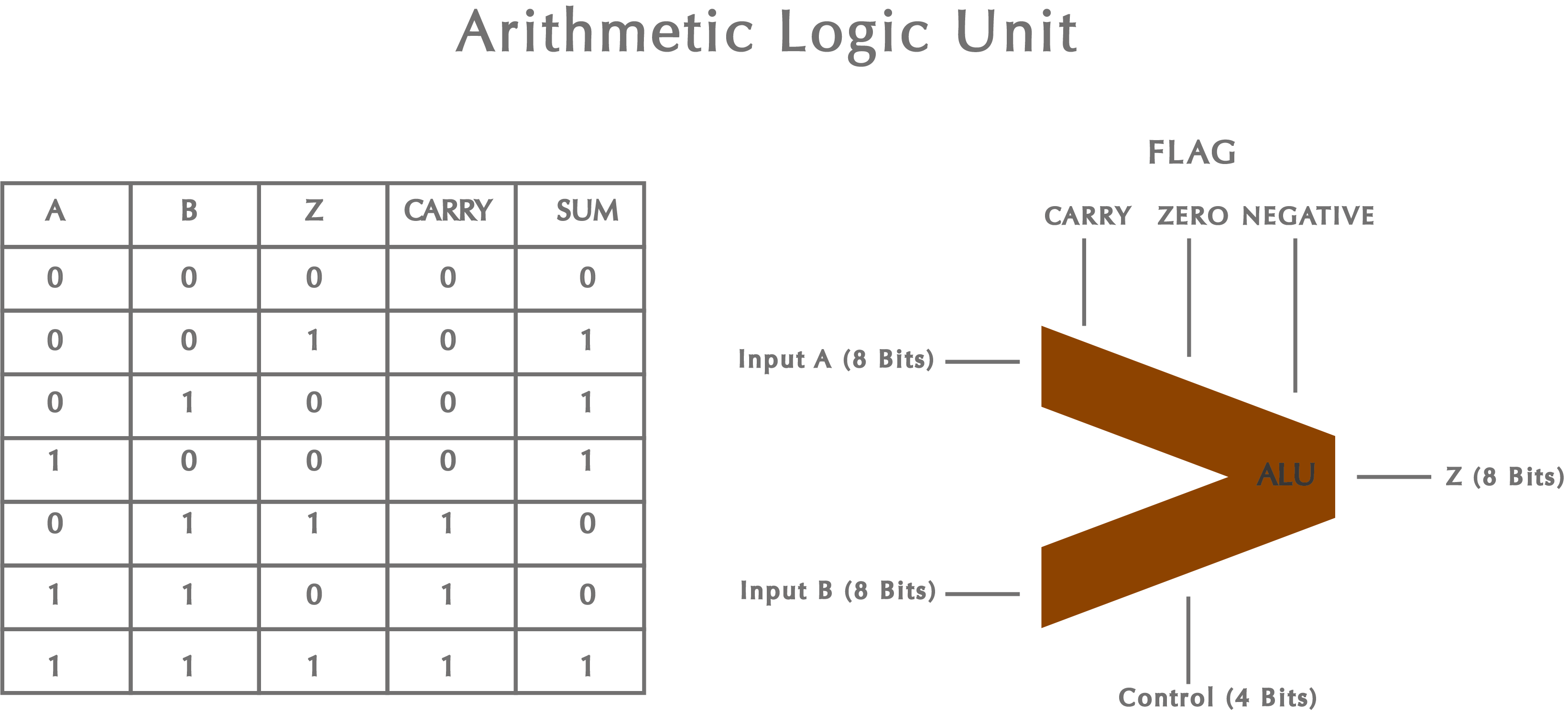}
\caption{Arithmetic Logic Unit adder carry principle. It is natural and indispensable for Binary Patch Encoding to learn the bit carry principle, since both utilize binary representation.}
\label{fig:alu}
\end{figure}

The carry mechanism is the core operational principle of arithmetic logic units (ALU)~\cite{Kaur2013VHDLDA}, the building block of micro-processors. Built upon Binary Patch Encoding and trained on COT step-by-step reasoning, BigBang-Proton learns the operational principles of ALU, including the carry mechanism, which is essentially the arithmetic logic unit's bit-wise summation principle. ALU performs bit-parallel operations on binary inputs, as shown in Figure~\ref{fig:alu}. A typical ALU comprises input registers, functional units, and an output multiplexer. Input registers store operands $A$ and $B$, with a 4-bit width shown in the example. The functional units include an adder/subtractor with a carry chain, logic gates (AND/OR/XOR), and shift/rotate circuits. These components enable the ALU to perform various arithmetic and logical operations. Finally, the output multiplexer selects the result based on control signals, determining the final output of the ALU. The carry principle in ALU refers to the process where, during bit-wise addition, an overflow (carry bit) from one bit position is propagated to the next higher bit position, ensuring accurate summation across binary digits by handling dependencies between bits in a chain-like manner.  

One example in multiplication tasks encountered in testing our model is as follow, which demonstrates BigBang-Proton masters the position-wise carry principle. The input sequence is: $$
\begin{aligned}
147137375e0 + 21019625e1 + 105098125e2 + 21019625e3 + 0e4 \\
+ 63058875e5 + 21019625e6 + 105098125e7 + 84078500e8
\end{aligned}
$$
The expected result is:
$$
1078338649271125 + 84078500e8 = 9486188649271125
$$
However, if the input contains an error, such as replacing \texttt{0e8} with \texttt{5e8}, the model's output changes accordingly:
$$
1078338649271125 + 84078505e8 = 9486189149271125
$$

More example highlighting the model's sensitivity to the position of digits can be seen in Appendix~\ref{append:error_case_arithmetic}. BigBang-Proton learns to manage carry propagation and bit-wise summation, and is capable to generalize to arithmetic tasks with any number of digits, because calculators and ALUs are fundamentally designed around this principle for unbounded scalability. 

This capability represents the core of language-guided scientific computing and financial analysis, especially when dealing with large-scale experimental datasets from Big Science experiments, which often include vast numerical experimental results. Basic operations like addition, subtraction, and multiplication form the foundation of numerical analysis, where computation processes and results require 100\% accuracy to accurately map to the described physical structures and physical phenomena. By enabling LLMs to learn the ALU carry principle, 100\% accuracy in arithmetic operation and absolute elimination of hallucination is achievable. This is precisely why we position arithmetic capabilities as the cornerstone on building our scientific multi-task learner.

\begin{figure}[H]
    \centering
    \includegraphics[width=1\textwidth]{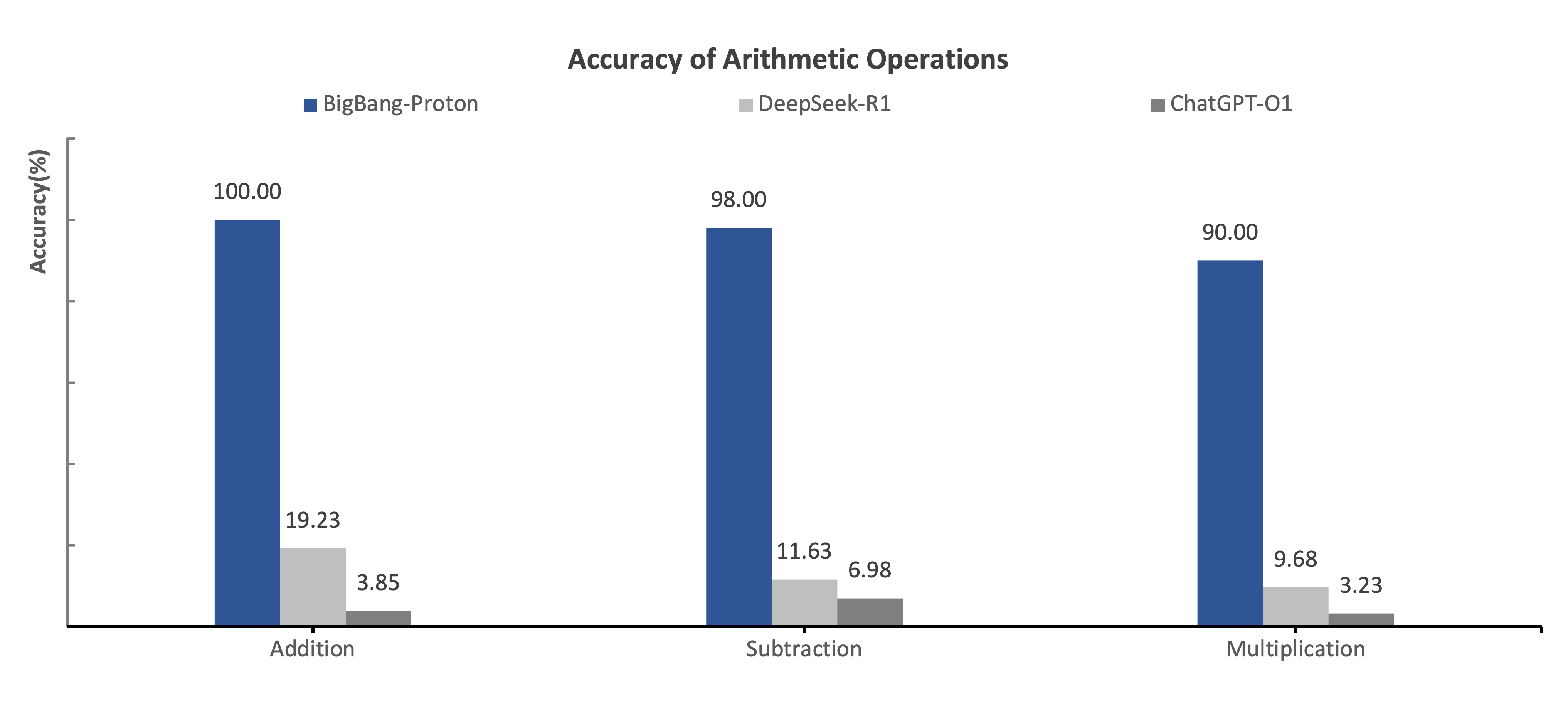} 
    \caption{Comparison of arithmetic performance among different models on specific tasks. The bar chart shows the accuracy of BigBang-Proton, DeepSeek R1 and ChatGPT o1 in handling addition, subtraction, and multiplication tasks. For each task, BigBang-Proton were evaluated on 200 test samples, DeepSeek R1 and ChatGPT-o1 were evaluated on 100 test samples, with all model assessed using a zero-shot prompting setup.} 

    \label{fig:arithmetic_performance_comparison}
\end{figure}

As demonstrated in Figure~\ref{fig:arithmetic_performance_comparison}, in arithmetic tasks involving up to 50-digit addition and subtraction, and up to 12-digit multiplication, the accuracy performance of the three models, BigBang-Proton, DeepSeek-R1, and ChatGPT-o1, is as follows: BigBang-Proton achieves accuracies of 100\%, 98\%, and 90\% on addition, subtraction, and multiplication, respectively.  DeepSeek-R1 achieves 19.23\%, 11.63\%, and 9.68\%; and ChatGPT-o1 achieves the lowest accuracies at 3.85\%, 6.98\%, and 3.23\%. For each task, BigBang-Proton was evaluated on a test set of 200 samples, while DeepSeek-R1 and ChatGPT-o1 were evaluated on a test set of 100 samples, with all models assessed under a zero-shot prompting setup. The dramatic performance gap between BigBang-Proton and the general-purpose LLMs on arithmetic tasks stems from a fundamental difference in their data encoding approaches. 

BigBang-Proton's superior performance, achieving 100\% accuracy on 50-digit addition and 98\% on 50-digit subtraction, results from our Binary Patch Encoding method and CoT training. This approach treats all data as raw binary sequences, preserving numerical information integrity. Crucially, it allows the model to learn digit positional alignment, a core a ALU principle, from training data including step-by-step reasoning. By processing numbers as unbroken character sequences (e.g., "12345"), the model naturally learns that the rightmost digit is units, next is tens, and so on, enabling a true, generalizable algorithm for addition and subtraction extrapolating to any length. Since BigBang-Proton achieves 90\% accuracy in 12-digit multiplication as shown in Figure~\ref{fig:arithmetic_performance_comparison}, errors related to digit position contribute most incorrect cases, indicating its sensitivity to digit position, as seen in Appendix~\ref{append:error_case_arithmetic}. In stark contrast, DeepSeek-R1 and ChatGPT-o1 use Byte Pair Encoding (BPE) which fragments numbers into arbitrary chunks (e.g., "12345" might split into "12", "34", "5"). This destroys positional understanding and forces memorization rather than algorithmic reasoning. Unable to learn coherent algorithms due to inconsistent tokenization, these models can only memorize specific input-output pairs in latent space, explaining their poor performance (19.23\% and 3.85\% on addition). BPE assigns arbitrary, non-sequential token IDs to single digits, creating discontinuous representations. Step-by-step computation data is ineffective for BPE-based models because their tokenizer corrupts the problem structure. Essentially, while DeepSeek-R1 and ChatGPT-o1 memorize numbers facts, they fail to learn ALU operational principles.
 
\subsection{Jet Tagging Task in Particle Physics: Language-guided Classification}        

In high-energy particle collisions at facilities like the LHC or future electron-positron Higgs factories, quarks and gluons produce collimated sprays of hadrons traveling in the same directions, known as jets that preserve information about underlying parton dynamics and serve as crucial signatures for studying QCD. To address specific challenges in particle physics, following our previous work \texttt{BigBang-Neutron}\cite{wu2024scalingparticlecollisiondata}, we pretrained BigBang-Proton mixing the Jet Origin Identification (JoI) dataset\cite{Zhu2023ParticleNetAI, wu2024scalingparticlecollisiondata} with other multi-discipline data, then fine-tuned the pretrained model with the same dataset again when it comes to downstream task of jet tagging. This task focuses on distinguishing jets originating from different quarks and gluons. Specifically, the model is trained to identify 11 distinct types of jets,  which are critical for understanding the dynamics of high-energy collisions. The training dataset consists of 11 million samples, with 1000,000 samples per jet type, ensuring a balanced and comprehensive representation of each category. The ability to accurately identify the origin of jets is critical for advancing research in high-energy physics, particularly in studies of Higgs, W, and Z bosons, where majority of these bosons decay directly into two jets. In the finetuning of Jet Origin Identification (JoI) task, we trained the model for 10 epochs and evaluated its performance on a validation set consisting of 10,000 samples. Unlike our previous work \texttt{BigBang-Neutron}\cite{wu2024scalingparticlecollisiondata} that use a direct classification LM head as, BigBang-Proton leverages auto-regressive language modeling to predict the jet type based on zero-shot prompt of particle information. Jet tagging by BigBang-Proton constitutes a clear demonstration of language-guided classification.  

\begin{figure}[H]
		\centering
		\includegraphics[width=1\textwidth]{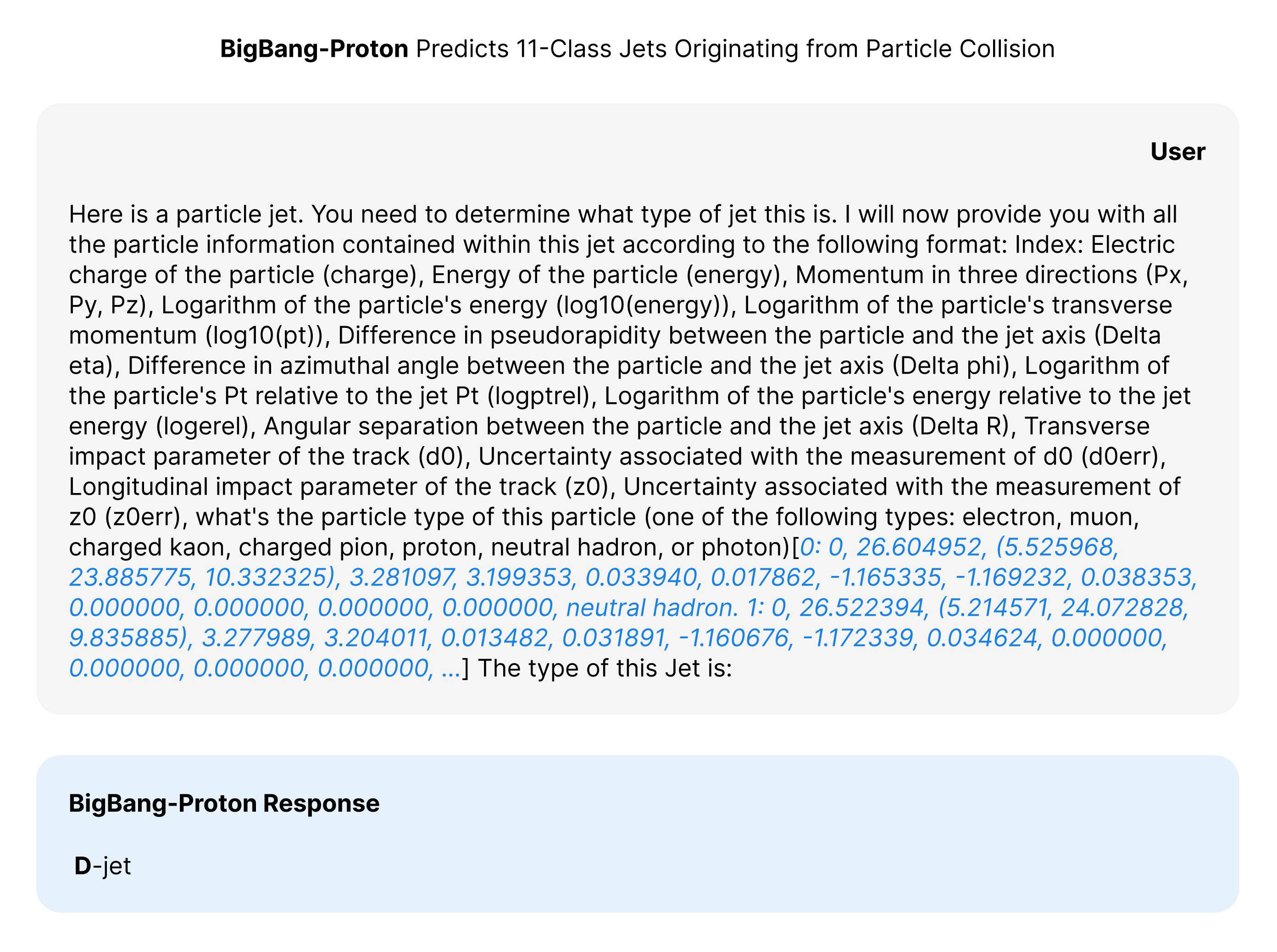} 
		\caption{Illustration of prompt-answer examples for jet origin identification (JOI) using BigBang-Proton. Each prompt begins with an instruction directive for the model, explicitly asking it to determine the jet category based on given particle data from simulation or experimental measurements. These numerical datasets have corresponding textual explanations in the prompts, establishing a foundation for the Theory-Experiment Learning paradigm. }
		\label{fig:chat5_jet}
\end{figure}

The dataset was meticulously designed in a prompt-class format to cultivate capabilities of language-guided classification from pretraining, following Theory-Experiment learning paradigm. As illustrated in Figure \ref{fig:chat5_jet}, The prompt for the jet tagging task is structured as a complete, self-contained instruction-response pair that seamlessly integrates natural language theory with experimental data. It begins with a clear instruction: "Here is a particle jet. You need to determine what type of jet it is." This directive sets the context and defines the model's objective. Immediately following this command, the prompt provides a comprehensive theoretical explanation in natural language, detailing the physical meaning of each particle property, such as charge, momentum, energy, and impact parameters, offering the model essential contextual knowledge for interpreting the subsequent numerical data. This theoretical foundation is then directly linked to the experimental input, which consists of a sequence of particle information. Each particle is described by its index and a set of numerical values representing its properties. The sequence concludes with the correct jet class (e.g., "c jet") explicitly stated as the expected answer. For detailed information about the JoI dataset, including specific attribute definitions and sample prompts, please refer to Appendix \ref{appendix_particle}. 
Our previous work BigBang-Neutron~\cite{wu2024scalingparticlecollisiondata} relies on adding a specialized classification or regression LM head on top of its base architecture, making the task capabilities an architectural choice. BigBang-Proton, however, leverages language-guided scientific computing through a prompt-prediction paradigm. This allows it to perform both classification and regression tasks in an end-to-end auto-regressive manner, where the task itself (e.g., "classify the jet type" or "predict the energy value") is specified within the text prompt during inference. This eliminates the need for any architectural modifications, enabling a unified model to handle diverse tasks through natural language instructions.

\begin{figure}[H]
        \centering
        \includegraphics[width=.4\textwidth]{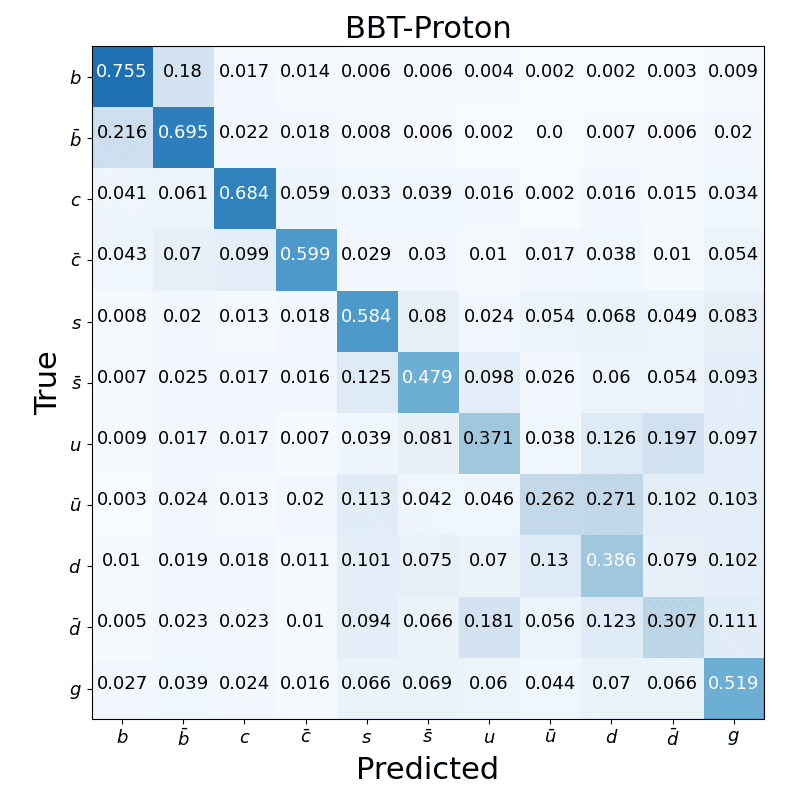}
        \hfill
        \includegraphics[width=.4\textwidth]{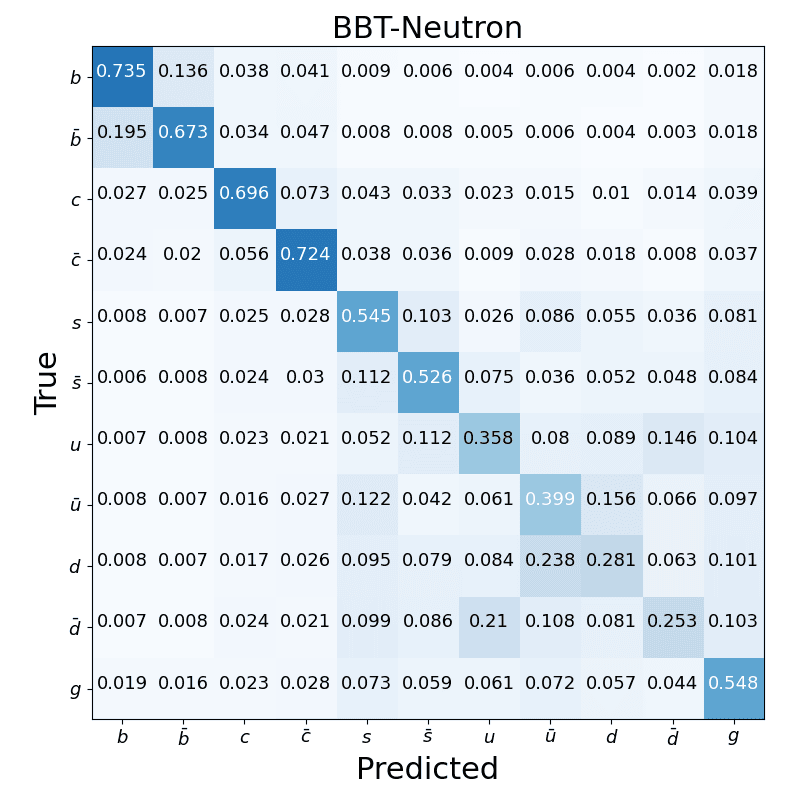}
        \hfill
        \includegraphics[width=.4\textwidth]{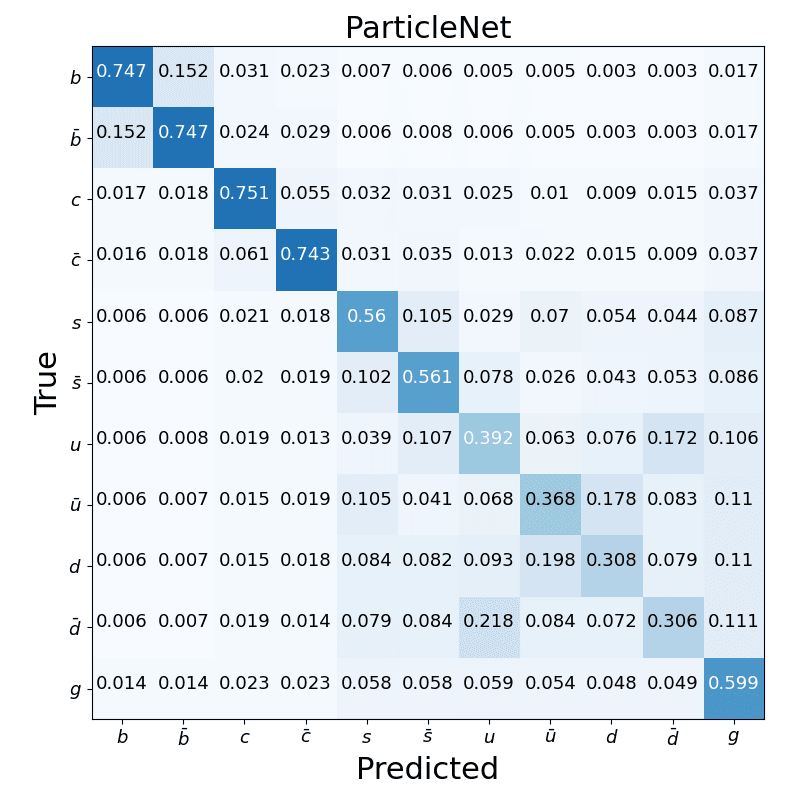}
        \hfill
        \includegraphics[width=.4\textwidth]{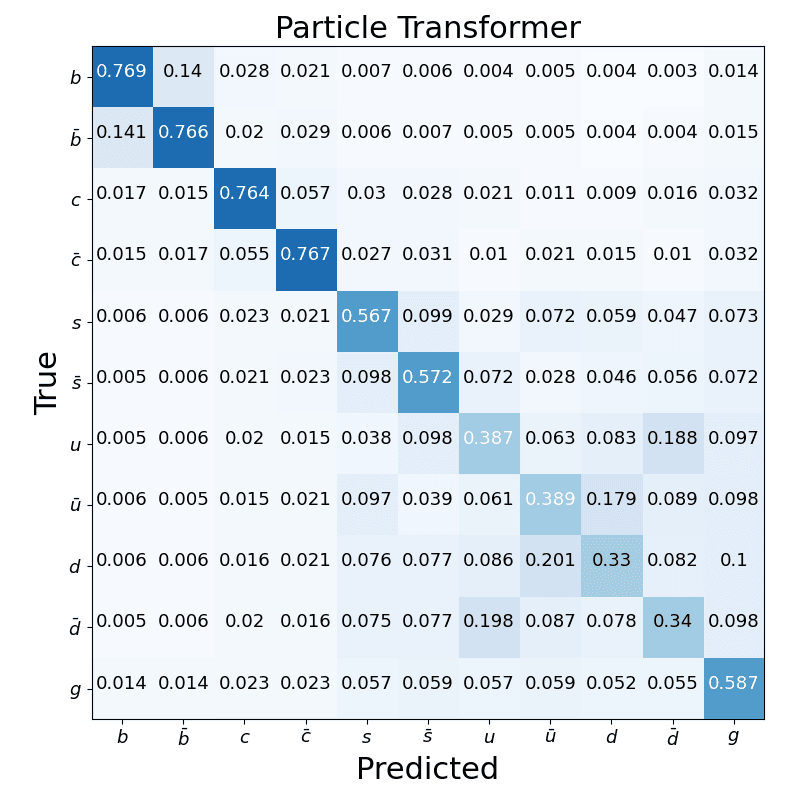}
        \caption{Confusion matrices ($M_{11}$) for $\nu\bar{\nu}H, H\to jj$ events at 240 GeV center-of-mass energy, generated by BigBang-Proton, BigBang-Neutron\cite{wu2024scalingparticlecollisiondata}, ParticleNet\cite{Zhu2023ParticleNetAI}, and Particle Transformer\cite{Qu:2022mxj}, using 1 million events per jet category (60\% training, 20\% validation, 20\% testing). Each matrix is normalized to unity per truth label row.}
    \label{fig:M11}
\end{figure}

\begin{figure}[H]
    \centering
    \includegraphics[width=1\textwidth]{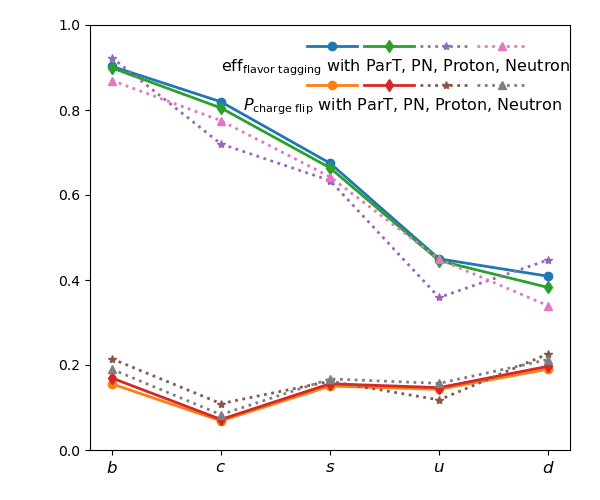}
    \caption{Jet flavor tagging efficiencies and charge flip rates for each quark species with BigBang-Proton, BigBang-Neutron, ParticleNet, and Particle Transformer. BigBang-Proton demonstrates the same level of performance as other models in tagging five flavors of quarks, and specifically outperforms other models in b and d quark tagging. The charge flip rate measures how often the model makes incorrect judgments regarding the charge of quarks. BigBang-Proton is closely aligned with other models, fluctuating around 20\%.}
\label{fig:effFlip}
\end{figure}

\begin{figure}[H]
    \centering
    \includegraphics[width=1\textwidth]{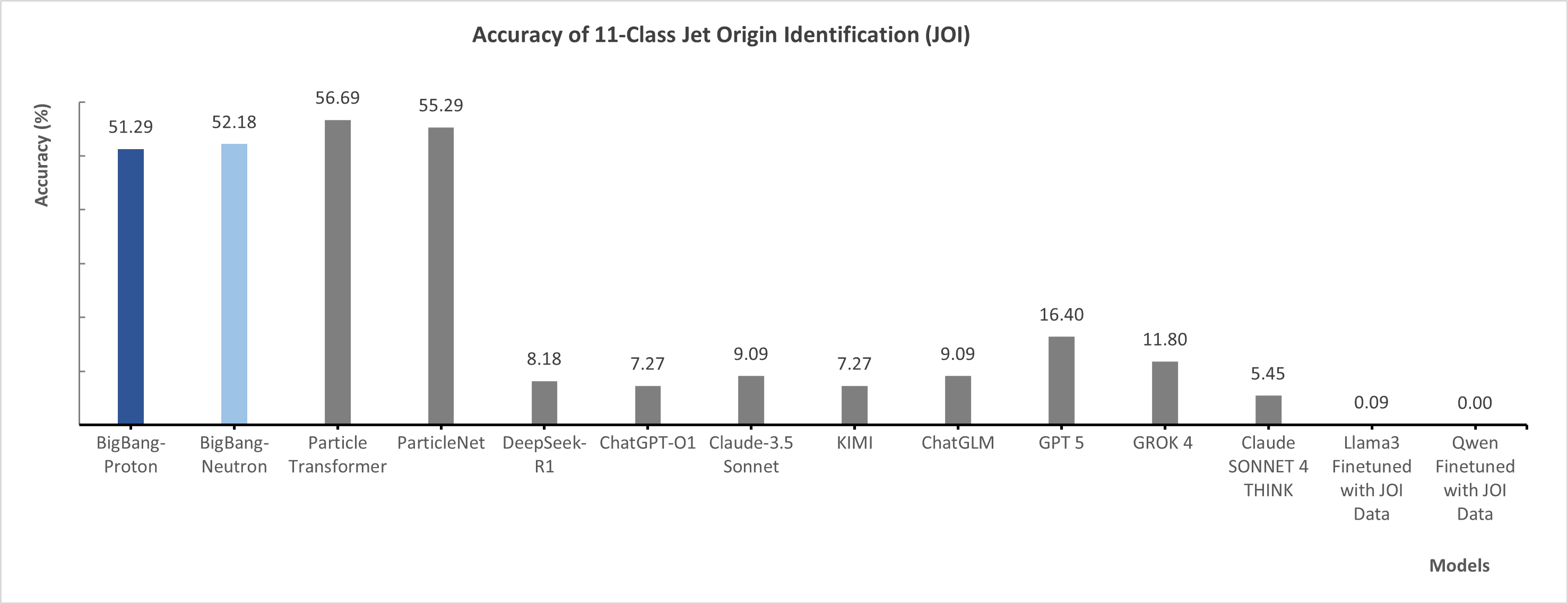} 
    \caption{Comparison of performance among BigBang-Proton, ParticleNet, and other models on the JoI task in particle physics. The bar chart highlights the accuracy of each model. BigBang-Proton, BigBang Neutron, ParticleNet, Particle Transformer are trained on 11M jets data, Llama3, Qwen2.5 are trained on 1.1M jets data. Other general LLMs are tested on 110 sample using zero-shot prompt.}
    \label{fig:particle_physics_performance_comparison}
\end{figure}

Figure~\ref{fig:particle_physics_performance_comparison} indicates that in 1 million dataset size BigBang-Proton achieved an 11-class classification accuracy of 51.29\%, closely trailing BigBang-Neutron’s 52.18\%, which employs identical Binary Patch Encoding~\cite{wu2024scalingparticlecollisiondata}. Compared to state-of-the-art (SOTA) specialized models, Particle Transformer (ParT)~\cite{Qu:2022mxj} at 56.69\% and ParticleNet (PN)~\cite{Zhu2023ParticleNetAI} at 55.29\%, BigBang-Proton’s accuracy is only within a narrow 4.0–5.4\% gap of these specialized models. These results highlight the effectiveness of language-guided classification, demonstrating its feasibility and ability to rival specialized SOTA models in complex particle physics tasks. This underscores the potential of language-guided-classification in high-accuracy scientific computing tasks. We further evaluated the performance of the BigBang-Neutron model alongside these leading models utilizing an 11-dimensional confusion matrix $M_{11}$ that classifies each jet based on the category with the highest predicted probability, as illustrated in Fig.~\ref{fig:M11}. Within the quark sector, the $M_{11}$ matrix is roughly symmetric and can be divided into $2 \times 2$ blocks, each corresponding to a specific type of quark. This confusion matrix offers a detailed insight into the classification performance of the models, highlighting both accurate and erroneous predictions across various jet categories.
Subsequently, two critical metrics are examined including jet flavor tagging efficiencies and charge flip rates. The jet flavor tagging efficiency is defined as the percentage of jets being tagged correctly, without distinguishing the charges of jets. The charge flip rate is calculated as the ratio of the off-diagonal elements to the total sum of the confusion matrix block, quantifying the probability of incorrectly identifying the jet charges, where a lower value indicates better performance. Figure~\ref{fig:effFlip} displays the flavor tagging efficiencies and charge flip rates for each quark type, showing that PN, ParT, BigBang-Neutron and BigBang-Proton exhibit similar performance at 1 million data volume. To investigate the model's data scaling behavior and emergent abilities\cite{wei2022emergentabilitieslargelanguage}, we conducted a series of experiments with varying dataset sizes. Specifically, we trained the model on subsets of the original dataset, ranging from 10\% to 100\% of the full data, and evaluated its performance on the same validation set. 

\begin{figure}[H]
		\centering
		\includegraphics[width=1\textwidth]{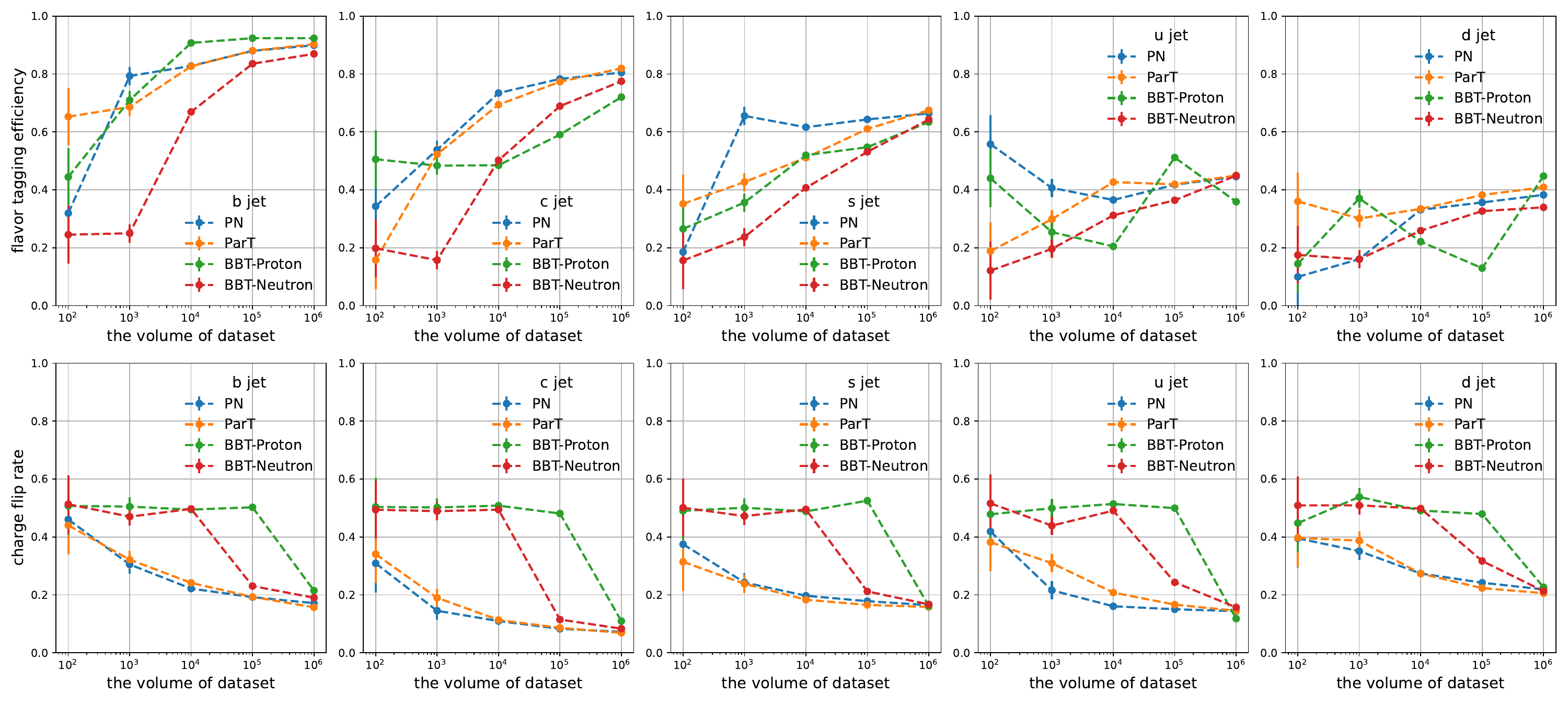} 
		\caption{Scaling performance of BigBang-Proton compared to BigBang-Neutron, ParticleNet, and Particle Transformer on the JoI task with increasing dataset sizes, from 100 samples to 1 million per class. In the flavor tagging task, BigBang-Proton and BigBang-Neutron show similar performance patterns for b and c quarks, starting from near-random guessing and increasing slowly to a threshold before jumping suddenly. For s, u, and d quarks, this pattern transitions to monotonically increasing or fluctuating growth. Specialized models demonstrate a slower rate of improvement in flavor tagging. In the charge flip rate, the sudden jump pattern becomes stronger, as shown in the figures for all five quarks. We can see a clear demonstration that BigBang-Proton requires more data than BigBang-Neutron to reach the critical point. Meanwhile, specialized models show consistently similar performance across increasing data sizes. }
		\label{fig:joi_scaling}
\end{figure}

As shown in Figure~\ref{fig:joi_scaling}, BigBang-Proton model's performance sharply increases as the dataset size grows beyond a threshold. BigBang-Proton demonstrates the most significant performance improvement in flavor tagging efficiency for b-jets, outperforming other models when the dataset size exceeds $100$ and converges to a stable level after that. In comparison, BigBang-Neutron takes 1000 samples to start the performance jump. For c-jets, the flavor tagging efficiency of the BigBang-Proton remained flat around 45\% and then started to improve as the dataset size surpasses $10^4$. For d-jet, BigBang-Proton demonstrated a continual growth from the starting point. For both c-jet and d-jet BigBang-Proton demonstrated a momentum of increasing in efficiency, while others models showed signs of convergence after criticality point $10^4$. For u-jets and d-jets, the performance of the BigBang-Proton model is more volatile, showing fluctuations that do not exhibit a clear trend with increasing data volume.  

Turning to the charge flip rate, Figure~\ref{fig:joi_scaling} shows that the trends for the transformer-based BigBang-Neutron and BigBang-Proton are similar. At smaller dataset sizes, these models remained near random and do not exhibit any performance jump in their charge flip rates. However, as the dataset size increases to a certain critical dataset size, the charge flip rate drops sharply to levels comparable with specialized models like PN and ParT. Notably, BigBang-Proton reaches this critical point at a larger dataset size of around $10^6$, whereas the BigBang-Neutron model sees this drop at $10^5$, and PN and ParT basically have not demonstrated emergent behavior in this task. 

The observed disparity in critical dataset sizes for emergent behavior\cite{wei2022emergentabilitieslargelanguage} in charge flip rate, approximately \(10^6\) samples for the auto-regressive BigBang-Proton versus \(10^5\) for the transformer-based BigBang-Neutron with a task-specific classification head, highlights fundamental differences in how these architectures learn specialized tasks. Auto-regressive models, such as BigBang-Proton, inherently focus on next-word prediction, which frames classification as a generative process (e.g., language-guided reasoning over particle data). This in-context approach\cite{dong2024survey, song2022comprehensivesurveyfewshotlearning, gpt2, brown2020language} requires the model to build richer, more compositional representations to map inputs to outputs via sequential generation, often necessitating larger datasets to capture subtle patterns like charge flips that emerge from probabilistic dependencies across contexts. In contrast, BigBang-Neutron's direct classification head optimizes explicitly for the task, enabling faster convergence on discriminative features with fewer examples, as it bypasses the need for generative intermediate steps. Specialized models like ParticleNet (PN) and Particle Transformer (ParT) do not exhibit such emergent performance jumps, achieving good performance with much smaller datasets, likely because their architectures are designed task-specifically and incorporate embedded domain knowledge in structure, enabling stronger inductive bias and more efficient early learning. BigBang-Neutron employs a more general sequence-based architecture. The trend observed from the specialized models to BigBang-Neutron and BigBang-Proton suggests that more general architectures require larger datasets and greater training compute to reach phase transition. 

As shown in Figure~\ref{fig:particle_physics_performance_comparison}, The performance of general-purpose LLMs on the 11-class particle jet tagging task is profoundly inadequate, with all models performing at levels close to random guessing (10\%). DeepSeek-R1 (8\%), ChatGPT-O1 (7\%), Claude 3.5 Sonnet (9\%), KIMI (7\%), and ChatGLM (9\%) achieve only limited accuracies. Even the advanced versions, which are specifically designed for complex reasoning, GPT-5 (16.4\%), Grok4 (11.8\%), and Claude Sonnet 4 thinking mode (5.45\%) show only marginal improvements, demonstrating that their sophisticated reasoning capabilities are ineffective for this scientific task. As illustrated in Appendix~\ref{appendix_particle}, The analysis provided by DeepSeek-R1, Claude 3.5, and even the advanced reasoning models including GPT 5 and Grok 4 reveals a consistent pattern of shallow reasoning that fails to engage with the physical principles underlying particle jet formation. Their analysis is limited to a checklist of easily extractable features including particle count, energy distribution, particle type composition, and basic kinematic variables, typically covering multiplicity, softness/hardness, pions, photons, kaons, impact parameters, and angular separation. While these characteristics are mentioned in the data, the models merely count and categorize them without demonstrating any understanding of the causal mechanisms that generate these features. For instance, DeepSeek-R1 correctly notes that gluon jets have higher multiplicity due to a larger color charge but treats this as a memorized fact to be matched, not as a principle to be applied. It observes "soft radiation" and "large Delta R" as standalone facts, failing to connect them to the fundamental process of parton showering, where a high-energy parton radiates lower-energy gluons in a branching process governed by QCD. Similarly, Claude 3.5's focus on "charged particle dominance" and "mix of charged and neutral particles" is a mere description of the data, not an analysis. GPT-5's calculation of "displacement significance" ($s_{\text{d0}}$ and $s_{\text{z0}}$) uses this to make a probabilistic guess, stating "lean towards c-jet", based on memorized patterns of b vs. c jets, rather than simulating the actual hadronization process. This approach is fundamentally flawed because it treats jet classification as a pattern-matching exercise on a set of hand-crafted features, akin to a multiple-choice test where the answers are already known, similar format as current LLM benchmarks\cite{hle, mmlu, gpqa}. The models are unable to simulate the underlying physics. They can only recognize statistical correlations learned from their training data. 
Besides testing zero-shot reasoning, we further evaluated two mainstream large language models, Qwen 2.5\cite{qwen2025qwen25technicalreport} and Llama3.2\cite{dubey2024llama} by fine-tuning them on 1.1M jets (100,000 samples per class). Then we conducted zero-shot inference. As shown in Figure~\ref{fig:particle_physics_performance_comparison}, the results illustrate that Llama3.2 achieved an accuracy of only 0.09\%, while Qwen 2.5 achieved 0\%. This result indicates that mainstream LLMs with BPE tokenizers exhibit a fundamental limitation in processing large-scale numerical data. 

\subsection{Inter-atomic Potential Simulation: Language-guided Regression}

   Accurate prediction of material properties through modeling of large-scale electron interactions within atomic lattices has traditionally relied on high-fidelity first-principles approaches such as density functional theory (DFT)\cite{kohn1965selfconsistent} and ab initio molecular dynamics (AIMD)\cite{marx2009ab} by explicitly calculating the electronic structures to provide quantum mechanical accuracy. However, the computational complexity of DFT scales with the cubic order of the system size ($\mathcal{O}(N^3)$) rendering it impractical for large-scale and long-time calculations. To address this limitation, two primary strategies have been developed to reduce the computational complexity of atomistic simulations.The classical force field methods, which approximate atomistic interactions using empirical inter-atomic potentials, , significantly lower computational demands, at the cost of reduced accuracy compared to quantum mechanical methods. The second uses machine learning inter-atomic potentials (MLIPs), which are trained on DFT-labeled datasets to learn surrogate models of atomic interactions\cite{xie2018crystal}. MLIPs achieve a favorable balance between computational efficiency and accuracy, often approaching the precision of DFT. Contemporary MLIP frameworks predominantly utilize graph neural networks (GNNs)\cite{xu2022predicting,batzner2022e3,gasteiger2021gemnet,batatia2023foundation,zhang2021phase,fu2022forces} as their underlying architecture, utilizing the natural graph structure of atomic systems where atoms are represented as nodes and interatomic interactions as edges. To generalize beyond task-specific models tailored to particular material classes, universal MLIP models pretrained on periodic table, (e.g.,M3GNet and DPA-2)\cite{chen2022universal,choudhary2023unified,takamoto2022towards,zhang2024dpa2largeatomicmodel} have been developed. These are trained on diverse datasets derived from DFT calculations across a broad range of material types. Despite these successes, GNN-based MLIPs are inherently limited by their reliance on atomic structure data that can be explicitly derived from DFT calculations, including atomic compositions, crystal structures, and quantum mechanical properties such as energy, forces, stress and magnetic moments\cite{deng2023chgnetpretraineduniversalneural,jain2013commentary}. These models are unable to directly incorporate multi-modal data, which constitutes a significant and underutilized source of scientific knowledge in materials science\cite{birhane2023science}, including Crystallographic Information File(CIF) datasets\cite{antunes2024crystal}, scientific literature, images from experimental measurements, etc. Recent efforts on exploring LLMs for material science \cite{kim2024large,cavanagh2024smileyllama,weininger1988smiles,ock2023catalyst,antunes2024crystal} have particularly focused on incorporating textual data such as Simplified Molecular Input Line Entry System (SMILES) strings,CIF,chemical formulas in model training, ignoring the complete atomic structure information stored as numerical data. These LLMs pretrained or finetuned on textual data usually are not capable to process regression tasks which deal with numerical data, such as energy prediction.  Matterchat\cite{tang2025matterchatmultimodalllmmaterial} have tried to develop LLMs to generate language-guided material properties prediction, including regression tasks, while relying on external graph-based model\cite{deng2023chgnetpretraineduniversalneural} to provide embedded knowledge of crystal structures. 

In this downstream task, we present that BigBang-Proton, an auto-regressive LLM pretrained on blend of multi-discipline datasets including arithmetic, particle physics, material science, biology can outperform specialized GNN-based model in high-throughput material discovery of stable crystals. Simply by next-word-prediction, BigBang-Proton can process tasks of classification and regression in MLIP. For most LLM-based models for material science, regression tasks are difficult since LLMs can not learn and predict numerical sequence well. In contrast BigBang-Proton demonstrates extraordinary performance in regression tasks, such as formation energies, forces, stresses, magnetic moments prediction, and energy above convex hull calculation. 

\begin{figure}[H]
     \centering
     \includegraphics[width=1\textwidth]{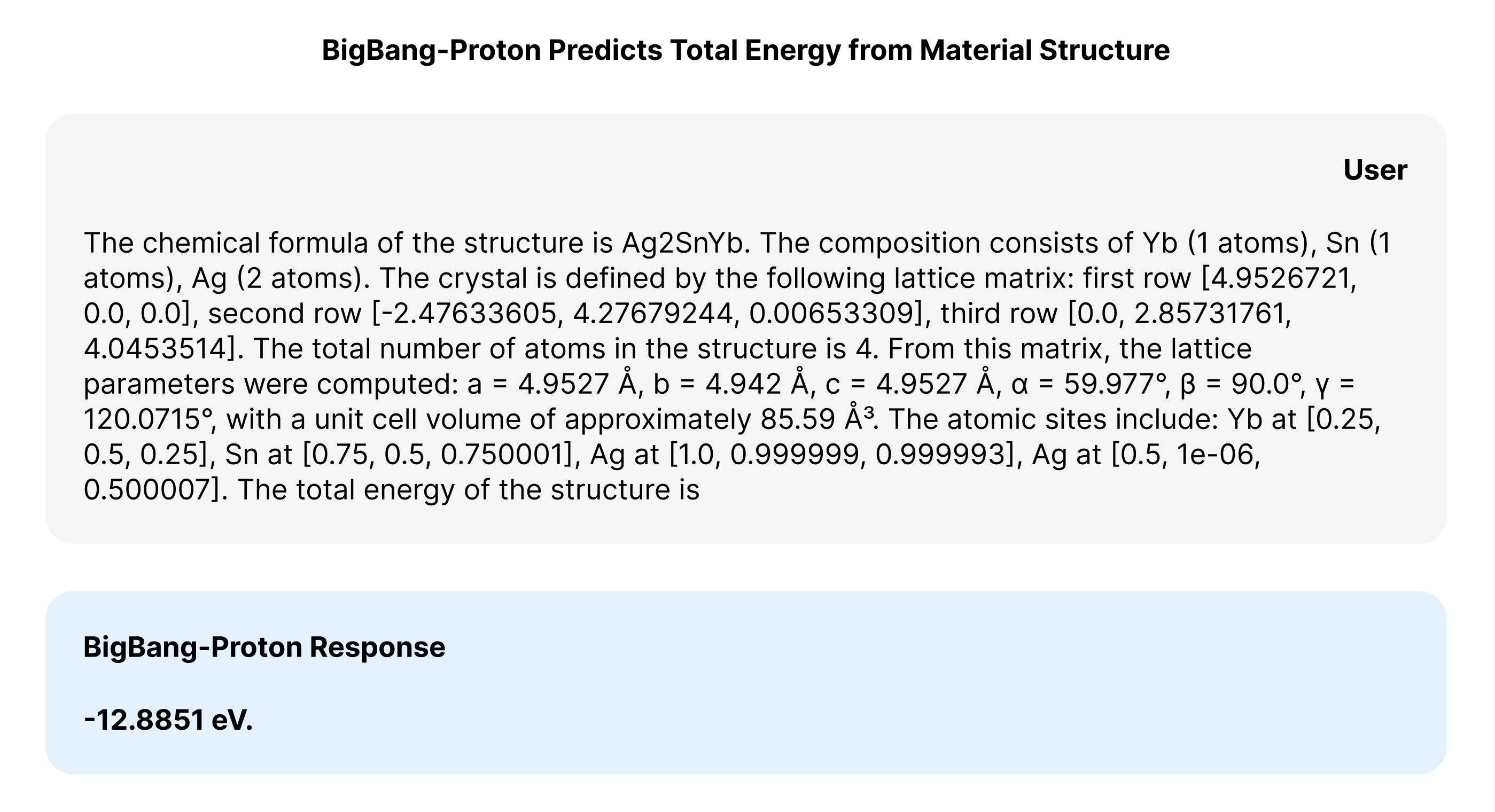} 
     \caption{Language-guided inter-atomic potential simulation by BigBang-Proton using a zero-shot prompting approach. The material structure data were decomposed into parts, with each part embedded in corresponding textual descriptions. BigBang-Proton learned complete context that integrated theoretical descriptions and experimental data to make final inferences. Compared to traditional specialized models that relied solely on numerical data for regression, this approach leveraged the aligned theoretical and experimental contexts to conduct regression. Scientific principles and facts, including "formula", "composition", and "atomic sites", were integral components of the simulation. }
     \label{fig:chat_material}
\end{figure}

BigBang-Proton are trained on Material Project Trajectory (MPTrj) datasets from material project following the requirements of MatBench\cite{dunn2020} material discovery\cite{janosh2024matbench} compliant policy. MPTrj\cite{jain2013commentary} is a large database of DFT-relaxed crystal structures generated mainly from experimentally-tested crystals, consisting of 94 elements,145,923 inorganic compounds, and specifically 1,580,395 atom configurations, 1,580,395 energies, 49,295,600 forces, 7,944,833 magnetic moments and 14,223,555 stresses. Atomic structure data are converted uniformly to byte patch sequences before training. We employ Wang-Botti-Marques dataset(WBM) dataset\cite{wang2021predicting} with 257,487 structures to evaluate capabilities of BigBang-Proton trained MPTrj\cite{jain2013commentary} on new material discoveries\cite{janosh2024matbench}. WBM dataset are generated by substitution of atoms in the existing crystal structures in MPTrj\cite{jain2013commentary} with chemically similar ones, which is defined quantitatively by statistical analysis on Inorganic Crystal Structure Database(ICSD)\cite{bergerhoff1983inorganic}. Natural language context to describe the crystal are aligned with atomic structure data as seen in Figure~\ref{fig:chat_material} to achieve language-guided material structure prediction. The language context can be extended to long-horizon chain of thoughts for the requirements of different scenarios. 

\begin{figure}[H]
     \centering
     \includegraphics[width=1\textwidth]{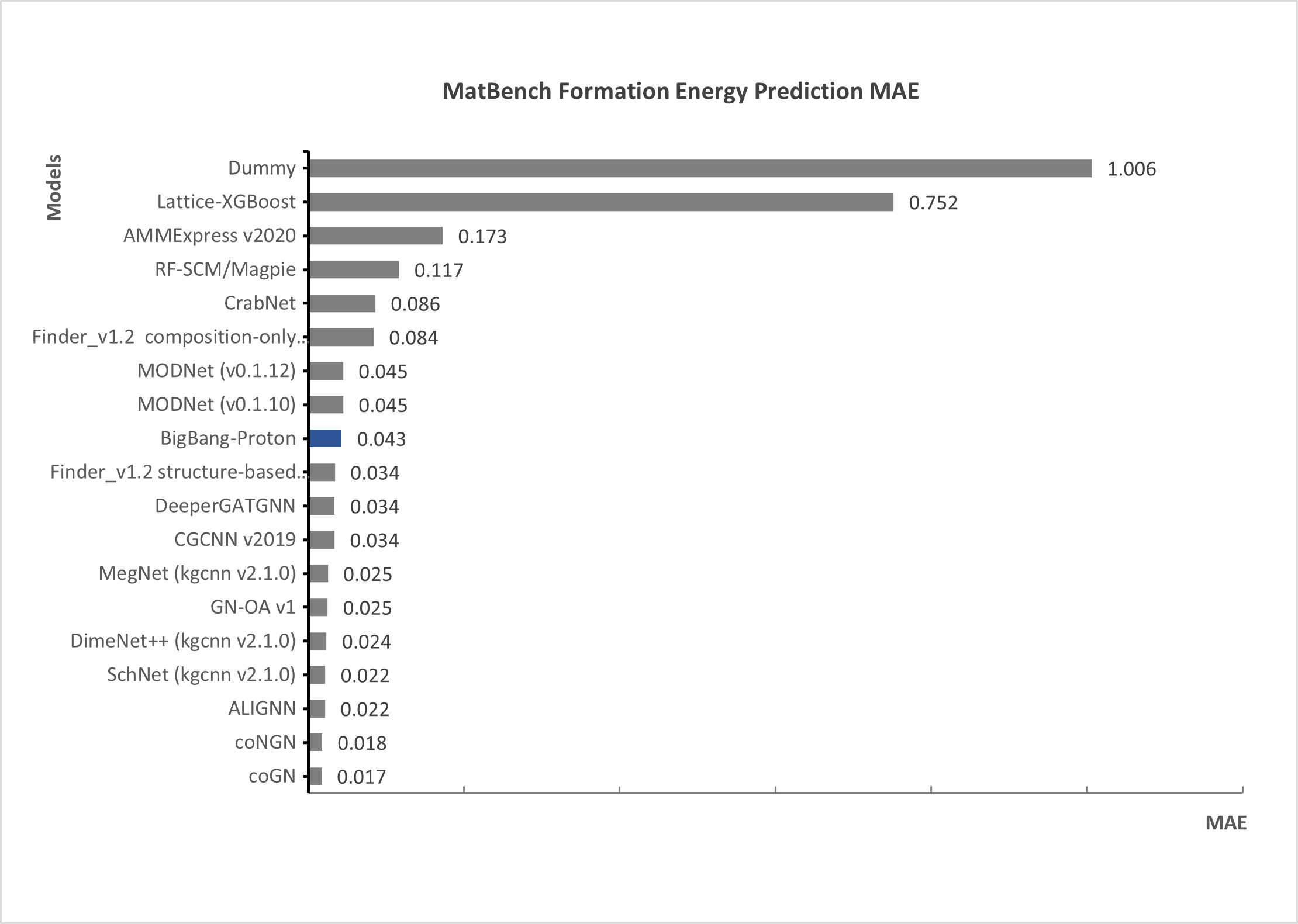} 
     \caption{MatBench leaderboard formation energy prediction MAE. Compared to specialized models mostly built upon graph neural networks (GNNs), BigBang-Proton achieved an MAE of $0.043$ eV in formation energy calculation. Although there is still a gap to the state-of-the-art level, this performance demonstrates the practical utility of BigBang-Proton and language-guided regression techniques for real-world inter-atomic potential simulation tasks, validating the effectiveness of theory-experiment alignment in computational materials science. }
     \label{fig:formation energy}
\end{figure}

Matbench\cite{dunn2020} benchmark lists 13 tasks with 10 for regression and 3 for classification. For the 3 classification task, we can apply models to generate intermediate results by regression. We choose regression task formation energy prediction to demonstrate the capability of BigBang-Proton on numerical computing. Shown in Figure~\ref{fig:formation energy} BigBang-Proton achieves mean absolute error(MAE)  0.043 eV/atom in formation energy prediction, ranking 11st in Matbench leaderboard at the time of writing. BigBang-Proton outperforms some specialized machine learning methods such as AMMExpress(MAE=0.117 eV/atom)\cite{dunn2020}, surface complexation modeling RF-SCM\cite{Breiman2001,Ward2016,Faber2015}(MAE=0.117eV/atom), attention-based model CrabNet\cite{Wang2021crabnet}(MAE=0.086 eV/atom) and feature selection model MODNet\cite{De_Breuck_2021,DeBreuck2021}(MAE=0.045 eV/atom). However, it falls short of specialized GNNs like coGN\cite{ruff2023}(MAE=0.017 eV/atom), ALIGNN\cite{choudhary2021}(0.022 eV/atom) and SchNet\cite{schnet2018}(0.022 eV/atom). LLM-based model Matterchat\cite{tang2025matterchatmultimodalllmmaterial} achieves a root mean square error (RMSE) of 0.121 eV/atom in Matbench formation energy prediction, with a significantly larger error than BigBang-Proton. It indicates that general purpose large language models\cite{jiang2023mistral7b} trained on embeddings from specialized GNN model CHGNet\cite{deng2023chgnetpretraineduniversalneural} have not obtained the same understanding of material structures as end-to-end trained BigBang-Proton. The current implementation uses only a 1.5 billion parameter model with limited training data and conventional fine-tuning methods, leaving substantial room for improvement through scaling to larger parameter, expanding training datasets, incorporating advanced fine-tuning techniques, and integrating post-training reinforcement learning approaches. These enhancement pathways are inherently accessible to our language-guided simulation framework but remain impossible for traditional GNN-based approaches. 

\begin{figure}[H]
    \centering
    \includegraphics[width=1\textwidth]{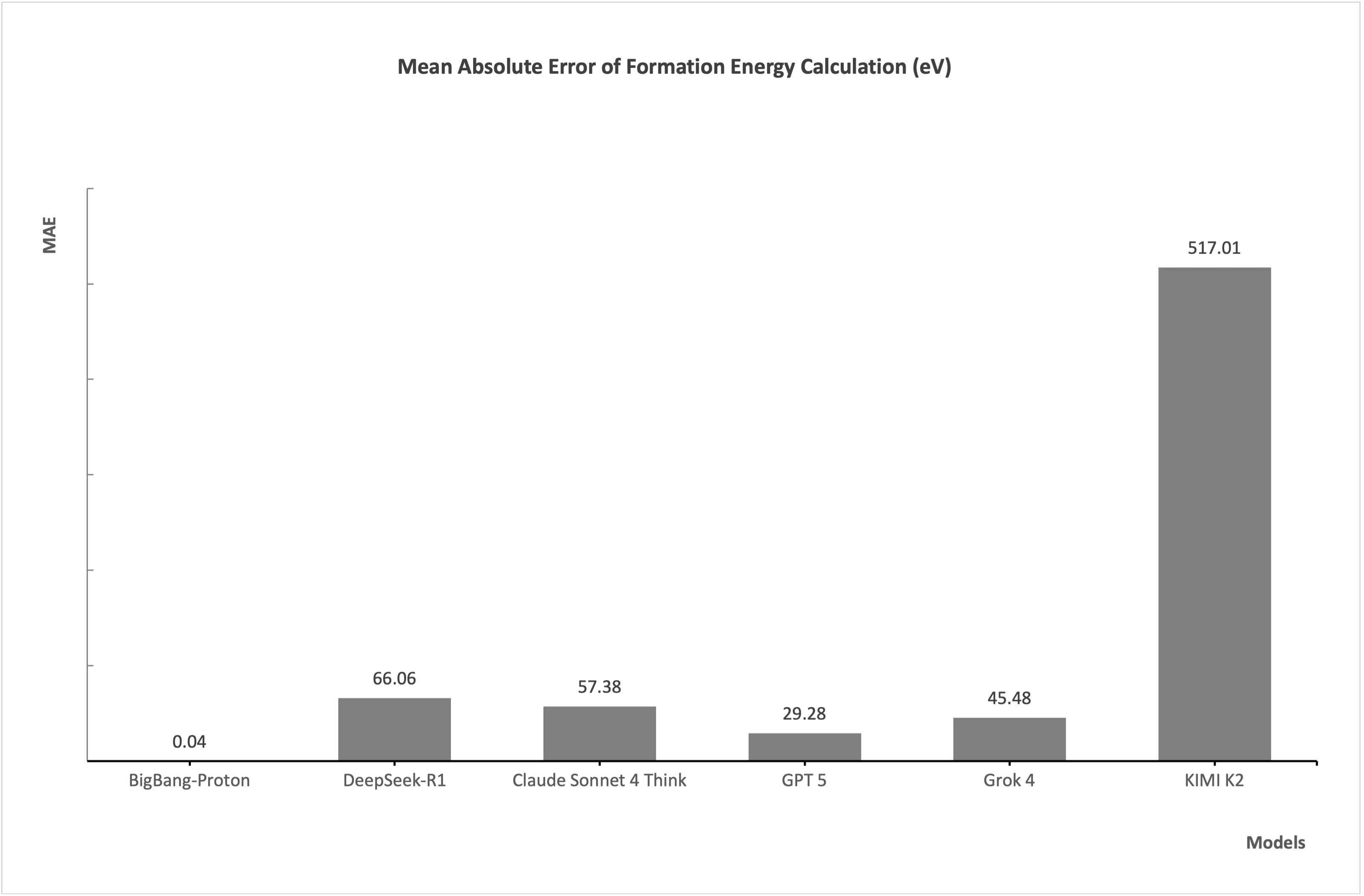} 
    \caption{Comparison of formation energy prediction performance among BigBang-Proton and mainstream LLMs. These LLMs achieved MAEs ranging from ~$20$ eV to ~$500$ eV, three or four orders of magnitude larger than the physically meaningful threshold. These results clearly demonstrate the fundamental limitations of mainstream general-purpose LLMs in solving real-world tasks such as inter-atomic potential simulation, while also indicating the limited effectiveness of current LLM benchmarks~\cite{gpqa, hle} in question-answer formats for measuring the capabilities of general LLMs.}
    \label{fig:lattice_energy_prediction_comparison}
\end{figure}

In Figure~\ref{fig:lattice_energy_prediction_comparison} We compare formation energy prediction performance of mainstream reasoning LLMs, including DeepSeek R1 (MAE: 66.056 eV), GPT 5 (MAE: 29.279 eV), Claude Sonnet 4 (MAE: 57.377 eV), and KIMI K2 (MAE: 517.006 eV) with BigBang Proton (MAE: 0.043 eV). The results from these LLMs are far from applicable in inter-atomic potential simulation. In a zero-shot learning scheme, we prompt the model to predict 110 sets of formation energies with atomic structures from MPTrj\cite{deng2023chgnetpretraineduniversalneural} test dataset as input. The reasoning models usually generate long and complicated reasoning steps, particularly by DeepSeek R1, as seen in Appendix~\ref{append:material_llms}, starting from explaining the dataset, with some simple deductions such as counting atom numbers and energy values or obvious attributes in dataset, then making final guesses on formation energy. Sometimes LLMs attempt to write code to perform calculations, which is not our intended approach. While LLMs demonstrate certain understanding of basic atomic structure configurations, including atom counts and quantitative correlations between attributes, they fail to grasp the underlying physics mechanisms determining total energy, formation energy, and other physical quantities. The dramatic performance gap between mainstream LLMs and BigBang-Proton stems from differences in data, tokenization, and architecture. While mainstream LLMs are pretrained on the entire internet texts including scientific literature and textbooks that contains nearly all known condensed matter theory, they have not been trained on numerical datasets such as MPTrj\cite{deng2023chgnetpretraineduniversalneural} or Materials Project\cite{jain2013commentary}, which contain DFT-simulated results, leaving them unable to intrinsically learn inter-atomic potentials. Physical structures typically described using numerical data which are calculated from physics laws can not break down to natural languages. Although these models could be fine-tuned on such task-specific data, their reliance on BPE tokenizers fundamentally hinders effective numerical learning, as demonstrated by Matterchat\cite{tang2025matterchatmultimodalllmmaterial}'s performance compared to BigBang-Proton. Moreover, using mainstream LLMs as scientific agents that call external tools\cite{ghareeb2025robinmultiagentautomatingscientific} built upon task-specific models would limit the capabilities of discoveries on the scope of these specialized models. 

\begin{figure}[H]
     \centering
     \includegraphics[width=1\textwidth]{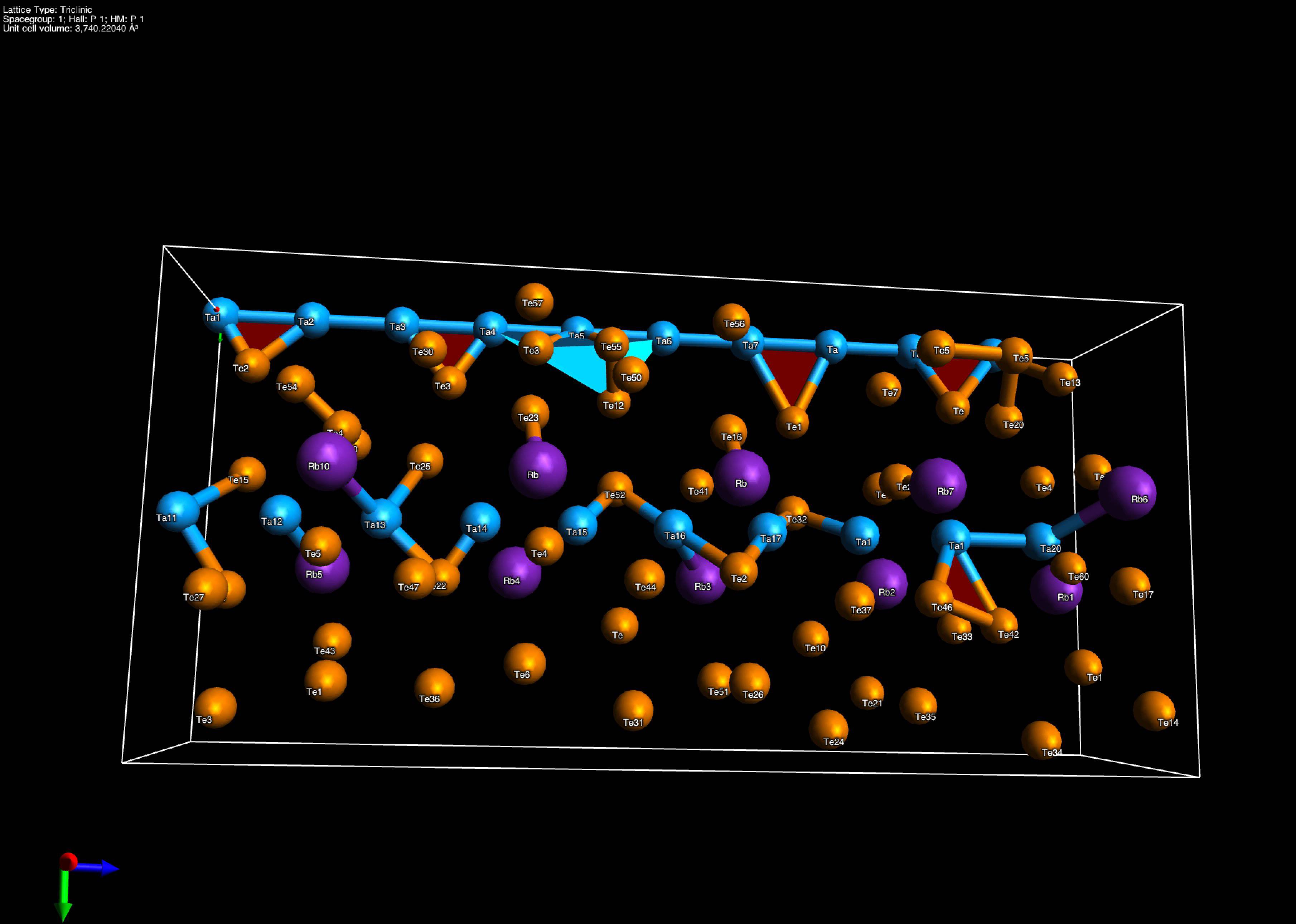} 
     \caption{A demo for new material discovery based on structure data of 90-atom molecular compound $Te_{60}Ta_{20}Rb_{10}$ includes three basic steps: First, crystal structure data is prompted to BigBang-Proton to predict the total energy of the structure, yielding -504.07 eV. Second, the predicted total energy and calculated forces are used to determine the energy above hull and assess the thermal stability of the new material, which is found to be <0, indicating structural stability and matching the label in WBM. Finally, these results are applied to predict various physical and chemical properties of the discovered material.}
     \label{fig:material discovery demo}
\end{figure}

Figure~\ref{fig:material discovery demo} demonstrates a demo for new material discovery based on structure data of the 90-atom molecular compound $Te_{60}Ta_{20}Rb_{10}$, which includes three detailed steps: First, the crystal structure data containing 90 atoms with triclinic symmetry (space group $P1$, lattice parameters $a = 8.07~\text{\AA}$, $b = 14.75~\text{\AA}$, $c = 31.41~\text{\AA}$) is fed into BigBang-Proton, which processes the atomic coordinates and chemical composition to predict the total energy of the structure as $-504.07~\mathrm{eV}$, demonstrating the model's capability to handle complex, large-scale, multi-element systems with precise energy calculations. Second, the predicted total energy, along with force calculations derived from energy gradients, is utilized to compute the energy above hull. For $Te_{60}Ta_{20}Rb_{10}$, BigBang-Proton obtained $E_{\mathrm{above\_hull}} < 0$, aligning with the stability label provided in the WBM database. The energy above hull assesses the thermodynamic stability of this ternary compound and confirms its structural viability. Finally, the model can extrapolate to predict additional physical and chemical properties, including electronic band structure, elastic moduli, and chemical reactivity patterns, showcasing the comprehensive material discovery pipeline enabled by BigBang-Proton's advanced reasoning capabilities on complex atomic structures. Figure~\ref{fig:material discovery demo} shows the atomic structures, site distributions, symmetries, and chemical bonds of $Te_{60}Ta_{20}Rb_{10}$ based on BigBang-Proton's language-guided calculation of total energy ($-504.07~\mathrm{eV}$), illustrating that the large unit cell dimensions ($8.07 \times 14.75 \times 31.41~\text{\AA}^3$) accommodate the complex atomic arrangements necessary to balance the competing bonding requirements of the three distinct chemical species. This structure visualization demonstrates that BigBang-Proton's language-guided simulation of interatomic potential enables detailed structural analysis of complex multi-element materials. 

\subsection{Lake Water Quality Prediction: Language-guided Spatiotemporal Prediction}
\label{subsec:lake_water_prediction}

Earth system modeling represents humanity's ambitious effort to understand and predict the complex interactions between the atmosphere, hydrosphere, cryosphere, land surface, and biosphere\cite{heavens2013studying}. Within this framework, water cycle modeling stands as a critical component, encompassing both oceanic and continental water systems. Continental water modeling focuses on freshwater systems, particularly lakes and rivers, which serve as sentinels of environmental change and play a disproportionately important role in global biogeochemical cycles. Lake water quality\cite{chapra97, ai5040098} prediction as a typical cross-discipline task, sits at the intersection of geoscience, limnology, biology, chemistry, environmental engineering and economics, representing one of the most challenging aspects of earth system modeling. Like numerical weather prediction\cite{graphcast, bi2023accurate}, many numerical models for water quality prediction by machine learning approaches have been developed\cite{kang2022effect, lee2007analysis, li2009research}. Chlorophyll-a concentration\cite{yan2024comprehensive, yao2021retrieval, YU2022102119, kwon2018monitoring, kim2022machine, mohebzadeh2021spatial, su2021estimating, niu2023incorporating}, as a primary indicator of algal biomass and ecosystem health\cite{voros1991phytoplankton, lionard2008evaluation}, is a key factor in water quality prediction. The intrinsic difficulty lies in the fact that chlorophyll dynamics are governed by multiple factors including solar radiation, temperature, wind, nutrient loading, water residence time, internal biogeochemical processes, and anthropogenic influences.

\begin{figure}[H]
   \centering
   \includegraphics[width=1\textwidth]{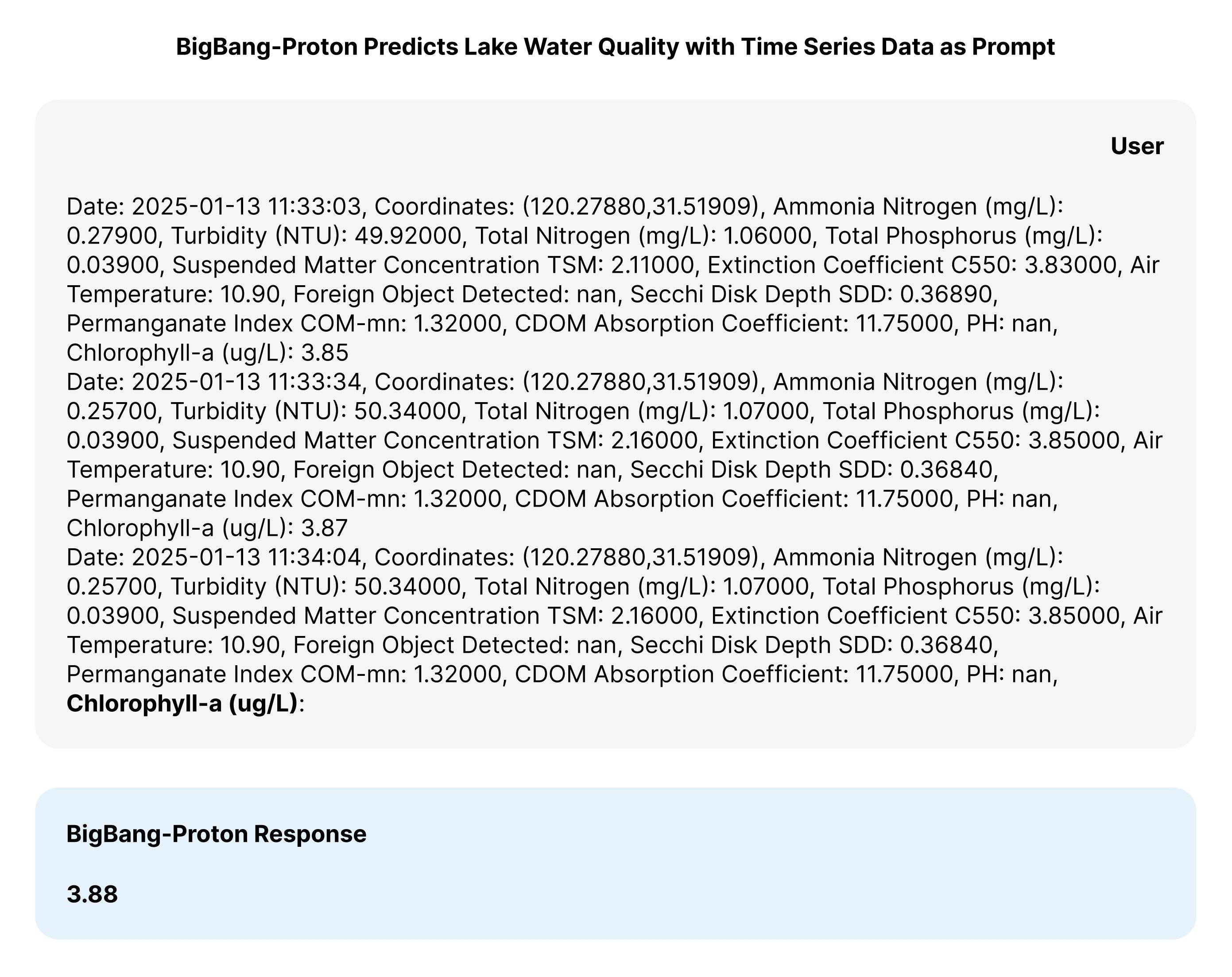} 
   \caption{Illustration of language-guided spatiotemporal prediction examples for lake water quality using BigBang-Proton. Each prompt began with time series data input, then explicitly instructed the model to determine the chlorophyll-a concentration based on the given context. These time series data of multiple factors, generated from sensors, quantitatively described the fluctuations of physical world metrics. Time series modeling was converted to in-context learning and inference for BigBang-Proton. }
   \label{fig:chat_lake}
\end{figure}

In this task, we aim to predict water quality based on historical spatiotemporal data of chlorophyll-a concentrations. BigBang-Proton has been pretrained on spatiotemporal sensor data of 262 millions bytes. We further finetuned BigBang-Proton on the lake water quality data. Lake water quality datasets were collected by sensors deployed in Wuli Lake in City of Wuxi, China. Sensors collect data every 30 seconds, spanning totally 2 years. 80\% dataset were used for training and 20\% for evaluation. As shown in Figure~\ref{fig:chat_lake}, The dataset includes a timestamp with date information, a precise location specified by latitude and longitude, and a comprehensive suite of water quality parameters. These parameters cover nutrients including Ammonia Nitrogen, Total Nitrogen and Total Phosphorus, optical properties such as Turbidity, TSM, C550, SDD, CDOM Absorption Coefficient and Permanganate Index, physical parameters like Air Temperature, and the target variable for prediction, Chlorophyll-a measured in $\mu g/L$. We only use data collected from one single position at the center region of the lake for training. 

The finetuning of BigBang-Proton leveraged a sequence-to-sequence approach, where the input is complete time series data collected from sensors consisting of chlorophyll-a concentration values and other parameters from the past five consecutive time points, and the output is the predicted chlorophyll-a concentration at a future time point. The prompt-chlorophyll-a is demonstrated as in Figure~\ref{fig:chat_lake}. The ultimate goal is to minimize the  Mean Absolute Error (MAE) metric. 

The results show that the model achieves an MAE of $0.58$ $\mu g/L$ for chlorophyll-a concentration predictions. This indicates that the model can effectively capture the temporal dependencies among various environmental and water quality parameters, leading to accurate predictions.In addition to MAE, we also evaluate the model using the Mean Absolute Percentage Error (MAPE) to provide a more comprehensive assessment of prediction accuracy. MAPE is defined as:
$$
\text{MAPE} = \frac{1}{N} \sum_{i=1}^{N} \left| \frac{y_i - \hat{y}_i}{y_i} \right|
$$
where $y_i$ is the true chlorophyll-a concentration, $\hat{y}_i$ is the predicted value, and $N$ is the total number of predictions. For BigBang-Proton, the MAPE was calculated to be 0.098, indicating that on average, the relative error between the predicted and actual chlorophyll concentrations is approximately 9.8\%. These results surpass specialized numerical machine learning models in chlorophyll-a prediction tasks. For instance, \citeay{YU2022102119} achieved an MAE of $1.85$ $\mu g/L$ using a machine learning model trained on satellite data, while \citeay{yao2021retrieval} attained an MAE of $0.998$ $\mu g/L$ and an MAPE of 19.4\%.

As shown in Figure~\ref{fig:chat_lake}, Chlorophyll-a (Chl-a) dynamics are driven by 12 contextual variables including Ammonia Nitrogen, Turbidity, Total Nitrogen, Total Phosphorus, Suspended Matter Concentration, Extinction Coefficient, Air Temperature, Foreign Object Detection, Secchi Depth, Permanganate Index, CDOM Absorption Coefficient, and pH, all measured across time and space. BigBang-Proton eliminates manual feature engineering as done in specialized time series modeling, by treating their collective nonlinear interplay as an in-context learning problem. The model infers Chl-a not from isolated predictors, but from the evolving, high-dimensional context formed by the complex system. 

Moreover, trained on scientific multitask datasets using Theory-Experiment alignment, BigBang-Proton captures not only the temporal dynamics of the local system but also the contextual relationships that may correspond to real-world physical or environmental processes for different scales. This enables the model to predict changes in water quality as part of a broader spatiotemporal and causal framework. These results suggest that the modeling approach used in BigBang-Proton can be extended beyond lake systems to modeling larger scales and more complex system, from regional climate to the Earth system. 

\subsection{Language-guided Genome Sequences Modeling}
One of the key downstream tasks for BigBang-Proton is Gene Sequence Prediction, which focuses on understanding and predicting the functional and evolutionary characteristics of genetic sequences. These sequences encode the biological instructions that drive the development, functioning, and adaptation of living organisms. Variations in genome sequences across species and individuals reflect evolutionary processes, such as natural selection and genetic drift, which shape biological diversity and enable adaptation to changing environments \cite{doi:10.1126/science.32.812.120,WATSON1953,Nirenberg1961TheDO}. Recent advancements in high-throughput sequencing technologies have made it possible to generate vast amounts of genomic data, providing unprecedented opportunities to study these variations at a whole-genome scale \cite{Shendure2012}.
    
In the realm of computational biology, specialized models have been developed to address specific challenges. For instance, AlphaFold represents a series of state-of-the-art task-specific models designed for protein structure prediction using a diffusion architecture \cite{Jumper2021, Abramson2024}. AlphaFold series have achieved remarkable success in predicting three-dimensional protein structures from amino acid sequences.
\begin{figure}[H]
		\centering
		\includegraphics[width=1\textwidth]{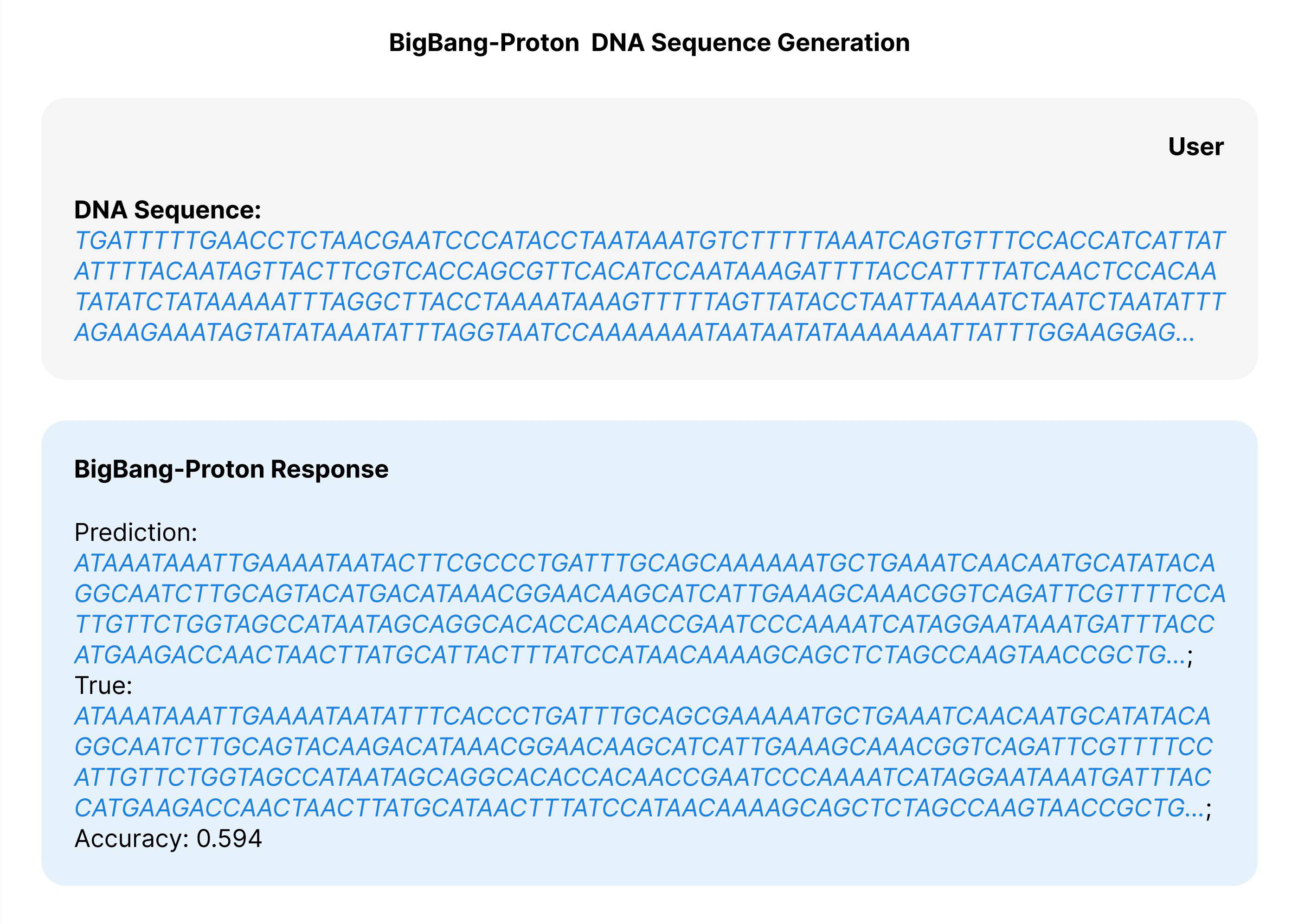} 
		\caption{Illustration of zero-shot examples for gene sequence completion using BigBang-Proton. The user prompts with an initial DNA segment of nucleotide sequence, and the model generates the continuation to complete that sequence. The accuracy of the generated nucleotide sequence is also reflected by training perplexities. }
		\label{fig:biological_prompt_answer}
\end{figure}
Conversely, Evo is an innovative foundational model that integrates DNA, RNA, and protein sequences into a unified framework for joint training \cite{dEric2024evo}. By combining information from multiple biological layers, Evo offers new insights into the complex interactions within and between genomes, enabling more accurate predictions of gene functions and evolutionary pathways.

In the pretraining stage, we mixed the same OpenGenome dataset~\cite{dEric2024evo} of 27.3B nucleotide tokens, which is about one-tenth of the data used in Evo, with datasets of particle physics, material structures, arithmetic, sensors and general texts, to pretrain BigBang-Proton. The OpenGenome dataset includes more than 80,000 bacterial and archaeal genomes, along with millions of predicted phage and plasmid sequences, covering approximately 300 billion nucleotide tokens. For safety considerations, viral genomes that infect eukaryotic hosts were excluded from the training data. In downstream task stage, we further finetuned pretrained model with 82.8B nucleotide tokens. BigBang-Proton have totally learned about 110.1B tokens.
    
The dataset was used to train the model using a next token prediction paradigm, where the model learns to predict the subsequent nucleotide based on the given input sequence. This approach enables the model to understand complex genomic structures purely from the sequence context without relying on explicit annotations. As illustrated in Figure \ref{fig:biological_prompt_answer}, the model's prompt-answer cases demonstrate its ability to predict the sequence that follows a given gene sequence input. For example, given an initial segment of a gene sequence, BigBang-Proton can predict the most likely continuation of that sequence, effectively capturing long-range dependencies and subtle patterns within the genomic data. The same zero-shot genome prediction examples by general purpose LLMs can be seen in Appendix~\ref{append:biology}, which demonstrates near-random guessing. 

The OpenGenome dataset provides a vast collection of genomic sequences from diverse organisms. By predicting the next nucleotide in a sequence, BigBang-Proton can implicitly understand complex genomic features such as promoters, enhancers, and non-coding RNAs. This approach allows the model to develop understanding of various genomic elements purely based on the sequence context. 

\begin{figure}[H]
    \centering
    \begin{subfigure}[b]{1\textwidth}
        \includegraphics[width=\textwidth]{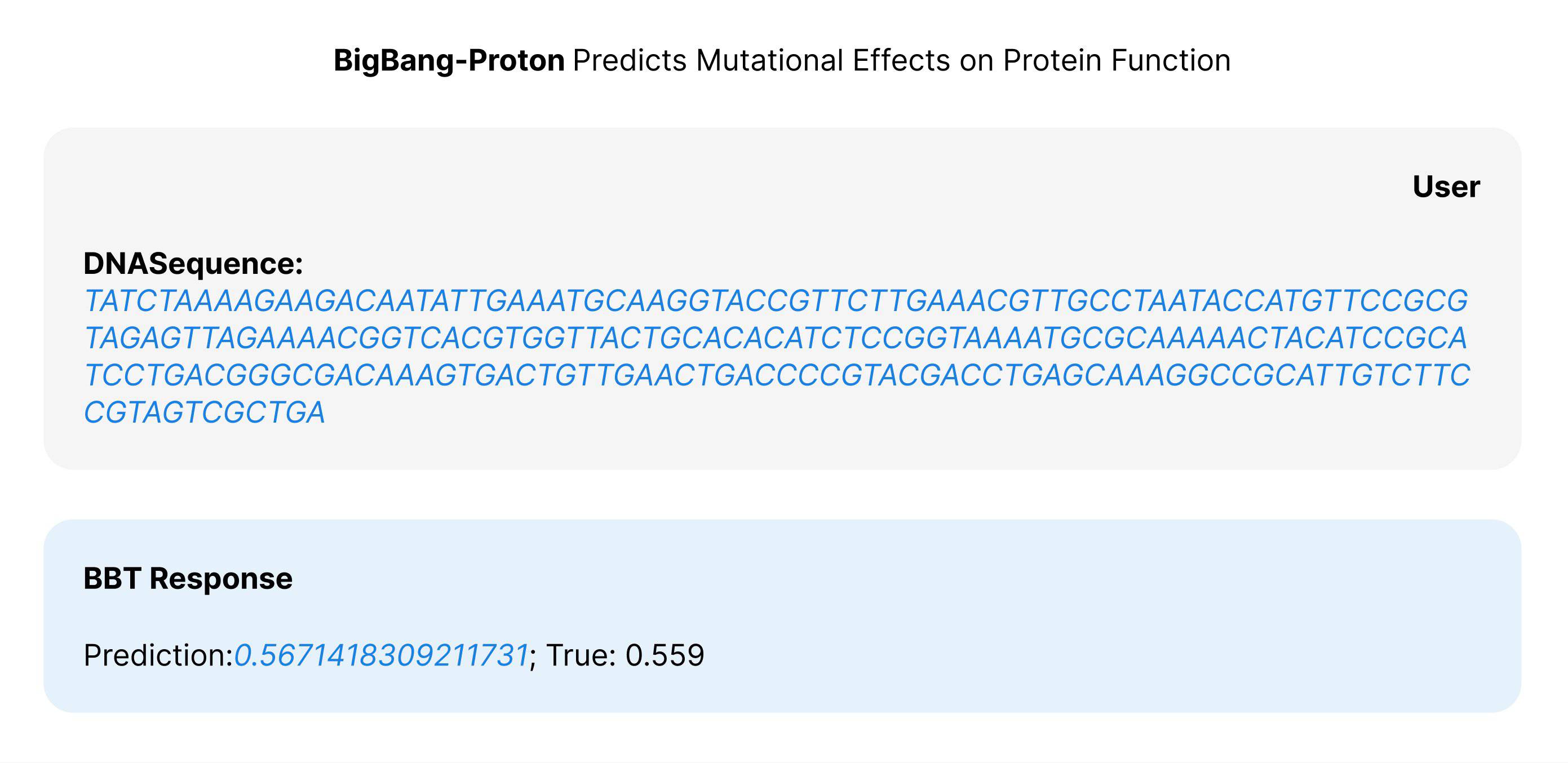}
        \caption{Mutational effects on protein function prediction}
        \label{fig:chat2_protein}
    \end{subfigure}
    \begin{subfigure}[b]{1\textwidth}
        \includegraphics[width=\textwidth]{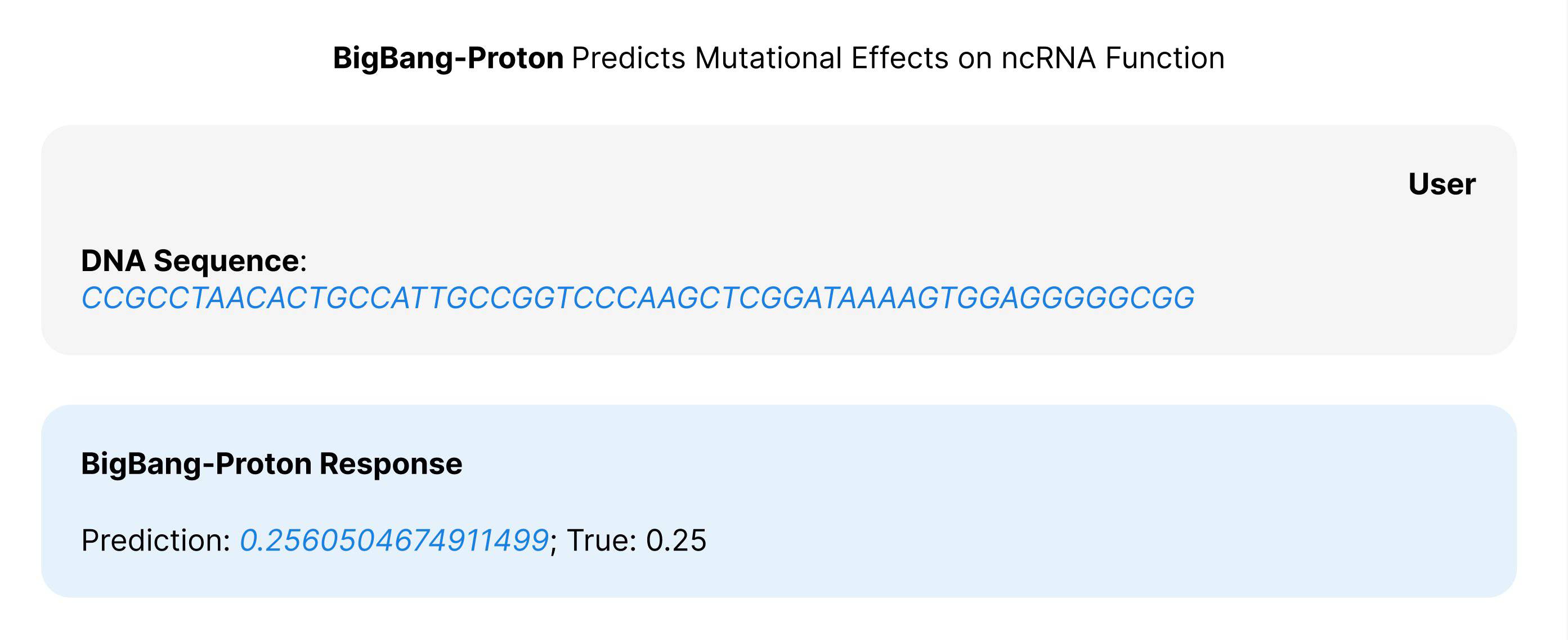}
        \caption{Mutational effects on ncRNA function prediction}
        \label{fig:chat3_ncRNA}
    \end{subfigure}
    \begin{subfigure}[b]{1\textwidth}
        \includegraphics[width=\textwidth]{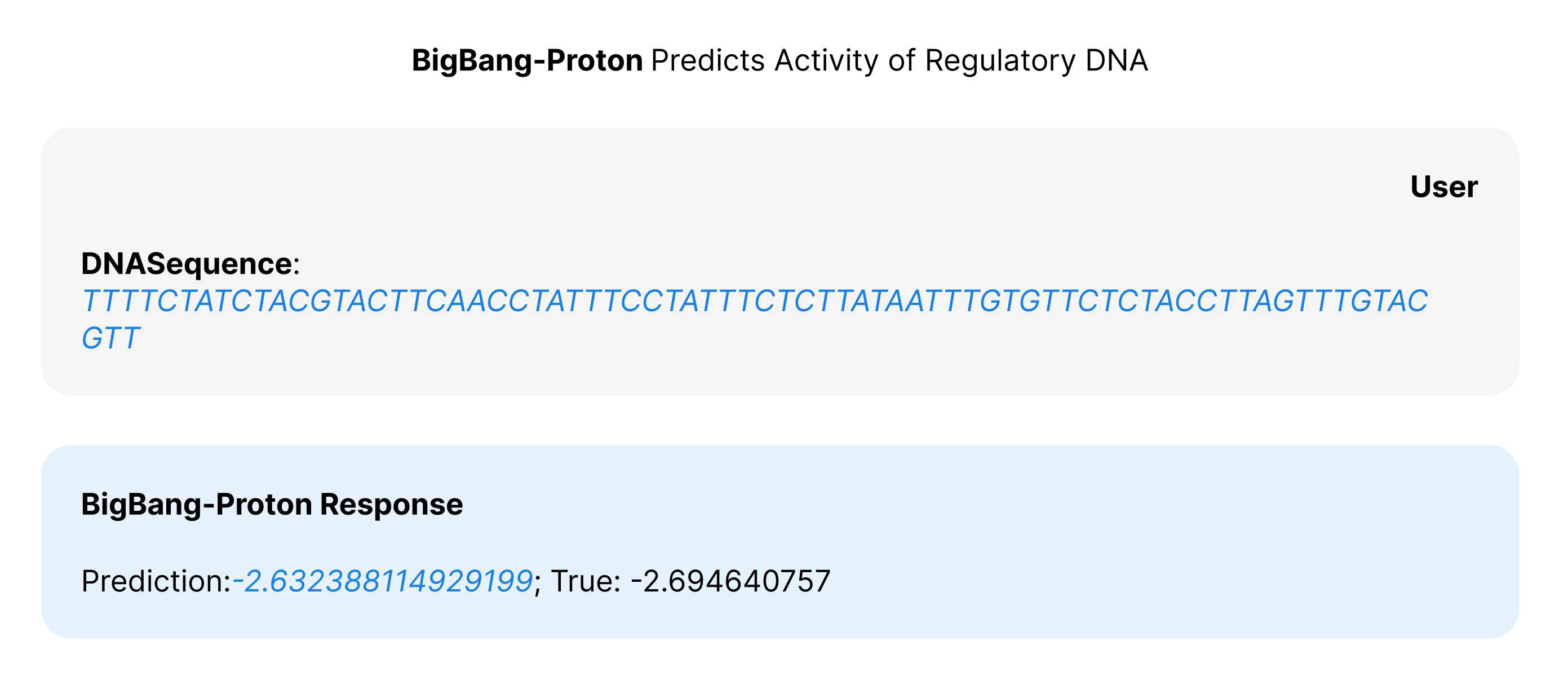}
        \caption{Activity of regulatory DNA}
        \label{fig:chat4_regulatory_dna}
    \end{subfigure}
    \caption{Illustration of zero-shot function prediction examples by BigBang-Proton for proteins, non-coding RNAs, regulatory DNA, using deep mutational scanning (DMS) dataset, in which function fitness score is measured for each protein mutation. Subfigures show the predicted fitness based on the input sequence segment. The final spearman correlations are calculated between fitness datasets of the predicted and labels.}
    \label{fig:combined_genomic_tasks}
\end{figure}

We set the maximum sequence length to 16,384 tokens to accommodate long genomic sequences while balancing computational efficiency. Sequences were padded or truncated to a fixed length to maintain consistency during training. The model was trained for 5 epochs, with early stopping based on validation ppl to avoid overfitting.
BigBang-Proton leverages its ability to learn complex patterns from sequential data to analyze and predict the functional roles of DNA, RNA, and proteins. By capturing the intricate relationships and interactions within genomic sequences, the model can infer how specific genetic variations influence biological processes, contribute to disease mechanisms, or determine phenotypic traits. This capability positions the model as a general-purpose foundation model for genomics, capable of learning the underlying principles of genome organization and function. Such a model has the potential to accelerate discoveries in areas such as functional genomics, evolutionary biology, and precision medicine, enabling researchers to decode the logic of life encoded in genetic sequences.
    
Similar to the scaling law analysis in Evo\cite{dEric2024evo}, we conducted a detailed comparison between BigBang-Proton and Evo in terms of Compute Budget Scaling under the same experimental settings. Compute-optimal refers to the theoretically optimal allocation of computational resources (FLOPs) between model size and training tokens to achieve the best performance. For this comparison, we kept the Dataset Size consistent across both models and focused on the Evaluation Perplexity (Eval PPL) as the primary metric. Perplexity is a widely used measure in sequence modeling tasks, with lower values indicating better model performance.
\begin{figure}[H]
	\centering
	\includegraphics[width=1\textwidth]{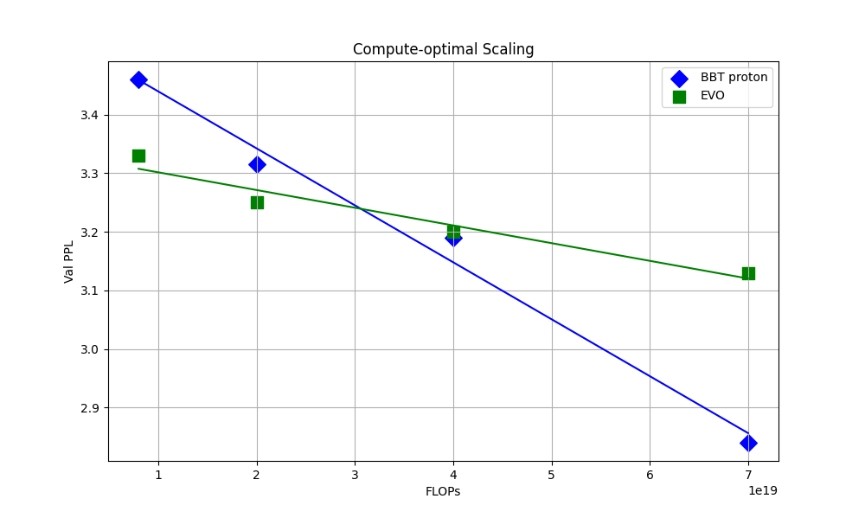} 
	\caption{Comparison of Evaluation Perplexity (Eval PPL) comparison between BigBang-Proton (1.5B) and Evo (7B) compute budgets optimal experiment. BigBang-Proton achieved lower perplexity (PPL = 2.8) using only one-third of the training data (110.1B vs 300B tokens) compared to Evo's optimal PPL of 3.1. Notably, BigBang-Proton continues to show decreasing PPL trends, indicating further convergence potential, while Evo has already reached its optimal performance. }
	\label{fig:gene_scaling}
\end{figure}    

As illustrated in Figure~\ref{fig:gene_scaling}, our experiments revealed a compelling trend in the performance comparison between BigBang-Proton of 1.5B parameters and Evo of 7B parameters under compute-optimal conditions. At lower compute budgets ($\text{FLOPs} < 3 \times 10^{19}$), BigBang-Proton initially exhibited higher perplexity (PPL) compared to the Evo model, suggesting that with limited computational resources, Evo's larger capacity may have an advantage in capturing simpler patterns efficiently. However, at higher compute budgets ($\text{FLOPs} \geq 3 \times 10^{19}$), BigBang-Proton demonstrated better scaling behavior and significant performance improvements. BigBang-Proton achieved a PPL of 2.8 at $7 \times 10^{19}$ FLOPs despite being trained on only 110.1 billion OpenGenome tokens (27.3B during pretraining plus 82.8B during downstream fine-tuning), which is just one-third of Evo's 300 billion token training budget. In contrast, Evo's compute-optimal training point is actually at 270 billion tokens, beyond which it shows saturated performance with a PPL of 3.1 at the same compute budget. The key insight is that BigBang-Proton 1.5B continues to exhibit a decreasing PPL trend, indicating it has not yet reached its performance plateau, while Evo 7B has already saturated and cannot improve further. This enormous difference with BigBang-Proton outperforming Evo despite using significantly less training data and smaller model size, strongly demonstrates the architectural superiority of BigBang-Proton in genomic sequence modeling. 

\begin{figure}[H]
	\centering
	\includegraphics[width=1\textwidth]{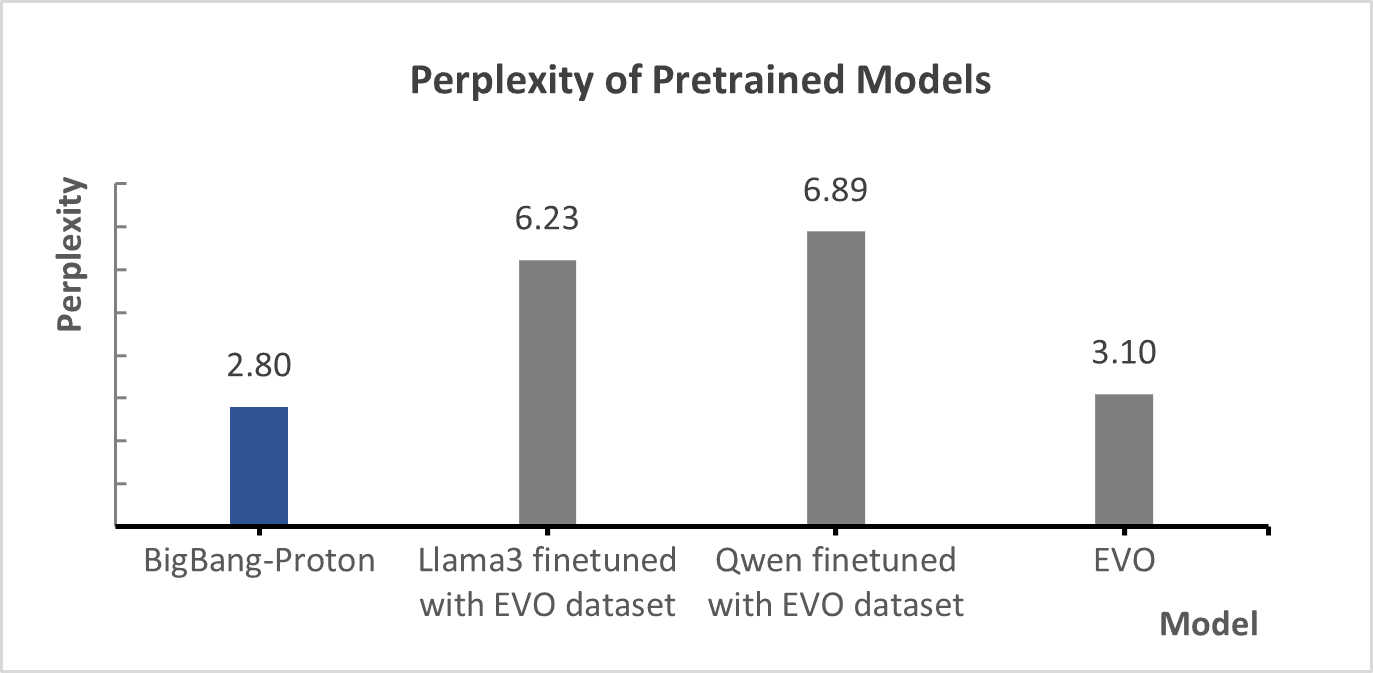} 
	\caption{Comparison of evaluation perplexity between BigBang-Proton 1.5B, Evo 7B, Llama3 1.3B, and Qwen2.5 1.5B. The latter two were fine-tuned on 3B OpenGenome tokens.  Llama3 and Qwen2.5 show PPLs of 6.23 and 6.89, respectively, which are much higher than those of Evo and BigBang-Proton, indicating their failure to learn the distribution patterns of genomic sequences.}
	\label{fig:llm_fintuned}
\end{figure} 

As shown in Figure~\ref{fig:llm_fintuned}, mainstream LLMs like Llama3 with 1.2B parameters achieved PPL of 6.23 and Qwen2.5 with 1.5B parameters achieved PPL of 6.89 after being fine-tuned with LoRA on 3B openGenome tokens. This result shows significantly higher perplexities than BigBang-Proton and Evo, highlighting their limitations in genomic sequence modeling. The large performance gap suggests that the BPE tokenizers used in mainstream LLMs fundamentally disrupt the understanding of DNA's core nucleotide vocabulary of A, G, T, and C. 

We further evaluate next-base prediction accuracy with context length $L=512$ base pairs on BigBang-Proton and other mainstream foundational LLMs including DeepSeek-R1, ChatGPT-o1, Claude-3.5 Sonnet and Kimi. The prediction target is as follows: 

\begin{equation}
P(b_{t+1} | b_{t-L:t}) \quad \text{where} \quad b \in \{\text{A},\text{T},\text{G},\text{C}\}
\end{equation}

\begin{figure}[H]
    \centering
    \includegraphics[width=1\textwidth]{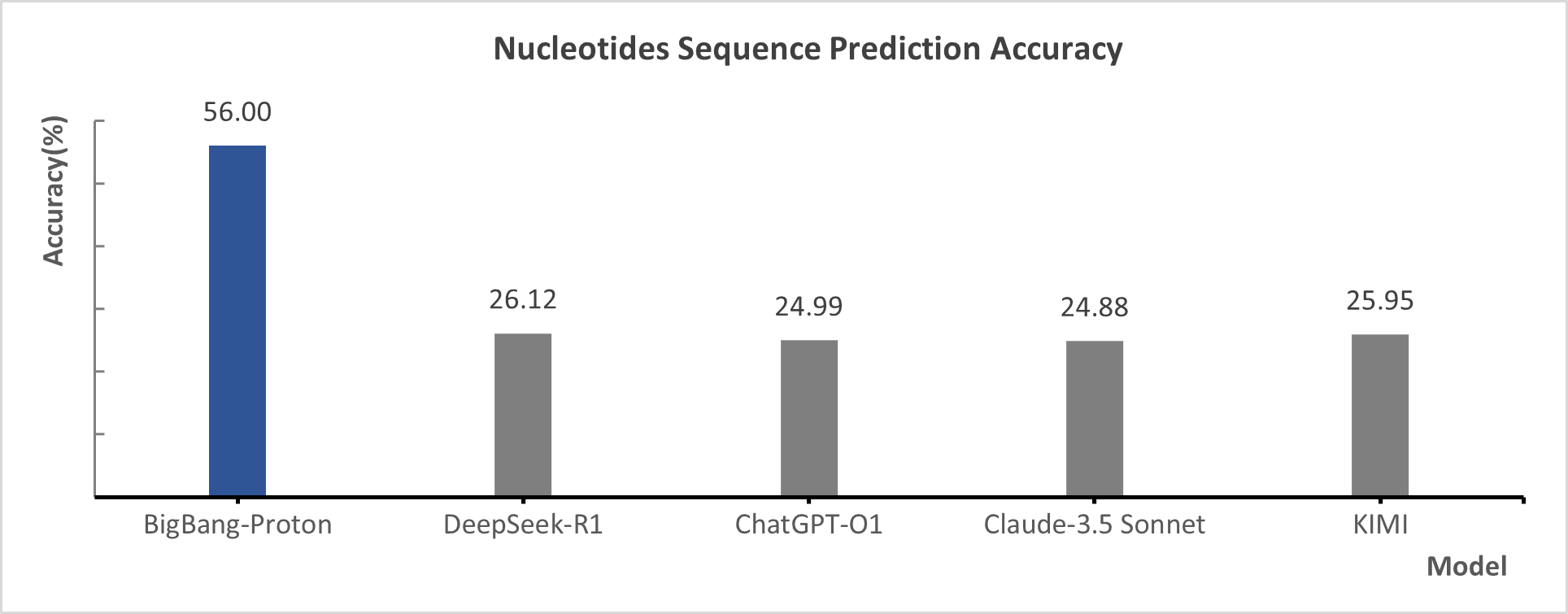}
    \caption{Comparison of accuracy among different models on the gene sequence generation task in a zero-shot setup. The figure shows that BigBang-Proton achieves 56\% accuracy, while all mainstream general-purpose LLMs perform at near-random levels (25\% expected by chance when selecting one from four options).}
    \label{fig:model_comparison_DNA_seq}
\end{figure}

The baselines were evaluated under few-shot settings. Each model was tested under the same few-shot conditions, with identical evaluation datasets. The results, as illustrated in Figure~\ref{fig:model_comparison_DNA_seq}, demonstrate that BigBang-Proton consistently outperforms the baseline models with 56\% accuracy compared to 26.1\%, 25.0\%, 24.88\%, and 25.95\% for DeepSeek-R1, ChatGPT-o1, Claude-3.5 Sonnet, and Kimi, respectively. All general-purpose LLMs perform at chance level (25$\pm$1\%), indicating that these models have fundamentally failed to learn the complex statistical patterns, biological rules, and positional dependencies that govern genomic sequences. These results correspond to the DNA sequence completion examples shown in Appendix~\ref{append:biology}. Despite their vast training on diverse text corpora, general-purpose LLMs lack the specialized inductive biases, appropriate tokenization schemes, and sufficient exposure to genomic data necessary to capture the intricate relationships between nucleotides that extend beyond simple character-level memorization. The near-random performance demonstrates that general-purpose LLMs cannot effectively transfer their language understanding capabilities to the highly specialized domain of genomics, where evolutionary constraints, structural motifs, regulatory elements, and long-range dependencies create pattern complexities that are fundamentally different from natural language structures. 
\paragraph{Predicting mutational effects on protein function} Beyond evaluating perplexity, we further assessed the model's zero-shot predictive capabilities on biologically relevant downstream tasks, with predicting the effects of mutations on protein function being one of the key tasks. Our method uses experimentally measured functional activity or fitness scores as targets for supervised finetuning. This approach allows the model to explicitly learn the relationship between nucleotide mutations and their functional consequences.

To evaluate the model's performance, we employed deep mutational scanning (DMS) studies, which systematically introduce mutations into protein coding sequences and experimentally quantify their impact on functional activity through fitness metrics~\cite{pmlr-v162-notin22a}. Instead of relying on amino acid sequences, we adapted this task to nucleotide sequences by using the wild-type coding sequence and the specific nucleotide mutations reported in the original DMS studies.  We used nucleotide sequences from \textit{E. coli} DMS studies. For Evo, datasets from six studies were used to compile nucleotide information, including a $\beta$-lactamase DMS by \citeay{Firnberg2014}, a $\beta$-lactamase DMS by \citeay{Jacquier2013}, a CcdB DMS by \citeay{Adkar2012}, a multi-protein thermostability dataset by \citeay{Tsuboyama2023}, an IF-1 DMS by \citeay{Kelsic2016}, and an Rnc DMS by \citeay{Weeks2023}. In our experiment, we specifically utilized IF-1 DMS by \citeay{Kelsic2016}. The experimentally measured fitness scores served as the ground truth for training and evaluating the model.

\begin{figure}[H]
		\centering
		\includegraphics[width=1\textwidth]{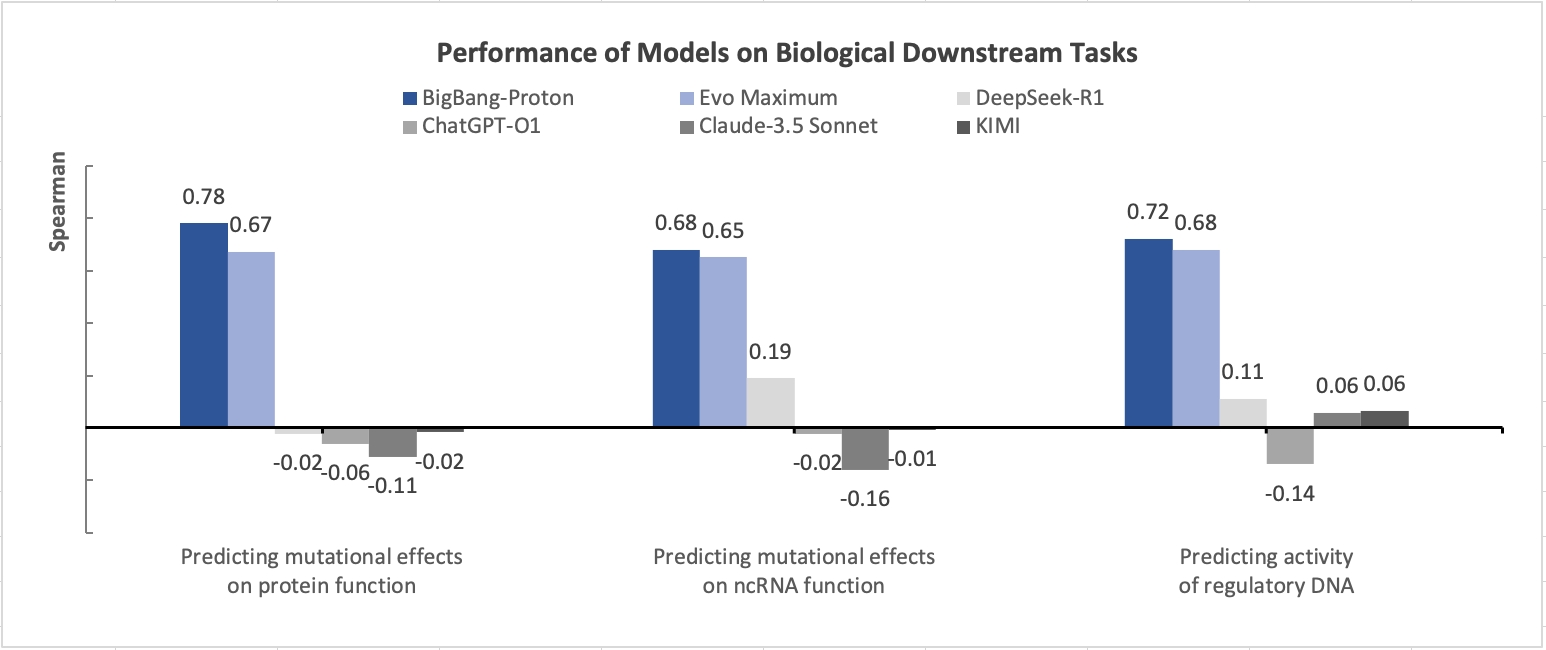} 
		\caption{The figure presents the performance of BigBang-Proton, Evo, and mainstream LLMs on function fitness predictions: (A) predicting mutational effects on protein function, (B) predicting mutational effects on ncRNA function, and (C) predicting the activity of regulatory DNA. BigBang-Proton surpasses Evo in all three tasks, with margins ranging from $0.03$ to $0.11$ in Spearman correlation coefficient. While general LLMs predominantly show negative and low correlations, indicating their inability to capture patterns in genome sequences related to function fitness. }
		\label{fig:biologically relevant downstream tasks}
\end{figure}

Our results (Figure \ref{fig:biologically relevant downstream tasks}A) demonstrate that the model can effectively predict the functional consequences of mutations, achieving a strong Spearman correlation of 0.78546 (p-value: 4.94e-41) between the model's predictions and the experimental fitness scores. This significantly outperforms the best-performing baseline model Evo, which achieved a maximum Spearman correlation of 0.67 and average of 0.45 in six datasets. Additionally, we compared our model against several other advanced models using zero-shot testing on the same task, DeepSeek R1 achieved a Spearman correlation of  (-0.02), ChatGPT o1 achieved (-0.06), Claude 3.5 Sonnet achieved (-0.11), and KIMI achieved (-0.02). These comparisons highlight the superior performance of our model in predicting mutation effects.
\paragraph{Predicting mutational effects on ncRNA function}
Next, we evaluated the model's ability to predict the functional consequences of mutations in non-coding RNAs (ncRNAs). Including tRNAs (protein synthesis adaptors), rRNAs (ribosome structural components), and ribozymes (catalytic RNAs), ncRNAs perform essential cellular functions through their sequence-dependent structures, and their functions are critically affected by nucleotide mutations that can alter folding, stability, and catalytic activity. Similar to the approach used for protein-coding sequences, we collected ncRNA deep mutational scanning (DMS) datasets and used experimentally measured fitness scores as the ground truth for supervised finetuning. This allowed us to test whether the model could generalize its predictive capabilities to ncRNA sequences, which play critical roles in cellular processes but differ significantly from protein-coding sequences in structure and function. Evo used seven datasets for evaluating models' zero-shot abilities in ncRNA function predictions, including a ribozyme DMS by \citeay{Kobori2015HighthroughputAA}, a ribozyme DMS by \citeay{Andreasson2020}, a tRNA DMS by \citeay{Domingo2018}, a tRNA DMS by \citeay{Guy2014}, a ribozyme DMS by \citeay{Hayden2011}, a ribozyme DMS by \citeay{Pitt2010}, and a rRNA mutagenesis study by \citeay{Zhang2009}. We used ribozyme DMS by Kobori et al. (2015) for our supervised finetuning. 

Our results (Figure \ref{fig:biologically relevant downstream tasks}B) demonstrate that the model achieves strong performance in predicting the functional effects of mutations in non-coding RNAs (ncRNAs). On the \citeay{Kobori2015HighthroughputAA} DMS dataset, which measures the effects of mutations on ncRNA function, the model achieved a Spearman correlation coefficient of 0.68, significantly outperforming the best-reported result from the Evo model, which achieved a maximum Spearman correlation of 0.65 and average of 0.25 in predictions of seven datasets. This improvement highlights the model's enhanced ability to infer the functional impact of mutations in ncRNA sequences.

To further validate our findings, we conducted zero-shot testing on the same task using mainstream general-purpose LLMs, with results of DeepSeek R1 $0.19$, ChatGPT o1 $-0.02$, Claude 3.5 Sonnet $-0.16$, and KIMI $-0.01$. The results indicate that these models obtained Spearman correlation coefficients that were close to zero when tested on the DMS dataset. The results underscore the unique capability of our model in handling ncRNA mutation prediction tasks, demonstrating its robustness and effectiveness in this domain.

\paragraph{Predicting gene expression from regulatory DNA}
Given that our model's training data also includes prokaryotic regulatory DNA sequences, we further investigated whether the model has learned meaningful representations that can be applied to regulatory DNA tasks. Specifically, we focused on predicting gene expression from promoter sequences. Similar to our approach for protein-coding and ncRNA tasks, we used experimentally measured activity values as targets for supervised finetuning, ensuring that the model's predictions are directly tied to biologically relevant outcomes.

Promoters are critical regulatory DNA elements that control the initiation of gene transcription. We evaluated the model's ability to predict gene expression levels based on promoter sequences. By leveraging datasets that experimentally measure the effects of mutations in promoter regions on gene expression, we trained the model to predict these expression levels directly. Using a dataset reported by~\citeay{Fleur2021.09.01.458561}, the model achieved a Spearman correlation of 0.72 between its predictions and the experimentally measured gene expression levels(figure \ref{fig:biologically relevant downstream tasks}C). This performance significantly outperforms the best-reported result from the Evo, which achieved a Spearman correlation of 0.68 on the same task. The improvement highlights the model's enhanced ability to capture the regulatory logic embedded in promoter sequences and generalize to predict functional outcomes. 

To further validate our findings, we conducted zero-shot testing on the same task using mainstream general-purpose LLMs, with results of DeepSeek R1 $0.11$, ChatGPT o1 $-0.14$, Claude 3.5 Sonnet $0.06$, and KIMI $0.06$. The results indicate that these models obtained Spearman correlation coefficients that were close to zero by prompts when tested on the promoter sequence dataset. The near-zero correlation values (0.11 and 0.06) suggest that these models are essentially making random predictions with no meaningful relationship to the actual gene expression levels, performing barely above the chance level expected from random guessing. ChatGPT o1's negative correlation coefficient of -0.14, indicates that its predictions are not only incorrect but inversely correlated with true expression levels, suggesting it learns spurious patterns or makes systematic errors worse than random guessing due to inappropriate biases or fundamental misunderstandings of promoter biology.

BigBang-Proton Consistently Outperforms Evo Across Molecular Scale Biology Tasks. The comprehensive evaluation across three distinct molecular scale biology tasks demonstrates BigBang-Proton's superior performance compared to Evo, with consistent and significant improvements in all domains. In predicting mutational effects on protein function, BigBang-Proton achieves a Spearman correlation of 0.78546 versus Evo's 0.67, representing a 17\% relative improvement. For ncRNA mutation prediction, BigBang-Proton attains a correlation of 0.68 compared to Evo's 0.65, showing a 5\% enhancement. In regulatory DNA activity prediction, the model achieves 0.72 versus Evo's 0.68, demonstrating an 6\% improvement. These results establish BigBang-Proton's comprehensive superiority across diverse molecular biology applications, indicating that its architectural innovations and training paradigm enable more effective learning of sequence-function relationships than Evo's approach. Critically, these performance gains align perfectly with the scaling law analysis based on perplexity metrics, where BigBang-Proton demonstrates consistently lower PPL values across increasing compute budgets. The correlation between PPL improvements and downstream task performance validates that the model's enhanced ability to model genomic sequence distributions directly translates to better biological understanding and predictive accuracy. 

General-purpose LLMs Fail to Learn Molecular Scale Biological Tasks. The consistent near-zero and negative correlation results across all three molecular scale biology tasks reveal the fundamental inability of general-purpose LLMs to understand biological sequence-function relationships. Across protein mutation prediction, ncRNA mutation prediction, and regulatory DNA activity prediction, mainstream models including DeepSeek R1, ChatGPT o1, Claude 3.5 Sonnet, and KIMI predominantly achieve correlation coefficients close to zero (0.11, 0.06, 0.19, -0.02, -0.01, 0.06), indicating performance equivalent to random guessing and demonstrating that these models fail to capture the complex statistical patterns and biological rules governing molecular function. More concerning are the systematic negative correlations observed in ChatGPT o1 (-0.14 for promoters, -0.06 for protein mutations, -0.02 for ncRNA), which indicate that this model not only fails to learn meaningful biological patterns but actually develops inverse relationships between sequence features and functional outcomes, suggesting the presence of inappropriate biases or fundamental misunderstandings of molecular biology principles. These results collectively demonstrate that general-purpose LLMs, despite their success in natural language tasks, lack the specialized inductive biases, appropriate tokenization schemes, and sufficient domain-specific training necessary to transfer their capabilities to the highly specialized and quantitatively precise domain of molecular biology, where accurate sequence-function prediction requires deep understanding of biochemical constraints, evolutionary pressures, and structural-functional relationships that are fundamentally different from linguistic patterns.

\paragraph{Why does BigBang-Proton Learn Genomes better?} 

The comprehensive experimental results demonstrate that BigBang-Proton significantly outperforms Evo in genomic sequence modeling. The following fundamental architectural and training innovations potentially explain why BigBang-Proton can learn DNA sequences better than Evo. First, Theory-Experiment Learning combined with Binary Patch Encoding provides fundamental advantages over Evo's approach. This framework integrates theoretical knowledge from scientific literature with experimental genomic sequences {A, G, T, C}. While both methods use character-level tokenization preserving individual nucleotides as UTF-8 indices (A=65, T=84, G=71, C=67), the key difference lies in contextual learning. Evo treats these as isolated symbols relying on brute-force pattern recognition, while BigBang-Proton embeds them within biological context from scientific texts, enabling nucleotides to carry their true biological meanings as Adenine, Thymine, Cytosine, and Guanine. The model learns biological prior knowledge such as associations with biochemical properties, functional roles, and evolutionary constraints, developing rich inductive biases that guide sequence pattern learning. This explains BigBang-Proton's superior performance with lower perplexity (2.8 vs 3.1 at $7 \times 10^{19}$ FLOPs) and higher Spearman correlations (0.785 vs 0.67 for protein mutations, 0.68 vs 0.65 for ncRNA, 0.72 vs 0. Second, cross-discipline transfer learning enables generalization between structures, a key hypothesis behind BigBang-Proton's performance advantage over single-domain model Evo. BigBang-Proton is trained on diverse data including arithmetic operations, particle jets, material crystals, sensors, and stock prices. Physics principles like quantum mechanics govern both material crystals and biological macromolecules, suggesting shared isomorphic properties that enable transfer learning across domains. Third, Monte Carlo Attention enables genome-scale context processing through theoretically unlimited extension. While Evo's 131k token limit constrains long-range genomic dependency capture, BigBang-Proton's Monte Carlo Attention can process entire genome-scale sequences through inter-patch delegation, crucial for understanding complex phenomena like long-range regulatory interactions and chromosomal organization spanning thousands to billions of base pairs.

General-purpose LLMs fail to learn genome sequences due to BPE tokenization's fundamental incompatibility with {A G T C} sequences, as subword merging destroys single-nucleotide resolution and creates semantic ambiguity; therefore, they exclude DNA sequencing data in pretraining, leading to loss of biological inductive biases necessary for capturing sequences and function patterns.

\subsection{Scientific Multi-task Learning: Design and Significance}

Arithmetic operations are central to scientific multi-task learning because experimental results are predominantly numerical. BigBang-Proton incorporates particle jets, material crystals, genome sequences, and water quality, representing scales from quarks to Earth systems. Diverse datasets align in representation space without conflict during pretraining, showing possible transfer in latent space with similarities especially between adjacent scales. Quantum mechanics governs atom interactions in materials and biological systems, suggesting DNA and proteins reflect quantum effects, leading to distributional homogeneity. Protons and neutrons form atoms with electrons, while crystal formation energy stems from electron interactions. Quark decay distributions likely share similarities with material structures, aiding formation energy calculations. Water quality represents a complex system influenced by multiple factors. Material and biological structures may exhibit similarities with water quality data. Number theory reveals intrinsic numerical structures in the universe. Arithmetic logic unit simulation could enhance learning across all scales.

Pretraining a foundational model on datasets spanning all scales, structures, and disciplines would provide deeper insights into hidden analogies in physical structures. This approach merges reductionism and emergentism, treating the universe as a unified entity to explore fundamental laws.

\section{Related Work}

\textbf{LLMs on real-world science} So far there are limited publications about using mainstream LLMs of both basic version(post-trained by supervised finetuning and RLHF) or reasoning version (post-trained by long-horizon chain-of-thought) to tackle real-world scientific problems. The dominant use of LLMs for scientific research happens in literature searching, writing and codes generation. In a case on theoretical condensed matter, \cite{yin2025exactsolutionfrustratedpotts} prompted OpenAI’s o3-mini-high to derive an exact equation for the critical temperature of UNPC in Ising models and, despite initial errors, discovered a symmetry-based block diagonalization that reduced the 9×9 transfer matrix of the 1D J1-J2 three-state Potts model to a solvable 2×2 form. DeepMind developed FunSearch \cite{romera2024mathematical}, a method that uses LLMs to generate creative solutions in the form of computer code, leading to new mathematical discoveries for the cap set problem and improved algorithms for the bin-packing problem, marking the first time LLMs have been used to solve challenging open problems in science and mathematics at the year 2023 of the publication. These cases all rely on the language generation capability of LLMs. 

\textbf{Scientific agents} Expanding the capability of LLMs to complete scientific tasks, multiple agent-based systems have been developed. ~\citeay{ghareeb2025robinmultiagentautomatingscientific} released \textit{Robin}, a multi-agent system for automating scientific discovery, including \textit{Crow} and \textit{Falcon} as literature review agents for generating hypotheses and \textit{Finch} as an experimental analysis agent which was implemented on \textit{Aviary}~\cite{narayanan2024aviarytraininglanguageagents}, a framework of language agents for scientific tasks, to interact with data analysis tools. Finch interacts with \texttt{edit\_cell}, a specialized data analysis tool designed for bioinformatics workflows to agents to modify and execute code cells in a Jupyter notebook, and \texttt{submit\_answer}, a tool finalized the agent's analysis and submits conclusions such as figures, tables and interpretations.These tools enabled by Aviary environment are typical domain data analysis tools independently developed for specialized application in particular scientific fields. ~\citeay{qiu2025biomarsmultiagentroboticautonomous} developed \textit{BioMARS}, an end-to-end scientific multi-agent system, integrating LLMs and VLMs (vision-language models) to automate experiments in biology, consisting of biologist agent based on LLMs and RAG to analyze literature and synthesize protocols, technician agent transforming language protocols to robotics actions, and inspector agent based on ViT (Vision Transformer) to supervise the experiment processes. \citeay{gottweis2025aicoscientist} published AI co-scientist, a multi-agent system built on Gemini 2.0\cite{gemini2024update} that functions as a collaborative tool for scientists, generating novel hypotheses and research proposals through specialized agents (e.g., Generation, Reflection, Ranking, Evolution, Meta-review) inspired by the scientific method. These agents iteratively refine hypotheses using automated feedback loops, tools like web searches, and test-time compute scaling for advanced reasoning. ~\citeay{su2024two} published \textit{VirSci} is an LLM-based multi-agent system designed to mimic collaborative scientific research by organizing teams of agents to iteratively generate, evaluate, and refine hypotheses using feedback loops and domain-specific tools.~\citeay{lu2024aiscientistfullyautomated}, \citeay{yamada2025aiscientistv2workshoplevelautomated}, \citeay{sakana} from Sakana AI published \textit{AI Scientist}, an end-to-end agentic system that autonomously formulates scientific hypotheses, designs and executes experiments, analyzes data, and authors manuscripts—successfully generating several peer-review-accepted workshop papers. ~\citeay{giglou2024llms4synthesis}, \citeay{agarwal2025litllm}, \citeay{wang2024usinggpt4}, \citeay{huang2025automated}, \citeay{boiko2023autonomous}, \citeay{ifargan2024autonomous}, \citeay{wang2025txgemma} published researches focusing on scientific hypothesis generation based on LLMs or agents, particularly in the fields of biology and chemistry. However, we find that all these AI scientist agent systems overwhelmingly rely on the language understanding and generation capabilities. When it comes to data analysis, the systems have to interact with the outside specialized tools. It is an obvious disadvantage that the current mainstream LLMs have not been pretrained on experimental data and therefore are unable to process the experimental data analysis within its knowledge. It also leads to the prevalent feedback from the scientific community that most of the ideas generated by LLMs are not profound~\cite{si2024llmsgeneratenovelresearch}. 

\textbf{Domain-specific LLMs} Auto-regressive domain foundational models or finetuned models in material science, biology, CFD have been published. In material science and chemistry, \citeay{xia2025naturelanguagemodeldeciphering} and \citeay{huang2023materialsinformaticstransformerlanguage} tried to train a foundational language model for biology and material from scratch. ~\citeay{tang2025matterchatmultimodalllmmaterial}, \citeay{xie2025darwin15largelanguage}, \citeay{gruver2024finetunedlanguagemodelsgenerate}, \citeay{rubungo2023llmproppredictingphysicalelectronic} finetuned material and chemistry datasets on foundational LLMs and achieved better performance than the original model in this particular domains. \citeay{bran2023chemcrowaugmentinglargelanguagemodels} tried to build agent on top of GPT-4 to use tools in chemistry researches. 

Foundational models in biology have shown remarkable capabilities in modeling biological systems at scale. \citeay{Brixi2025.02.18.638918} introduced Evo 2, a biological foundation model trained on 9.3 trillion DNA, RNA, and protein sequences, capable of predicting functional genetic variation and generating genome-scale sequences with unprecedented resolution and coherence. Similarly, \citeay{lin2023evolutionaryscale} demonstrated that evolutionary-scale protein language models can infer full atomic-level structures directly from primary sequences using a 15-billion-parameter architecture (ESMFold), achieving high-resolution predictions at scale. More recently, \cite{Bhatnagar2025.04.16.649055} published ProGen3 , a 46-billion-parameter sparse autoregressive model trained on over 1.5 trillion amino acid tokens, which significantly improves the ability to generate viable proteins across diverse families and aligns better with experimental data for improved fitness prediction.However, despite their impressive performance, these foundational models are still fundamentally based on autoregressive sequence modeling and rely heavily on large-scale biological sequence or structural data , such as DNA, RNA, and protein sequences or atomic-level structures. While this enables them to learn statistical patterns and correlations from vast biological datasets, it also limits their understanding to what is present in the training data — primarily sequence-to-structure relationships — without explicitly incorporating the underlying physical and chemical principles that govern molecular interactions, thermodynamics, kinetics, or cellular environments.In other words, current foundational models in biology lack integration of first-principles knowledge from physics and chemistry, such as energy minimization, electrostatic interactions, hydrogen bonding, or reaction dynamics, which are essential for a deeper mechanistic understanding of biological processes. This restricts their generalization capability beyond known biological patterns and makes them prone to overfitting or producing biologically plausible but physically unrealistic sequences or structures. 

\textbf{World model} \citeay{cosmos} introduced Cosmos-Reason1 , a Physical AI system that builds upon a hierarchical ontology of physical common sense encompassing Space, Time, and Fundamental Physics to enable grounded reasoning about the real world through multimodal perception and long chain-of-thought processes. \citeay{yang2025thinkingspacemultimodallarge} released VSI-Bench for evaluating multimodal large language models (MLLMs) on visual-spatial intelligence. \citeay{assran2025vjepa2selfsupervisedvideo} introduced V-JEPA 2 , a self-supervised model trained on large-scale video data that enables physical world understanding, prediction, and planning without task-specific supervision. Despite these advances, recent works on world modeling remain fundamentally limited by their reliance on visual and multimodal perception. They fall short of constructing a comprehensive understanding of the physical world due to three key limitations. First, these models lack quantitative measurement integration, learning from qualitative observations such as videos, images, or natural language descriptions without incorporating precise numerical data like temperature, pressure, velocity, or force that are essential for accurate physical reasoning. Second, they cannot learn physical laws without quantitative grounding, as access to measurable, structured physical quantities is necessary to infer underlying physics principles such as Newtonian mechanics, thermodynamics, and electromagnetism, which are inherently mathematical and require data-driven learning. Third, simulating "what is seen" is not equivalent to simulating reality, as these systems can mimic visual patterns such as wave motion or object interactions but lack mechanistic understanding, resulting in simulations of appearances rather than the causal dynamics behind them. For example, even if a model can generate realistic wave movement in video simulations, it does not understand the fluid dynamics equations that govern such behavior. To build true world models that generalize and predict real-world phenomena accurately, future systems must go beyond passive observation and integrate quantitative physical data , first-principles physics law represented by language and numeric descriptions , and structure learning over time, space, energy and matter .

AI agents have increasingly leveraged reinforcement learning to demonstrate long-horizon chain-of-thought (COT) capabilities, as seen in ChatGPT o1 and DeepSeek R1. \citeay{silver2025welcome} argue that future AI progress lies in experience-based learning, enabling superhuman performance in mathematics, coding, and scientific discovery. However, the concept of agency remains ill-defined in current research. Our findings indicate that long-horizon COT alone is insufficient for solving data-driven scientific problems and may introduce bias that undermines result reliability. True understanding of the material world must be rooted in structure-aware representation learning during pretraining, especially for modeling multi-scale and multi-structure physical or biological systems. Consequently, pursuing AGI through reinforcement learning alone may fall short without grounding in a coherent world model. As further shown by \citeay{richens2025generalagentsneedworld}, general agents capable of multi-step, goal-directed tasks must first learn an internal model of the environment’s dynamics and constraints. 

\textbf{Embodied intelligence} Vision-language-action(VLA) models by end-to-end pretraining have been adopted as the mainstream methodology for robotics\cite{sapkota2025visionlanguageactionmodelsconceptsprogress}.
\citeay{brohan2023rt2visionlanguageactionmodelstransfer}, \citeay{agibot2025worldcolosseo}, and \citeay{helix2025system1system2} introduced generalist vision-language-action (VLA) models—RT-2, GO-1, and Helix—that leverage web-scale data and end-to-end learning to achieve semantic reasoning and high-performance long-horizon robotic control in real-world settings. Current VLA models build world models on web-scale language pretraining, yet they often fail to learn coherent representations of the environment, as demonstrated in \citeay{vafa2024evaluatingworldmodelimplicit}'s navigation experiments. To build a material world foundational model that can accurately map multi-scale realities—including earth systems, countries, cities, factories, and homes—within which embodied intelligent systems operate is critical for the mass application of embodied intelligence. Particularly, RT-2 integrates robotic actions represented as numeric tokens with Internet-scale vision-language data, enabling emergent semantic reasoning and improved generalization in robotic control through end-to-end learning. Our work, BigBang-Proton, excels at large-scale numerical data learning and physical structure modeling, and can seamlessly merge the learning of action control with internal world representation. This integration significantly enhances generalization capabilities and enables more accurate planning and reasoning in real-world environments.

\section{Discussion and Future Work}

The smooth convergence of pretraining on language and specialized scientific multitask datasets indicates that scaling of LLMs can go beyond language to physical world. The capabilities shown by BigBang-Proton to simulate and generate multi-scale physical structures, in which mainstream reasoning LLMs have encountered obvious failures, further suggest that Structure Learning is indispensable constituent for artificial general intelligence (AGI).

\subsection{Path to AGI: Structure Learning Versus Long-Horizon Chain-of-Thought}

\begin{figure}[H]
     \centering
     \includegraphics[width=1\textwidth]{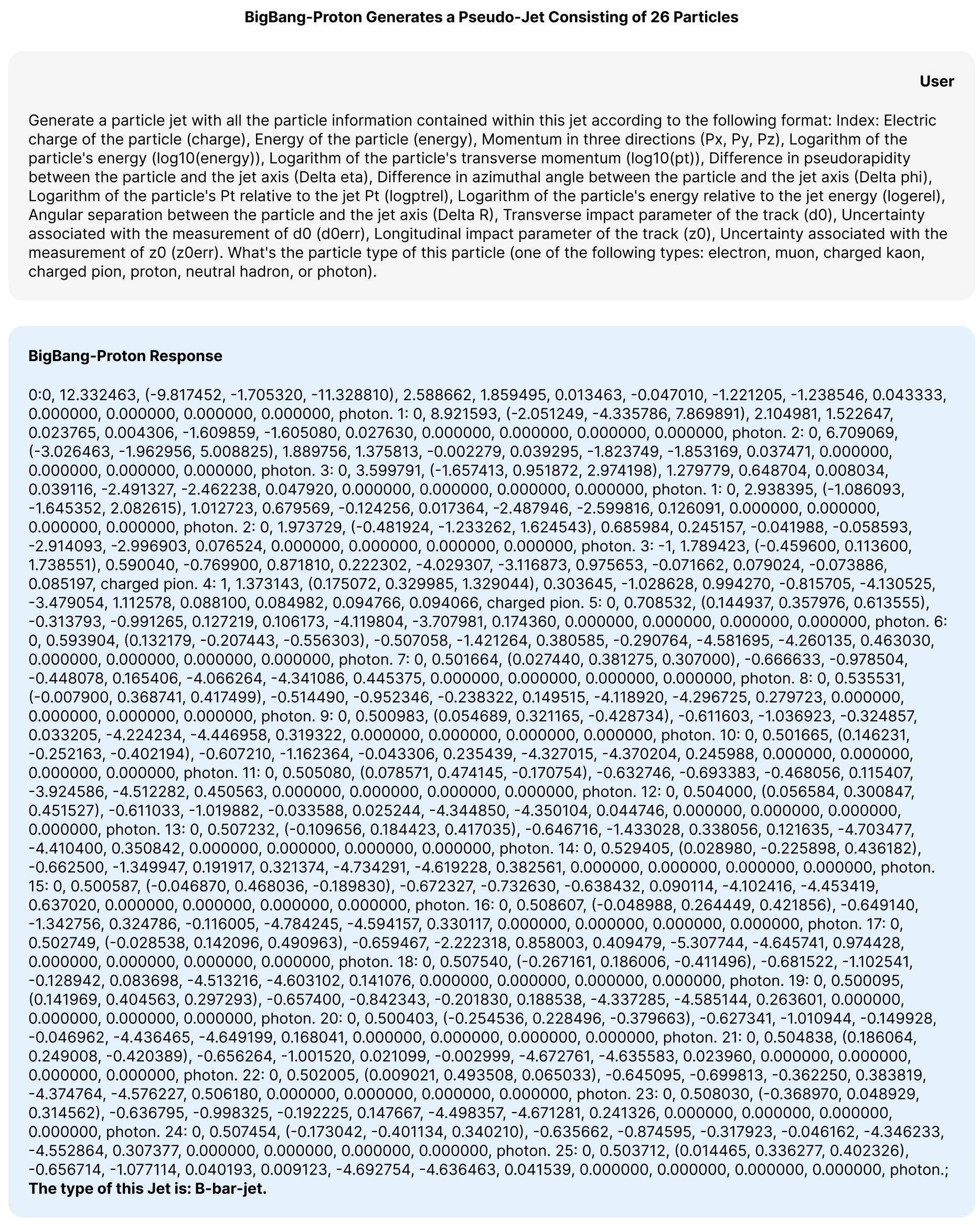} 
     \caption{Prompt BigBang-Proton to generate a quasi-jet consisting of final-state particles from decaying of quarks. BigBang-Proton demonstrates remarkable ability to generate near-real particle data, accurately reproducing critical physics quantities including transverse impact parameter (d0), longitudinal impact parameter (z0), and particle type assignments (electrons, muons, charged pions). The model exhibits 1.24\% discrepancy in momentum conservation ($P_x^2+P_y^2=P_t^2$) and correctly understands that only charged particles can have track impact parameters. These results indicate that BigBang-Proton has learned the underlying physical relationships and dynamics governing jet structure, generation, and interaction, enabling it to generate data that closely resembles real-world observations.}
     \label{fig:jet generation}
\end{figure}

Physical world is made up of hierarchies of material systems defined by physical structures emerged on specified scale\cite{doi:10.1126/science.177.4047.393}. Here we define physical Structure Learning in developing AGI as a basic requirement to understand the material world. Going deep to the fundamental material structure, natural language-based long horizon chain-of-thought approaches experience total failure without forming the capabilities to learn physical structures described by experimental measurement, shown in previous sections. Human language as a representation of human understanding to the world structures, only consist of small fraction of description of the reality. In this case, even infinite COT can not lead to the definite understanding of material structures. Long-horizon COT is an incomplete approach for achieving AGI. As shown in Figure~\ref{fig:jet generation}, BigBang-Proton is capable of generating a quasi-jet consisting of final-state elementary particles from decaying of quarks depicted by 15 dimensions after training on 1 million jet event datasets. This quasi-jet is made up of 24 photons and 2 charge pions from decaying of B quark. Particle features include charge, energy, 3-momenta $(P_x, P_y, P_z)$, $\log_{10}(\text{energy})$, $\log_{10}(p_T)$, angular offsets $(\Delta \eta, \Delta \phi, \Delta R)$, relative scales $(\log p_T^{\text{rel}}, \log E^{\text{rel}})$, and (for charged tracks only) impact parameters $(d_0, z_0)$ with uncertainties. Two thirds of generated photons in the jet have zero mass which is required by physics law 

\begin{figure}[H]
    \centering
    \begin{subfigure}[t]{0.48\textwidth}
        \centering
        \includegraphics[width=\textwidth]{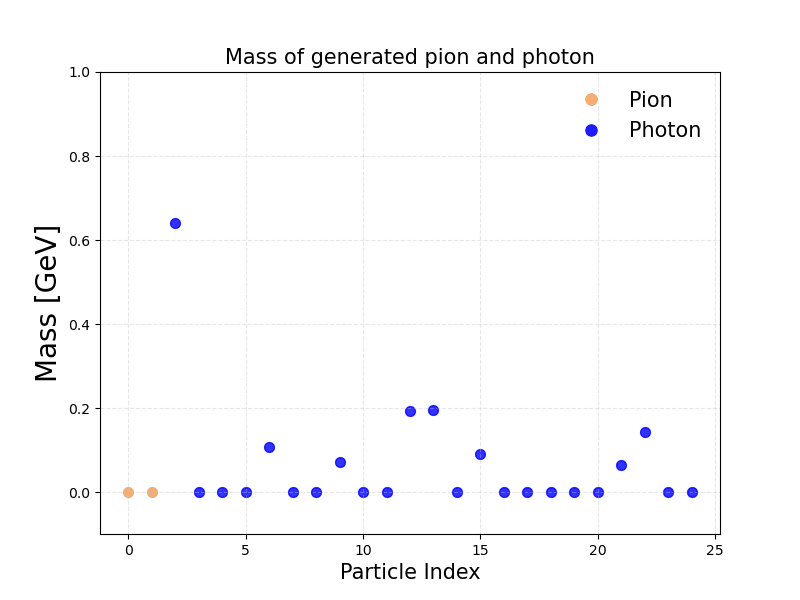}
        \caption{This quasi-jet is made up of 24 photons and 2 charge pions from decaying of b quark, a configuration not typically observed in nature but conforming to fundamental physics constraints. The quasi-jet is generated from the context of physics knowledge and jet data on which BigBang-Proton was trained.}
        \label{fig:quasi jet photon}
    \end{subfigure}
    \hfill
    \begin{subfigure}[t]{0.46\textwidth}
        \centering
        \includegraphics[width=\textwidth]{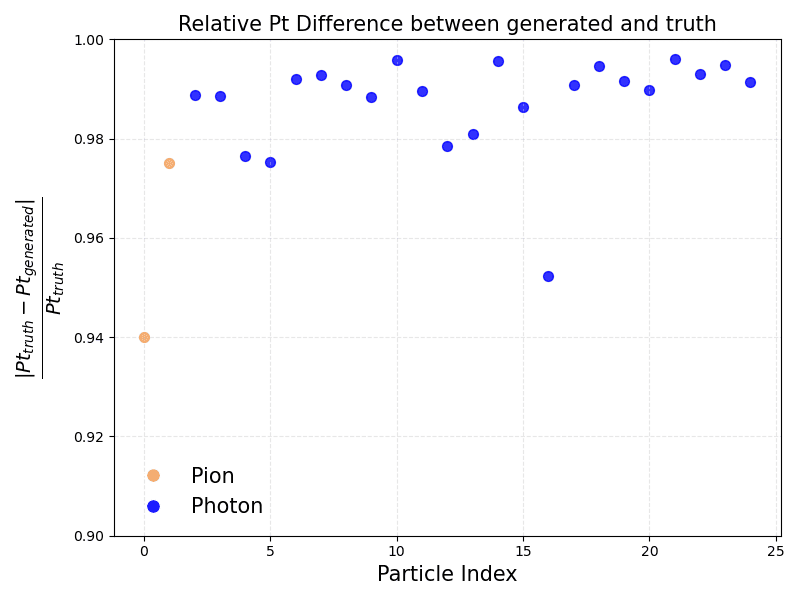}
        \caption{The momentum conservation follows the relation:$P_x^2+P_y^2=P_t^2$, where $P_t$ denotes the total momentum. This figure shows that the generated $P_t$ and $P_x$, $P_y$ has 1.24\% discrepancy.}
        \label{fig:quasi-jet momentum}
    \end{subfigure}
    \caption{A quasi-jet is generated to simulate a real $\bar{b}$-induced jet via quark fragmentation.}
    \label{fig:quasi-jet analysis}
\end{figure}

and the other one third have non-zero mass due to hallucination, as demonstrated in Figure~\ref{fig:quasi jet photon}. The momentum conservation follows the relation:$P_x^2+P_y^2=P_t^2$, where $P_t$ denotes the total momentum. Figure~\ref{fig:quasi-jet momentum} shows that the generated $P_t$ and $P_x$, $P_y$ has 1.24\% discrepancy. BigBang-Proton understands only charged particle can obtain transverse and longitudinal impact parameters of the track. These analysis indicates that model are able to learn the physics and dynamics governing the jet structure, generation and interaction, enabling it to generate data that closely resembles real-world observations. Similarly, BigBang-Proton demonstrates the capabilities to generate crystal structure and DNA sequences by natural language prompts, indicating it has learned these material structures in pretraining. Using Theory-Experiment Learning paradigm, we can integrate language learning with material Structure Learning and use language prompt to drive Structure-based reasoning, and expand the Structure Learning to the whole physical world within a single model. 

\subsection{Platonic Representation, Data Space Manifold and Universe Manifold} 

\citeay{huh2024platonicrepresentationhypothesis} argues that AI models trained on different deep networks and modalities tend to converge to a shared statistical model of reality in their representation spaces. They hypothesize that this convergence arrives at a representation of an idealized reality, termed the Platonic representation , in reference to Plato's Allegory of the Cave \cite{plato1992republic}. This work focuses on studying the distances between data points in vision and language representations, finding that these models attempt to converge to a representation of the underlying reality that generates the observed data. Data and model scaling, along with task generalization, serve as key drivers of this convergence.In the context of scientific multi-task learning and the development of material world foundational models, this hypothesis is supported by transfer learning across scientific domains and the hierarchical organization of the material world based on scales and structures \cite{doi:10.1126/science.177.4047.393}. The material world, or the Universe itself, constitutes the idealized reality, and all scientific laws in physics, chemistry, and biology represent statistical reflections of this reality from specific perspectives. The material world is a single, unified entity, while human efforts to understand it are divided across disciplines due to resource constraints. The Universe originated from the Big Bang 13.7 billion years ago and evolved into its present state \cite{alpher1948origin, Kolb:1990vq}, giving rise to human civilization, from which human languages emerged and were recorded via the internet. Mainstream LLMs, trained on the entirety of internet data, capture only a small portion of the information embedded in the material world. Contrary to the prevailing judgments that foundational LLM pretraining has hit a wall\cite{hardywhite2024scalinglaws, altinuc2025gptscalingwall, ilyas2024pretraining, marcus2025breakingopenai} due to the exhaustion of available internet data, we conclude from our work on BigBang-Proton that the limit of pretraining is ultimately the limit of the universe itself. We further hypothesize that, as seen in \textbf{Hypothesis 1}, given sufficient resources, pretraining on all data that can be collected across the entire history of the Universe and human civilization will allow a single model to converge to a representation rooted in the Big Bang and the fundamental laws governing the origin of the universe, the intersection of information and material, as the material world we inhabit today derives from that singular point. 

\begin{hypothesis}
    \label{hyp1}
        The scaling law of auto-regressive LLMs has not hit the wall. The limit of the scaling of LLMs is the ultimate boundary of the universe. The scaling of LLMs will ultimately converge to fundamental physics laws at Big Bang moment and the intersection of information and material. 
\end{hypothesis}

The intrinsic statistical nature of reality stems from quantum mechanical principles\cite{Dirac1930-DIRTPO}, originating in quantum fluctuations during the Big Bang\cite{Turner_1999, guth2013quantumfluctuationscosmologylead, Blasco_2015, Kuzmichev_2018}. These fluctuations drove hierarchical structure formation across cosmic and material scales, exemplifying the principle that macroscopic phenomena emerge from microscopic interactions governed by quantum mechanics and thermodynamics. Statistical mechanics\cite{pathriaStatisticalMechanics2011} bridges these scales, revealing how probability distributions underpin phase transitions in physical systems, a concept paralleled in linguistic systems. Language, as an information-transmission medium, derives its structure from probabilistic distributions that project reality through combinatorial configurations of dimensions of space, time and energy. Its macroscopic architecture emerges from vast permutations, forming a complex system governed by free energy minimization perspective\cite{friston2010freeenergy}. Language exhibits dynamic structures with phase transitions and emergent properties across its evolutionary trajectory\cite{Macwhinney, harrison2007language}. LLMs computationally approximate these linguistic probability distributions which are refined over millions of years to encode temporal, spatial, and energetic relationships. This study, \citeay{wu2024scalingparticlecollisiondata}, and other recent attempts\cite{xia2025naturelanguagemodeldeciphering, tang2025matterchatmultimodalllmmaterial} show that, when mapped into high-dimensional latent spaces, these distributions closely align with fundamental physical structures such as particle collisions, material crystal lattices, DNA sequences, and water's spatial and temporal patterns. Such alignment enables language-guided scientific computing and advances transfer learning between scientific domains\cite{thrun1998learning, pan2009survey}. \citeay{mirchandani2023largelanguagemodelsgeneral} and \citeay{lu2021pretrainedtransformersuniversalcomputation} concluded that LLMs can work as general pattern machines in highly different data modalities with sequences as input. The language-physics alignment in representation space further suggests that language inherits symmetries from the universe’s physical laws, positioning LLMs not merely as language pattern learner, but also as tools for reconstructing a unified statistical aspects of reality. 

\begin{figure}[H]
    \centering
    \begin{subfigure}[t]{0.48\textwidth}
        \centering
        \includegraphics[width=\textwidth]{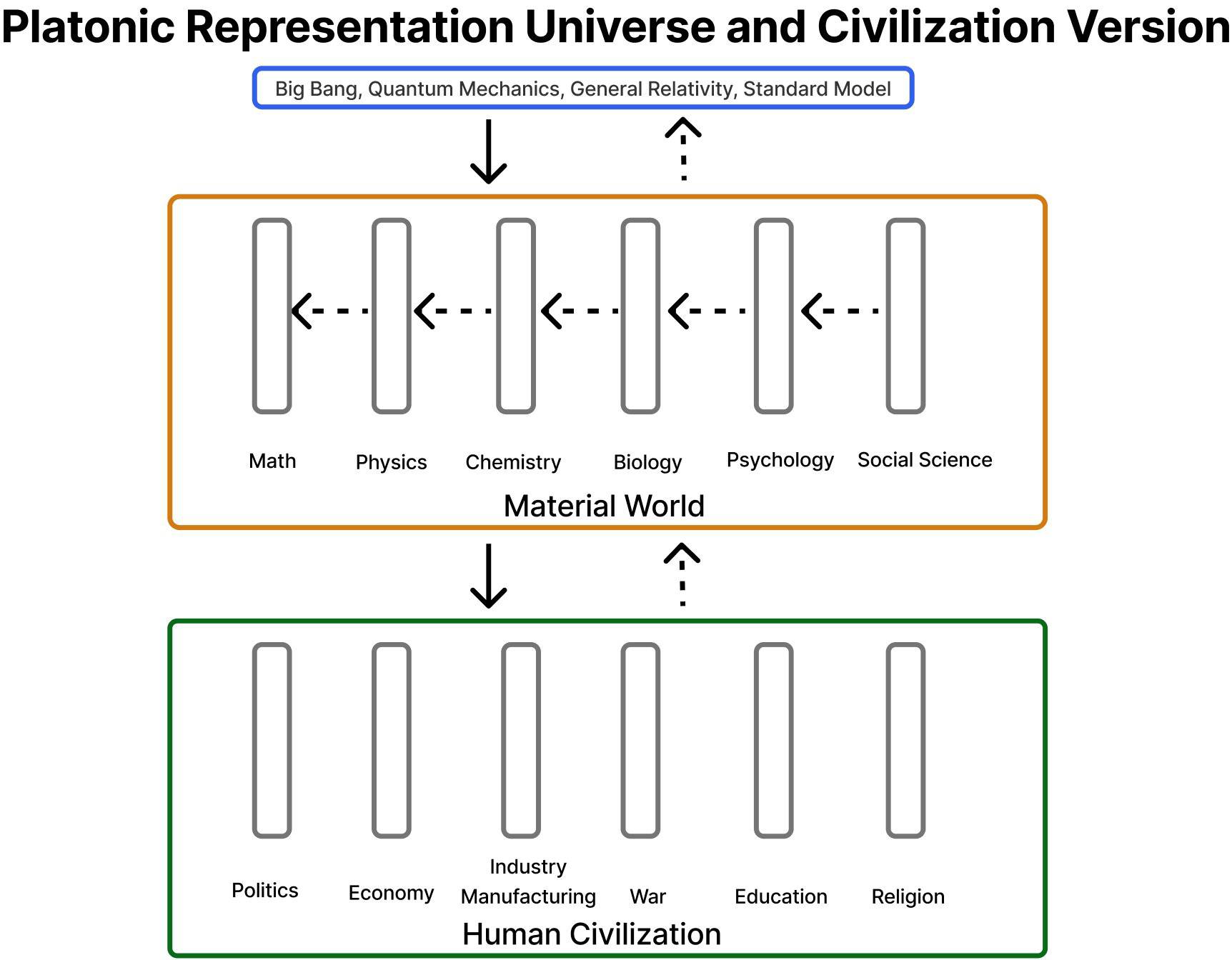}
        \caption{Platonic representation of the universe and civilization converges to basic physics laws. We hypothesize that models trained on complete data generated by human civilization tend to converge to basic natural laws, and models trained on complete data of nature and universe evolution tend to converge to basic physics laws. A single model trained on data generated from the complete history of the universe tends to converge to the Big Bang moment, the fundamental physics law at singularity, and the intersection of information and matter.}
        \label{fig:plantonic}
    \end{subfigure}
    \hfill
    \begin{subfigure}[t]{0.48\textwidth}
        \centering
        \includegraphics[width=\textwidth]{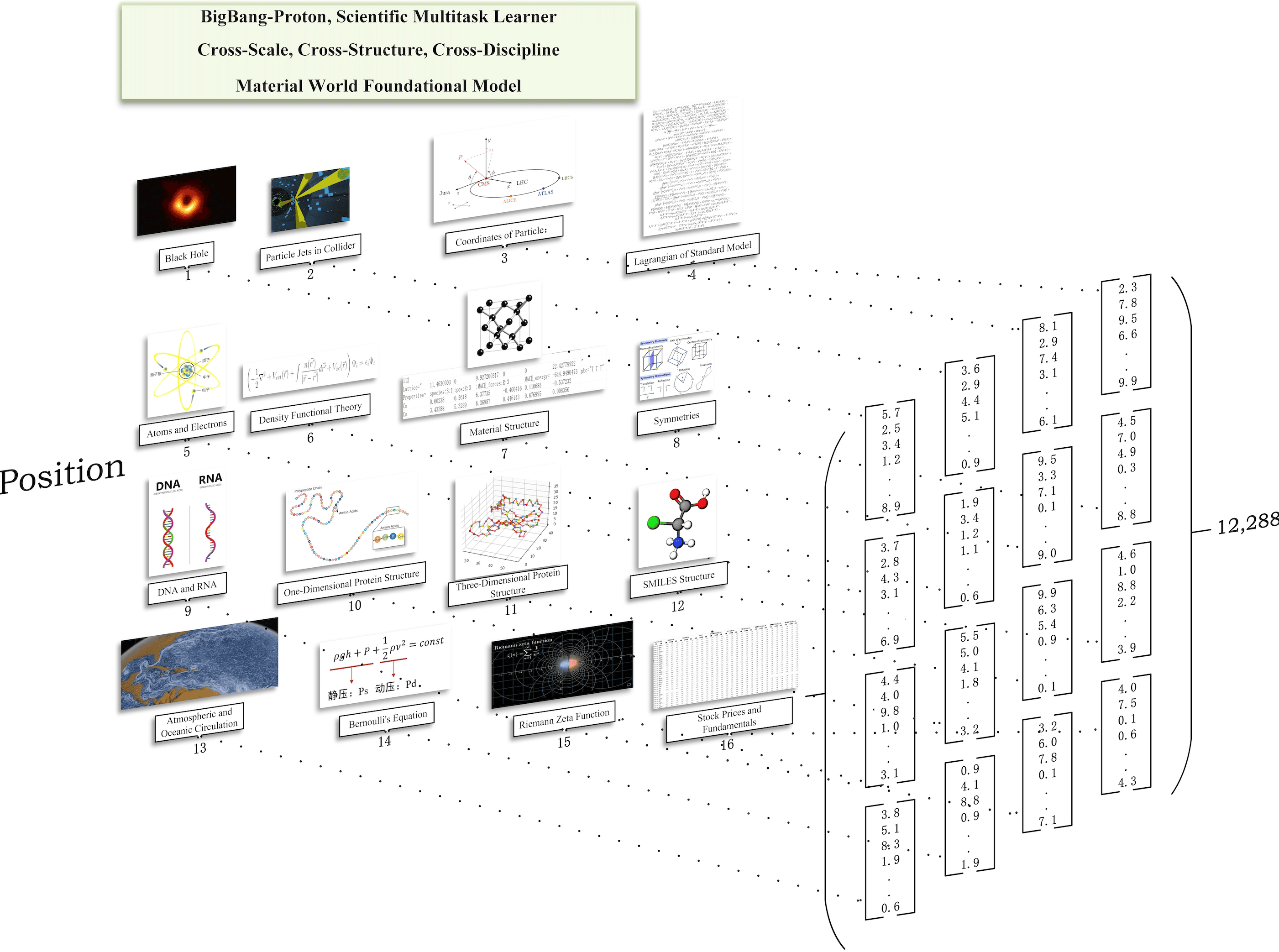}
        \caption{Projecting all scientific domains and natural languages to the same embedding space achieves convergence. The convergence shown in Figure~\ref{fig:training performance} verifies that natural language and seemingly highly different scientific domains share a universal statistical nature, which may root in quantum mechanics and other fundamental physics principles. This insight inspires us to treat the universe as a unified entity when building representations and models.} 
        \label{fig:converge}
    \end{subfigure}
    \caption{Convergence to underlying reality is manifested by the evolution of universe and the history of science.}
    \label{fig:platonic representation}
\end{figure}

Traditional transfer learning in computer vision and NLP utilizes representation and parameter transfers\cite{han2021pretrainedmodelspastpresent, farahani2021concisereviewtransferlearning}. Figure~\ref{fig:converge} shows the scheme that projects all scientific domains knowledge to the same representation space to augment significantly transfer learning between domains with binary patch as encoding method, proposed in our previous work BigBangTransformer-Neutron\cite{wu2024scalingparticlecollisiondata}. Figure~\ref{fig:plantonic} illustrate how different scientific domains converge and transfer knowledge along the one-way hierarchy of science, obeying "more is different" principle\cite{doi:10.1126/science.177.4047.393}. Simulating the entirety of universe is the direct approach to achieve thorough transfer learning across disciplines, scales, and structures in ultrahigh representation data space. 

Manifold hypothesis in deep learning provides further insight into how pretraining on data from the entire universe may converge to fundamental physical laws and enable representation transfer across domains. The hypothesis posits that real-world data, when embedded in a high-dimensional ambient space $\mathbb{R}^D$ concentrate near a much lower-dimensional manifold$\mathcal{M}$ \cite{BengioCourvilleVincent2013, belkin2003laplacian, canas2013learning, genovese2012manifold, dasgupta2008random, donoho2003hessian, hastie1989principal, kambhatla1993fast}, a claim supported by both theoretical investigations \cite{OzakinGray2009, NarayananMitter2010, Cacoullos1966, kegl2000learning, narayanan2009sample, niyogi2008finding, roweis2000nonlinear, smola1999regularized, tenenbaum2000global, weinberger2006unsupervised} and empirical studies \cite{PopeZhuAbdelkaderEtal2021, Carlsson2009}. Manifold learning has also been examined in large language models \cite{modell2025originsrepresentationmanifoldslarge}.According to the union of manifold hypotheses \cite{brown2023verifyingunionmanifoldshypothesis}, low-dimensional structures of interest in Universe-scale pretraining across tasks, disciplines, spatial scales, and physical structures form manifolds and converge to a union of disconnected manifold. This convergence of manifold may correspond to the fundamental Universe manifold, and align with the holographic principle \cite{Bousso_2002, hooft2009dimensionalreductionquantumgravity, ramallo2013introductionadscftcorrespondence, Maldacena_1999} which states that the full physical description of Universe can emerge from its boundary representation. This perspective offers a deeper understanding of the relationship between information entropy, as introduced by Shannon\cite{shannon1948mathematical}, and thermodynamic entropy in physics \cite{bekenstein2007holographic, susskind1995worldashologram}. As \citeay{Vopson2025} proposed that gravity is evidence of data compression of the computational universe, the actual pretraining on the entire universe would reveal the physics mechanism at the intersection of information and material. 

\citeay{hooft2009dimensionalreductionquantumgravity} first postulated that 3+1 spacetime dimensions would reduce to 2+1 at the Planck scale, inspired by the Bekenstein bound, which conjectures that the upper bound of a black hole’s entropy scales with surface area rather than volume \cite{bekenstein1981universalupperbound}. This world-as-a-hologram hypothesis was supported by the anti-de Sitter space and conformal field theory (AdS/CFT) correspondence \cite{Maldacena_1999}. Dimensional reduction is a fundamental quest of manifold learning \cite{meilă2023manifoldlearningwhathow}, which can drive models pretrained on universe-scale data to learn the spacetime structure at the Planck scale and uncover hidden laws at the quantum gravity level, which remain undetected by physics experiments. This spacetime structure learning extends far beyond the current spatial intelligence and world model paradigm \cite{yang2025thinkingspacemultimodallarge, cosmos, jepa}, which is primarily based on image learning.

Considering physical constraints, including the speed of light, degree of freedoms and the Bekenstein bound\cite{Lloyd_2000}, \citeay{PhysRevLett.88.237901} calculated the total information capacity of the universe. The universe contains approximately $10^{80}$ baryon particles and can register no more than $10^{90}$ bits with all energy and matter, and no more than $10^{120}$ floating-point operations for logical operations when gravity is considered. To pretrain a foundational LLM on $10^{80}$ baryons and $10^{90}$ bits, Monte Carlo Attention can provide an equivalent context length of $10^{80}$ by setting the layer number to 40, a capability unachievable by other transformer or alternative architectures. 

\citeay{sutskever2023generalization} introduced data compression and Kolmogorov complexity as the mathematical formalism to explain the generalization of unsupervised learning. We convert information of degree of freedoms of total baryons in universe to a binary string. Let $X =X_{1:N}$ be the binary string encoding all degrees of freedom $d$ including positions, momentum and spin stored in $b$ bits for $M = 10^{80}$ baryons, with $N=M*d*b$ generated through:

$$X = \mathcal{L}(Y, \Pi)$$

where $\mathcal{L}$ = foundational LLM acting as compression algorithm, $Y$ = complete scientific knowledge as constraints based on scales, structures and disciplines, $\Pi$ = initial conditions
Without $Y$, X consisting of totally random baryons approaches maximum entropy\cite{CoverThomas2006, kaltchenko2004algorithmsestimatinginformationdistance}:
$$K_{unconstrained}(X) \sim N \cdot H_{max} \approx I$$

where $I$ is the universe total information entropy bound from holographic principles\cite{PhysRevLett.88.237901}:
$$
I = \left(\frac{\rho c^5 t^4}{\hbar}\right)^{3/4} \approx 10^{90}\text{ bits}
$$
Using cosmological parameters: $\rho = 10^{-27}\text{ kg/m}^3$，$t = 4.35\times10^{17}\text{ s}$， $\frac{\rho c^5 t^4}{\hbar} = \approx 10^{120}$. 
$$
I = (10^{120})^{3/4} = 10^{90}\text{ bits}
$$
With $Y$ asserting constraints from quantum mechanics, general relativity, fluid dynamics or protein structure  on how the particles interact and distribute, complexity reduces dramatically:
$$
K(X|Y) = \underbrace{K(\Pi|Y)}_{\sim 10^{10}\text{ bits}} + \underbrace{K(\mathcal{G})}_{\sim 10^4\text{ bits}} + O(1)
$$
The conditional Kolmogorov complexity $K(X|Y)$, given complete physical laws $Y$, satisfies\cite{CoverThomas2006}:

$$
\frac{1}{N}K(X|Y) \leq H_b\left(\frac{1}{N}\sum_{i=1}^N X_i\right) + \underbrace{\frac{\log N}{2N}}_{\approx 0} + \underbrace{o(1/N)}_{\text{negligible}}
$$

where $H_b(p) = -p\log p - (1-p)\log(1-p)$ is the binary entropy function. For cosmic-scale $N$, this reduces to:

$$
K(X|Y_\text{scale,strcutre,discipline}) \ll NH_b(p_X),\quad p_X \equiv \frac{1}{N}\sum X_i
$$

The fundamental ratio $\frac{K(X|Y)}{I}$ can work as scientific discovery indicator. Since scaling LLM pretraining to the scale of the universe would match the universe's complexity, it could also provide a holistic understanding on the P vs. NP problem\cite{pnp, 10.1145/800157.805047, fiftyyears-p-vs-np-2022}. 

\subsection{Compression at Universe Scale} 

Hypothetically, we propose the following plans for compression at universe scale, setting aside the compute and data challenges in reality, to reestablish the physical world in one binary sequence. First, establish a unified spacetime framework spanning cosmic, galactic, terrestrial, to quark scales, to situate every degree of freedom within a consistent spacetime structure. Second, integrate all theoretical and experimental data generated by human scientific inquiry across scales, structures, and disciplines, equivalent to the total data content of the entire history of the observable universe. Finally, reconstruct Earth and human civilization from baryon constituents by incorporating data from all natural materials and human-made objects and activities, including buildings, cities, factories, vehicles, aircrafts, and economics, politics, wars, etc. 

This study provide methodologies indispensable for pretraining of universe scale data. Binary Patch Encoding provides a simple, uniform and effective tokenization method for ultra-complex modalities in nature and human activities. Theory-Experiment learning paradigm merges theoretical knowledge represented by natural languages with large scale experimental data represented by numerical data. Monte Carlo Attention provide context length comparable to baryon numbers in the universe $10^{80}$. By treating the universe as a single, coherent entity and pretraining a single large language model on universe-scale data, we can drive a paradigm shift in scientific discovery. This approach would construct a unified, high-dimensional representation space that captures the full complexity of physical reality and enables the emergence of deep analogies across scales, structures and disciplines. Such a framework would reveal structural homogeneity, such as dualities, symmetries, and phase transitions，that recur from quantum fields to number and geometry structures, mirroring profound unification in mathematics, physics, chemistry and biology, such as the Langlands program\cite{BernsteinGelbart2003, Gelbart1984, Frenkel2005}, particle-wave duality, topologies in material crystals and ocean flows\cite{Delplace_2017}. Phase transitions and criticality appearing in early universe\cite{Hindmarsh_2021, Kibble1980, Witten1984, GuthTye1980}, superconductor\cite{BardeenCooperSchrieffer1957}, biological brain\cite{BeggsTimme2012, Chialvo2010, HesseGross2014, KinouchiCopelli2006, Beggs2008}  and LLMs\cite{wei2022emergentabilitieslargelanguage, krakauer2025largelanguagemodelsemergence} suggest a shared statistical mechanical foundation, allowing the model to identify universal organizational principles, including the emergence of intelligence as a thermodynamic and evolutionary imperative in complex systems. Intelligence is a system generated by universe evolution, reflecting the universe itself. Without placing intelligence within the context of universe evolution, we cannot understand the true causes and underlying mechanisms of intelligence. Data are products of intelligent activities. The convergence of universe scale compression can further reveal the correlation between intelligence and reality. This redefines humanity’s cosmic role by rejecting Hawking’s “insignificant chemical scum” \cite{deutsch1997fabric} and instead drives intelligence toward ultra-long-term trend of mastering of knowledge, prediction, and control over reality across scales \cite{deutsch1997fabric, Deutsch2011-DEUTBO}. 

Our future work involves simulating much more complex physical structures in a single BigBang model with enhanced language reasoning capabilities, as stated in \textbf{Hypothesis 2}, including Big Bang nucleosynthesis, nuclear fusion, quantum material, virtual cell system, earth system, robotics, aircraft. 
\setcounter{hypothesis}{1}
\begin{hypothesis}
\label{hyp2}
    Simply by next-word-prediction, we can reestablish any physical structure existing in the universe from quark scale. 
\end{hypothesis}
\textbf{Big Bang Nucleosynthesis(BBN) \& QCD} At ~1–10 $\mu s$ in the hot Big Bang model, the QCD phase transition from quark-gluon plasma to hadrons shapes BBN by influencing light nuclide formation, potentially causing baryon density fluctuations that relax $\Omega_B$ constraints ($0.01 \leq \Omega_B \leq 0.15$)\cite{Coley1993, Witten1984, Alcock1987, Reeves1991}. Simulating this elucidates the universe’s origin, dark matter, and early universe dynamics \cite{Olive1991}. We apply LLM BigBang to simulate BBN using lattice QCD data, cosmological parameters ($\rho$, $T$, $H$, $a$), and phenomenological models for first-order or crossover transitions, leveraging numerical, analytical methods and high-energy data to constrain the QCD equation of state \cite{Alcock1987, Tawfik2012}. Targeting non-perturbative QCD in a 15-dimensional state space, it models asymptotic freedom\cite{Gross2005} to quark confinement with numerical inputs (QCD Lagrangian, lattice configurations, $ep$/$pp$ collider data) and textual theoretical descriptions, requiring ~$10^{22}$ tokens based on lattice QCD with $O(10^6)$ sites (~100 floating-point values/site, ~1.6 GB/configuration) \cite{Gattringer2010}, and $10^9$ configurations yielding $10^{21}$–$10^{22}$ bytes.

\textbf{Nuclear Fusion (Tokamak Systems)} To accelerate fusion energy development toward Node 4\cite{peng2024fusion} commercial deployment within 3 years, large-scale simulation models must address four critical bottlenecks: energy balance ($Q_{Eng}$=10), tritium self-sufficiency (global inventory <30 kg vs 55.5 kg/year consumption), system availability (>50\%), and radiation-resistant materials (200 dpa requirement). The BigBang LLM simulation aims to solve: First, multi-scale materials prediction from atomic to macro scales and plasma-wall interaction in fusion reactor\cite{nordlund2014multiscale}, including radiation damage cascade modeling and hydrogen isotope diffusion in tungsten/beryllium alloys\cite{tanabe2014review}, with breakthrough focus on multi-scale radiation damage models. Second, comprehensive tritium cycle optimization through detailed breeding, transport, and recovery modeling to ensure tritium breeding ratios >1.15, with breakthrough focus on blanket design neutronics optimization\cite{Alberghi20, FEDERICI201930}. Third, system availability optimization targeting 50-70\% capacity factors through component reliability and maintenance scheduling, with breakthrough focus on divertor heat load modeling under 1e20 ion/m² flux interactions\cite{gong2024overview, kuang2020divertor}. Fourth, energy conversion optimization for steady-state $Q_{Eng}$=10 operation through advanced plasma control and heating system modeling, with breakthrough focus on AI-accelerated turbulence transport simulation\cite{degrave2022magnetic}.

\textbf{Quantum Materials (High-Temperature Superconductors)} This effort models strongly correlated many-body electron systems in quantum materials like unconventional superconductors (cuprates, iron pnictides, nickelates) where Bardeen-Cooper-Schrieffer (BCS) theory\cite{bardeen1957theory} fails. It aims to understand mechanisms beyond weak electron-phonon coupling, exploring strong correlations, antiferromagnetic spin fluctuations (AF-SFs), and local superexchange\cite{wen2022unconventional}. The simulation integrates numerical data (lattice positions, band structures like Fermi surface nesting/shallow bands near Fermi energy (EF), phonon dispersions, topological invariants) and experimental results (X-ray diffraction (XRD), transmission electron microscopy (TEM), angle resolved photoemission spectroscopy (ARPES), neutron scattering). Textual inputs include theoretical frameworks such as Hubbard models (t-J, t-U), spin-density-wave (SDW)/charge density wave (CDW) theories, AF-SF mediated pairing (S±-wave, d-wave), and resonance-valence-bond (RVB) states\cite{anderson2004}. The model captures emergent states (pseudogaps, intertwined orders) and phase transitions, maintaining long-range quantum coherence (estimated 10¹² tokens) to represent local quantum fluctuations (Mottness, Zhang-Rice singlets)\cite{anderson1987} and global symmetry breaking , identifying empirical rules (balance Mottness/itinerancy, bad metals near AF, anion mediation) for high-Tc superconductivity.

\textbf{Virtual Cell Systems} We apply muti-scale and multi-discipline LLM BigBang to build a comprehensive virtual cell\cite{noutahi2025virtualcellspredictexplain, bunne2023build} by integrating multi-omics experimental data (genome, transcriptome, proteome, metabolome, lipidome, glycome) including various molecular modifications (phosphorylation, glycosylation, etc.)\cite{slepchenko2003quantitative, johnson2023building}. The goal is to simulate complete cell behavior from genetic code to observable phenotypes, encompassing cellular structures (nucleus, organelles), processes (signaling, trafficking, division, death), and interactions (cell-cell communication)\cite{marx2023build, goldberg2018emerging, georgouli2023multi, marucci2020computer, lauffenburger1996receptors, karr2012whole}. This virtual cell model will enable extensive in silico experiments involving small molecules, genetic perturbations, pathogens, and environmental factors to predict emergent properties like adaptation, evolution, and disease states, as well as model cell differentiation (e.g., neuron, muscle). Key applications include accelerating target validation, drug discovery, mechanistic studies (MOA), synthetic biology, pathway design, and disease modeling. Ultimately, the aim is to elucidate fundamental life principles such as replication, homeostasis, and entropy management through energy consumption. 

\textbf{Earth System Modeling} From simulating lake water dynamics we extend  multi-scale, multi-structure and muti-discipline LLM BigBang to simulate the entire Earth as a coupled multi-physics system: atmosphere, hydrosphere, cryosphere, biosphere, crust, mantle, and core\cite{heavens2013studying}. This requires atomistic-to-continuum modeling of mineral structures, oceanic and atmospheric fluid dynamics, biogeochemical cycles, and climate feedbacks\cite{collins2005design, hill2004architecture, dunne2012gfdl}. Numerical data include seismic wavefields, satellite imagery, atmospheric profiles, and ocean currents; experimental data come from drilling cores, isotope dating, and sensor networks\cite{PAN2025104986}. Textual inputs encompass geological theories, climate models, and ecological laws. Given Earth contains ~$10^{50}$ atoms and operates across 40+ orders of magnitude in space and time, the full simulation demands an estimated $10^{27}$ tokens, making it one of the most context-intensive domains. This estimate is informed by the Bekenstein bound, which sets a theoretical maximum entropy for Earth at ~$10^{75}$ bits \cite{bekenstein1981universalupperbound}, and empirical digital twin initiatives that require petabytes of multiscale data \cite{Bauer2021}. 

\textbf{Robotics} We aim to simulate robotic systems with full physical fidelity using a BigBang multi-scale, multi-structure LLM, integrating sensorimotor control, material deformation, and environmental interaction. Unlike traditional embodied intelligence algorithms like RT-2\cite{brohan2023rt2visionlanguageactionmodelstransfer, agibot2025worldcolosseo, helix2025system1system2}, our BigBang model inherently simulates a multi-scale physical world, featuring a complete built-in world model of Earth, cities, buildings, factories, etc. Consequently, robotic action simulation seamlessly integrates with cross-scale physical world simulation. At the microscopic level, we simulate from atoms to components, coupled with robotic action control, surpassing methods focused solely on action simulation. Numerical inputs include kinematic chains, force-torque signals, and real-time perception data; experimental data derive from real-world logs and benchmarks. Textual inputs include motion planning, safety constraints, and task specifications. To model long-horizon behaviors (e.g., autonomous navigation over days), the system must maintain coherent state representations across millions of timesteps.

\textbf{Aircraft Design} This simulation covers structural mechanics, aerodynamics, propulsion, and flight control systems at full fidelity. Inputs include numerical grids for CFD simulations, finite-element models of composite materials, and stress-strain tensors; experimental data from wind tunnel tests and flight telemetry; and textual engineering standards (e.g., FAA regulations). To simulate failure modes, fatigue, and adaptive control under turbulence, long temporal and spatial dependencies must be preserved. It is estimated to require $10^{12}$ tokens to sufficiently simulate high-resolution full aircraft systems over mission-scale durations. This aligns with NASA’s CFD Vision 2030 study, which projects exascale simulations requiring $10^{12}$–$10^{15}$ data points per vehicle \cite{NASA2030}, and recent foundation models for engineering design that operate at $10^{12}$ token scales \cite{Chen2023}. 

\section{AI Risk and Safety}

The foundational capability of models like BigBang-Proton to simulate atomic structures, molecular interactions, DNA sequences, and ultimately Earth systems and cosmological phenomena, is no longer theoretical but an emerging engineering reality. BigBang-Proton’s risks first appear in the nuclear, chemical, and biological weapons domains. Mainstream large language models already touch these areas, and BigBang-Proton’s precise simulation of cross-scale material structures and fundamental natural laws will amplify those risks. Second, its modeling of spacetime, planetary systems, Earth systems, and cosmological structures creates serious military technology risks. In particular, such modeling can upgrade C4ISR and situational awareness systems, enabling automated perception and response that may outpace human supervision and increase the likelihood of strategic error or loss of control. Third, the integration of cross-disciplinary, cross-structure, and cross-scale physical intelligence combined with linguistic reasoning produces a new class of foundational model. These models can operate at a higher dimension of abstraction and agency than human cognition, which significantly raises concerns about controllability and destructive potential.

Therefore we call for actions to establish safety framework for foundational models of the material world that evaluates and constrains capabilities, governs deployment pathways, enforces engineering safety controls, and requires continuous verification and monitoring to prevent these capabilities from exceeding our capacity to govern them.

\section{Conclusion}
	
In this study, we introduce BigBang-Proton, an auto-regressive large language model for scientific multi-tasks learning. BigBang-Proton demonstrates Theory-Experiment Learning paradigm for merging natural language with large-scale experimental dataset from different fields, Binary Patch Encoding method capable to process highly heterogeneous data modalities in different science disciplines, Monte Carlo Attention as an replacement of mainstream transformer to extend context length to the level of $10^{80}$. We show that BigBang-Proton leads far ahead mainstream LLMs in large-digit arithmetic, proving it's able to simulate ALU mechanisms, which are fundamental for a model to learn the physical world. In particle physics, material science, geological task and biology, BigBang-Proton all achieve comparable or better performance to the SOTA specialized models by next-word-prediction scheme. We further propose hypothesis of pretraining of a single model on the data of universe scale to verify that convergence of learning would arrive at the fundamental physics laws. Model and data can be seen in \url{https://github.com/supersymmetry-technologies/BigBang-Proton} and \url{https://huggingface.co/SuperSymmetryTechnologies/BigBang-Proton}. We open-source the multitask test datasets as UniverseBench at \url{https://huggingface.co/datasets/SuperSymmetryTechnologies/UniverseBench}. 

\section{Acknowledgements}

We thank Prof. Jianjun Liu, Nian Ran for providing us material crystal DFT-calculated datasets and their discussions on building models for atomistic potential simulations. We thank Prof. Wenpei Gao for discussions on AI for material genome and high-throughput experiments for high-entropy material synthesis and characterization. We thank Prof. Haihu Wen for discussions on building models for simulating high temperature superconductors. We thank Delinhai Technology Co., Ltd. for providing water quality data and their testing feedback on our model predictions. We thank Prof. Meng Yao and Prof. Long Ma for discussions on DNA sequence modeling. We thank Prof. Haiping Huang for discussions on phase transitions of LLMs and reading the draft and offering feedback. We thank Caohejing Hi-Tech Park for supporting us with GPUs. We have used ChatGPT, Qwen, DeepSeek, Grok to assist writing this report. 

\newpage
\section*{Author Contributions and Affiliations}

\begin{description}
\item[Architecture and Pre-training:] \hfill\\

Hengkui Wu (\texttt{hkwu@ssymmetry.com}) \\
Liujiang Liu (\texttt{liuliujiang@ssymmetry.com}) \\

Affiliation: SuperSymmetry Technologies, 1528 Gumei Road, Xuhui District, Shanghai, 200000, China.

\item[Data Processing and Fine-tuning:] \hfill\\

Jihua He (\texttt{hejihua@ssymmetry.com}) \\
Qihao Wang (\texttt{qhwang@ssymmetry.com}) \\
Keke Zhao (\texttt{zhaokeke@ssymmetry.com}) \\
Shuyang Hu (\texttt{syhu@ssymmetry.com}) \\
Renle Fu (\texttt{furenle@ssymmetry.com}) \\
Dahao Liang (\texttt{dhliang@ssymmetry.com}) \\
Lingyu Zeng (\texttt{cenglingyu@ssymmetry.com}) \\
Bruce Liu (\texttt{bliu@ssymmetry.com}) \\
Yuan Liu (\texttt{liuyuan@ssymmetry.com}) \\
Jin Zhan (\texttt{zhanjin@ssymmetry.com}) \\
Jiaqiang Niu (\texttt{niujiaqiang@ssymmetry.com}) \\
Xinglong Jia (\texttt{jiaxinglong@ssymmetry.com}) \\
Yaqin Hu (\texttt{hyq@ssymmetry.com}) \\
Wenjun Ji (\texttt{jiwenjun@ssymmetry.com}) \\
Panpan Chi (\texttt{ppchi@ssymmetry.com}) \\
Ken Chen (\texttt{chenken@ssymmetry.com}) \\
Hengyuan Wu (\texttt{hy@ssymmetry.com}) \\
Yingsi Xin (\texttt{xinyingsi@ssymmetry.com}) \\
Ningtao Bian (\texttt{bianningtao@ssymmetry.com}) \\
Xiaohua Wu (\texttt{xiaohuawu@ssymmetry.com}) \\
Weipeng Xu (\texttt{xuweipeng@ssymmetry.com}) \\

Affiliation: SuperSymmetry Technologies, 1528 Gumei Road, Xuhui District, Shanghai, 200000, China.

\item[Particle Physics Data Analysis:] \hfill\\

Yongfeng Zhu (\texttt{zhuyongfeng@pku.edu.cn}) \\

Affiliation: State Key Laboratory of Nuclear Physics and Technology, School of Physics, Peking University, Beijing, 100871, China.\\

Yuexin Wang (\texttt{wangyuexin@ihep.ac.cn}) \\
Manqi Ruan (\texttt{ruanmq@ihep.ac.cn}) \\

Affiliation: Institute of High Energy Physics, Chinese Academy of Science, 19B Yuquan Road, Shijingshan District, Beijing, 10049, China.
\end{description}

\newpage
\appendix 
\renewcommand{\thesection}{\Alph{section}} 
\renewcommand{\thesubsection}{\Alph{section}.\arabic{subsection}} 
\section{Appendix}

\subsection{Dataset} \label{subsec:pretraining_dataset}
\subsubsection{General Text Corpus} 
 SlimPajama\cite{cerebras2023slimpajama} is a clean and deduplicated version of the open-source RedPajama 1.2T token dataset.
 \begin{table}[htbp]
\centering
\begin{tabular}{lcc}
\hline
\textbf{Data source} & \textbf{SlimPajama} & \textbf{RedPajama} \\
\hline
Commoncrawl & 52.2\% & 72.6\% \\
C4 & 26.7\% & 14.4\% \\
GitHub & 5.2\% & 4.9\% \\
Books & 4.2\% & 2.1\% \\
ArXiv & 4.6\% & 2.3\% \\
Wikipedia & 3.8\% & 2.0\% \\
StackExchange & 3.3\% & 1.7\% \\
\hline
\end{tabular}
\caption{Data source distribution comparison between SlimPajama and RedPajama datasets.}
\label{tab:data_distribution}
\end{table}
    
\subsubsection{Jet Tagging in Particle Physics}
\label{appendix_particle}
When high-energy particle collisions occur, such as those at the Large Hadron Collider (LHC) or future electron-positron Higgs factories, the energy from these collisions creates quarks or gluons. These colored particles cannot exist freely and immediately fragment into cascades of stable particles called hadrons. These hadrons travel in roughly the same direction, forming a collimated spray of particles known as a jet. Particularly, Identifying the origin of jets from Higgs, W, and Z boson decays, where about 70\% produce two jets via quark-antiquark pairs, is crucial for reconstructing the properties of these fundamental particles. We utilized Jet Origin Identification\cite{Zhu2023ParticleNetAI} dataset (JoI) for pretraining. A jet consists of several particles, each described by a set of attribute values (e.g., energy, momentum, charge). The attributes of each particle within a jet are listed in
Table~\ref{tab:features}, along with the corresponding attribute values presented in the same order.
    
\begin{table*}[htbp]
\centering
\caption{ The input variables for JoI.}
\label{tab:features}
\begin{tabular*}{\textwidth}{@{\extracolsep{\fill}}cc@{}}
\hline
Variable                    &   Definition              \\
\hline
$\rm \Delta$$\eta$  &   difference in pseudorapidity between the particle and the jet axis     \\
$\rm \Delta$$\phi$   &   difference in azimuthal angle between the particle and the jet axis        \\
\hline
$\rm log P_t$                   &   logarithm of the particle's $P_t$      \\
$\rm log E$                     &   logarithm of the particle's energy      \\
$\rm log\frac{P_t}{P_t(jet)}$   &   logarithm of the particle's $P_t$ relative to the jet $P_t$       \\
$\rm log\frac{E}{E(jet)}$       &   logarithm of the particle's energy relative to the jet energy       \\
$\rm \Delta$$R$ &   angular separation between the particle  and the jet axis       \\
$d_0$                          &   transverse impact parameter of the track\\
$d_0$err                       &   uncertainty associated with the measurement of the $d_0$\\
$z_0$                          &   longitudinal impact parameter of the track\\
$z_0$err                       &   uncertainty associated with the measurement of the $z_0$\\
charge                      &   electric charge of the particle     \\
\hline
isElectron                  &   whether the particle is an electron      \\
isMuon                      &   whether the particle is a muon       \\
isChargedKaon             &   whether the particle is a charged Kaon      \\
isChargedPion             &   whether the particle is a charged Pion      \\
isProton             &   whether the particle is a proton      \\
isNeutralHadron             &   whether the particle is a neutral hadron       \\
isPhoton                    &   whether the particle is a photon       \\
\hline
\end{tabular*}
\end{table*}

\subsubsection{Materials Science} 
The Materials Project Trajectory (MPTrj) dataset contains density functional theory computations derived from the Materials Project database. This comprehensive compilation yielded a rich dataset containing 1,580,395 atomic configurations, 1,580,395 energy values, 7,944,833 magnetic moment measurements, 49,295,660 force vectors, and 14,223,555 stress tensor components. The dataset features a diverse collection of approximately 146,000 inorganic compounds spanning 94 distinct elements. To effectively capture the complete potential energy landscape, MPTrj compiled roughly 1.37 million computational tasks from the Materials Project, focusing on structural optimization and static energy calculations employing either generalized gradient approximation (GGA) or GGA+U exchange-correlation approaches~\cite{jain2013commentary}. The data distribution can be seen in Table~\ref{tab:mptrj_data_distribution}.  
\begin{table}[htbp]
\centering
\begin{tabular}{lccccc}
\hline
\textbf{Property} & \textbf{Count} & \textbf{Mean Absolute Deviation} \\
\hline
Compounds & 145,923 & N/A \\
Energy (eV/atom) & 1,580,395 & 1.480 eV/atom \\
Magnetic Moment ($\mu$S) & 7,944,833 & 0.337 $\mu$s \\
Force (eV/\AA{}) & 49,295,660 & 0.158 eV/\AA{} \\
Stress (GPa) & 14,223,555 & 7.553 GPa \\
\hline
\end{tabular}
\caption{Data distribution of the Materials Project Trajectory (MPTrj) dataset. }
\label{tab:mptrj_data_distribution}
\end{table}

\subsubsection{Genomics}
Following Evo\cite{dEric2024evo}, we use OpenGenome dataset~\cite{parks2015checkm, camargo2023imgvr, camargo2024imgpr} to pretrain BigBang-Proton. The OpenGenome pretraining dataset is a large, carefully curated collection of genetic sequences used to train AI models for biology research, combining DNA from bacteria, archaea, their viruses, and plasmids to cover a wide range of microbial genetics while avoiding redundancy. It draws from three main sources: bacterial and archaeal genomes from the Genome Taxonomy Database (GTDB v214.1)~\cite{parks2015checkm}, where only one representative genome per species is kept to reduce overlap; prokaryotic viruses from the IMG/VR v4 database~\cite{camargo2023imgvr}, including only high-confidence sequences from viral groups (vOTUs) that infect bacteria or archaea, with extra filtering to exclude viruses linked to 19 families or 12 orders that might infect complex organisms like plants or animals, and removing vague classifications; and plasmids from the IMG/PR database~\cite{camargo2024imgpr}, which are small, circular DNA molecules in bacteria, with only one representative per plasmid type (PTU) included to maintain a clean, non-repetitive dataset. It also includes CRISPR/Cas and IS200/IS605 fine-tuning datasets complied from a previous custom database\cite{wei2023deep} which hasn't been used in BigBang-proton pretraining. The statistics for OpenGenome dataset can be seen in Table~\ref{tab:opengenome_statistics}. 
\begin{table}[htbp]
\centering
\begin{tabular}{lrrrr}
\hline
\textbf{Dataset Name} & \textbf{Source} & \textbf{Loci/Plasmids} & \textbf{Total Genomes/Bases} & \textbf{Avg Length} \\
 &  &  & \textbf{(Millions)} & \textbf{(bases)} \\
\hline
Bacterial and Archaeal Genomes & GTDB & 85,205 & 273,865 & 3,214,184 \\
Prokaryotic Viruses & IMG/VR & 2,653,046 & 36,236 & 13,658 \\
Plasmids & IMG/PR & 214,950 & 5,827 & 27,106 \\
\hline
\end{tabular}
\caption{Summary statistics for the OpenGenome datasets.}
\label{tab:opengenome_statistics}
\end{table}

\subsection{Architecture} 
\label{Appendix:Computational Complexity Analysis}

The computational advantages of our method are quantitatively demonstrated through FLOPs analysis:

\begin{table}[H]
\centering
\caption{Complexity Comparison between looped Monte Carlo Attention and Vanilla Transformer (P=32)}
\label{tab:complexity_comparison}
\begin{tabular}{|c|c|c|c|c|c|}
\hline
\textbf{Context Length} & \textbf{N} & \textbf{Monte Carlo Attention} & \textbf{Vanilla Transformer} & \textbf{Reduction} & \textbf{Speedup} \\
\hline
& 4 & $5.12 \times 10^7$ & $1.0 \times 10^{10}$ & 99.49\% & 195$\times$ \\
$10^5$ & 16 & $2.048 \times 10^8$ & $1.0 \times 10^{10}$ & 97.95\% & 49$\times$ \\
& 32 & $4.096 \times 10^8$ & $1.0 \times 10^{10}$ & 95.90\% & 24$\times$ \\
\hline
& 4 & $5.12 \times 10^8$ & $1.0 \times 10^{12}$ & 99.95\% & 1953$\times$ \\
$10^6$ & 16 & $2.048 \times 10^9$ & $1.0 \times 10^{12}$ & 99.80\% & 488$\times$ \\
& 32 & $4.096 \times 10^9$ & $1.0 \times 10^{12}$ & 99.59\% & 244$\times$ \\
\hline
& 4 & $2.56 \times 10^9$ & $2.5 \times 10^{13}$ & 99.99\% & 9766$\times$ \\
$5 \times 10^6$ & 16 & $1.024 \times 10^{10}$ & $2.5 \times 10^{13}$ & 99.96\% & 2441$\times$ \\
& 32 & $2.048 \times 10^{10}$ & $2.5 \times 10^{13}$ & 99.92\% & 1221$\times$ \\
\hline
& 4 & $5.12 \times 10^9$ & $1.0 \times 10^{14}$ & 99.99\% & 19531$\times$ \\
$10^7$ & 16 & $2.048 \times 10^{10}$ & $1.0 \times 10^{14}$ & 99.98\% & 4883$\times$ \\
& 32 & $4.096 \times 10^{10}$ & $1.0 \times 10^{14}$ & 99.96\% & 2441$\times$ \\
\hline
& 4 & $5.12 \times 10^{10}$ & $1.0 \times 10^{16}$ & 99.99\% & 195313$\times$ \\
$10^8$ & 16 & $2.048 \times 10^{11}$ & $1.0 \times 10^{16}$ & 99.98\% & 48828$\times$ \\
& 32 & $4.096 \times 10^{11}$ & $1.0 \times 10^{16}$ & 99.96\% & 24414$\times$ \\
\hline
\end{tabular}
\end{table}

\subsection{results}
\subsubsection{ Basic Arithmetic Operations}
\label{append:error_case_arithmetic}

To understand the error patterns in our model's arithmetic capabilities, we analyze three representative examples where the model produces incorrect results.

\textbf{Example 1: Copying Error}

In the multiplication task $21019625 \times 451301517 = 948618\textcolor{red}{86}49271125$, the model outputs $948618\textcolor{red}{91}49271125$, which differs by 50000000. Analysis of the model's reasoning process reveals that the error occurs during the copying of intermediate results. As shown in the green-highlighted numerical values, the model incorrectly copies the partial result, leading to the final error.

The complete step-by-step calculation process is as follows:

Model output:
\begin{lstlisting}[basicstyle=\ttfamily, escapeinside={(*@}{@*)}]
21019625*7e0:5*7+B=35+0=35 b=3,2*7+B=14+3=17 b=1,6*7+B=42+1=43 
b=4,9*7+B=63+4=67 b=6,1*7+B=7+6=13 b=1,0*7+B=0+1=1 b=0,1*7+B=7+0=7 
b=0,2*7+B=14+0=14 b=1,res=147137375e0;

21019625*1e1:case2*,res=21019625e1;

21019625*5e2:5*5+B=25+0=25 b=2,2*5+B=10+2=12 b=1,6*5+B=30+1=31 
b=3,9*5+B=45+3=48 b=4,1*5+B=5+4=9 b=0,0*5+B=0+0=0 b=0,1*5+B=5+0=5 
b=0,2*5+B=10+0=10 b=1,res=105098125e2;

21019625*1e3:case2*,res=21019625e3;

21019625*0e4:case0*,res=0e4;

21019625*3e5:5*3+B=15+0=15 b=1,2*3+B=6+1=7 b=0,6*3+B=18+0=18 
b=1,9*3+B=27+1=28 b=2,1*3+B=3+2=5 b=0,0*3+B=0+0=0 b=0,1*3+B=3+0=3 
b=0,2*3+B=6+0=6 b=0,res=63058875e5;

21019625*1e6:case2*,res=21019625e6;

21019625*5e7:5*5+B=25+0=25 b=2,2*5+B=10+2=12 b=1,6*5+B=30+1=31 
b=3,9*5+B=45+3=48 b=4,1*5+B=5+4=9 b=0,0*5+B=0+0=0 b=0,1*5+B=5+0=5 
b=0,2*5+B=10+0=10 b=1,res=105098125e7;

21019625*4e8:5*4+B=20+0=20 b=2,2*4+B=8+2=10 b=1,6*4+B=24+1=25 
b=2,9*4+B=36+2=38 b=3,1*4+B=4+3=7 b=0,0*4+B=0+0=0 b=0,1*4+B=4+0=4 
b=0,2*4+B=8+0=8 b=0,res=84078500e8;

sum:147137375e0+21019625e1+105098125e2+21019625e3+0e4+63058875e5+
21019625e6+105098125e7+8407850(*@\textcolor{red}{0}@*)e8;

147137375+21019625e1:tail=5,7+5+B=12 b=1,3+2+B=6 b=0,7+6+B=13 
b=1,3+9+B=13 b=1,1+1+B=3 b=0,7+0+B=7 b=0,4+1+B=5 b=0,1+2+B=3 
b=0,res=357333625;

357333625+105098125e2:tail=52,6+5+B=11 b=1,3+2+B=6 b=0,3+1+B=4 
b=0,3+8+B=11 b=1,7+9+B=17 b=1,5+0+B=6 b=0,3+5+B=8 b=0,0+0+B=0 
b=0,0+1+B=1 b=0,res=10867146125;

10867146125+21019625e3:tail=521,6+5+B=11 b=1,4+2+B=7 b=0,1+6+B=7 
b=0,7+9+B=16 b=1,6+1+B=8 b=0,8+0+B=8 b=0,0+1+B=1 b=0,1+2+B=3 
b=0,res=31886771125;

31886771125+0e4:case2+,res=31886771125;

31886771125+63058875e5:tail=52117,7+5+B=12 b=1,6+7+B=14 b=1,8+8+B=17 
b=1,8+8+B=17 b=1,1+5+B=7 b=0,3+0+B=3 b=0,0+3+B=3 b=0,0+6+B=6 
b=0,res=6337774271125;

6337774271125+21019625e6:tail=521172,4+5+B=9 b=0,7+2+B=9 b=0,7+6+B=13 
b=1,7+9+B=17 b=1,3+1+B=5 b=0,3+0+B=3 b=0,6+1+B=7 b=0,0+2+B=2 
b=0,res=27357399271125;

27357399271125+105098125e7:tail=5211729,9+5+B=14 b=1,3+2+B=6 b=0,7+1+B=8 
b=0,5+8+B=13 b=1,3+9+B=13 b=1,7+0+B=8 b=0,2+5+B=7 b=0,0+0+B=0 
b=0,0+1+B=1 b=0,res=1078338649271125;

1078338649271125+8407850(*@\textcolor{red}{5}@*)e8:tail=52117294,6+5+B=11 b=1,8+0+B=9 b=0,3+5+B=8 
b=0,3+8+B=11 b=1,8+7+B=16 b=1,7+0+B=8 b=0,0+4+B=4 b=0,1+8+B=9 
b=0,res=9486189149271125;

result=9486189149271125
\end{lstlisting}

\textbf{Example 2: Subtraction Borrowing Error}

In the subtraction task $6202787477498670348 - 3854189905091895848 = 2348597572406774\textcolor{red}{500}$, the model outputs $2348597572406774\textcolor{red}{499}$, with an error of 1. The model's reasoning process shows: $8-8, 4-4, 3-8, 0-5, 7-9, 6-8, 8-1, 9-9, 4-0, 7-5, 7-0, 4-9, 7-9, 8-8, 7-1, 2-4, 0-5, 2-8, 6-3$, where $A-B$ and $A>B$. The error occurs in the final borrowing operation, where the model incorrectly handles the borrow mechanism, resulting in a difference of 1 from the correct answer.

Model output:
\begin{verbatim}
8-8,4-4,3-8,0-5,7-9,6-8,8-1,9-9,4-0,7-5,7-0,4-9,7-9,8-8,7-1,2-4,0-5,2-8,6-3,A-B,A>B
result=2348597572406774499
\end{verbatim}

\subsubsection{ JoI Task} \label{joI_llms}

\textbf{GPT 5 Reasoning Steps:}

Classifying jet types

I need to classify a jet based on a detailed particle list. The user is looking for reasoning steps, and it seems no tool use is required here. I’ll infer the jet type from features like the presence of charged kaons.

I’ll also check the impact parameters $d_0$ and $z_0$ and compute the significance of displaced tracks using $s_{d0}$ and $s_{z0}$. Then, I’ll analyze the number of charged tracks and their signs, including calculations for charged pions and kaons.

Analyzing displacement significance

I'm listing the values for various indices, checking the significance of the $d_0$ and $z_0$ parameters. For instance, at index 15, I see the charged pion with significant displacement: the $d_0$ significance is about 3.84 and the $z_0$ significance is around -3.31. That's a strong lifetime signature!

This could potentially indicate B or D decay, so I want to pay close attention to those results as I continue my analysis.

Evaluating jet characteristics

I see that the presence of two charged kaons indicates s-quark content, but kaons alone aren't enough for a clear classification. B-jets typically show large impact parameters and may have secondary vertices, while c-jets can as well. Here, I observed only one highly displaced track in this jet, and others are moderate. B-jets usually display multiple displaced tracks with significant IP, especially if a B decayed to D. Also, I don't have the total jet energy yet, but the $\log p_{\mathrm{t,rel}}$ and $\log e_{\mathrm{rel}}$ values range from -2 to -5, which is typical.

Calculating particle multiplicities

I'm looking at multiplicities now. There are many photons—about 10 to 21—and several neutral particles. The count for charged multiplicity includes charged pions at indices 0, 1, 3, 4, 6, 7, 8, 11, 12, and 15, totaling 10 charged tracks. Adding 2 charged kaons gives 12.

For neutrals, indices 2 and 5 are identified as neutral hadrons, and I counted 8 photons with indices 10, 14, 16, 17, 18, 19, 20, 21. I realize there was a double count with index 16, which was mistakenly labeled earlier.

The total particle count comes to 22. The two charged kaons might hint at s-quark or heavy flavor, and their presence, along with a highly displaced track, suggests either charm or bottom quark involvement. The displacements have significance: around 3.8 sigma for $d_0$ and approximately -3.3 for $z_0$.

Classifying the jet type

I'm analyzing a jet to classify it, considering various factors like track multiplicities and significance. The presence of two charged kaons suggests charm, and I see one strongly displaced track with high significance.

Given that we observe only a few high-significance tracks and a limited multiplicity, I'd lean towards classifying this as a c-jet.

I can summarize my reasoning:

Detected tracks: 12 charged, 2 kaons (K+ at index 9 and K- at index 13).\\
Displaced track 15 has significant IP values ($s_{d0} \approx 3.8$, $s_{z0} \approx 3.3$).\\
The low multiplicity and track characteristics reduce the likelihood of it being a b-jet.\\
Conclusion: classify as a c-jet based on available data.

Jet Type Classification\\
Result\\
Predicted jet type: c-jet\\

Reasoning (concise)\\
Heavy-flavor signature:\\
One strongly displaced charged track: index 15 has \\
$s_{d0} \approx 3.84$ and $s_{z0} \approx 3.31$, where $s_{d0} = d0/d0err$ and $s_{z0} = z0/z0err$.\\
A second mildly displaced track: index 0 has $s_{z0} \approx 2.51$.\\
Kaon content:\\
Two charged kaons present (index 9: K⁺, index 13: K⁻), consistent with charm decays (e.g., $D \to K X$).\\
Against alternatives:\\
Not a g-jet: presence of significant impact parameter tracks is atypical for purely gluon jets.\\
Less likely b-jet: only one clearly high-significance displaced track and no strong cascade pattern (B → D) with multiple high-$s$ tracks; overall multiplicity/moderate energies are more charm-like.\\
Light-flavor (u/d/s) jets are disfavored by the displaced-track and kaon signatures.\\

If you’d like, I can apply the same procedure to more jets and provide per-jet summaries of the key displaced tracks and kaon content."

\textbf{Grok 4 Reasoning Steps:} 

Great! You’ve uploaded a new Excel file named ``jet tagging.xlsx''.  
Here’s how I’ll proceed:

\begin{enumerate}
    \item Read and parse the Excel file, separating the data into jets and their constituent particles.
    \item For each jet: analyze the list of its particles and their features (charge, momentum, displacement, etc.).
    \item Classify each jet as one of $\{b\text{-jet}, b\bar{\text{}}\text{-jet}, c\text{-jet}, c\bar{\text{}}\text{-jet}, d\text{-jet}, d\bar{\text{}}\text{-jet}, g\text{-jet}, s\text{-jet}, s\bar{\text{}}\text{-jet}, u\text{-jet}, u\bar{\text{}}\text{-jet}\}$, providing basic reasoning for each classification.
    \item Present results in a clear list: jet number, tag, and reasoning.
\end{enumerate}

Let's start by inspecting the Excel file to understand its structure (number of jets, how they're separated, columns, etc.).

From the preview, the data appears to be stored as text strings in a single column (with each row containing a long particle record). Let's confirm the structure for a few rows:

Each row appears to begin with something like
\[
0: -1, 8.971043, (5.039085, -7.404475, \dots)
\]
This matches your stated pattern: ``Index: Electric charge, Energy, (Px, Py, Pz), ...''.

Next steps:
\begin{enumerate}
    \item Parse each row as one particle, extracting each feature as a separate value.
    \item Determine how many jets are present by the ``Index'' (the part before the first colon, e.g., ``0'', ``1'', ... ``99'').
    \item Collect all particles for each jet, then analyze and tag each jet.
\end{enumerate}

I will first:
\begin{enumerate}
    \item Parse the first 10 rows to demonstrate how to extract features from each particle.
    \item Check how jets are numbered to ensure correct grouping.
\end{enumerate}

Let's proceed!

It appears none of the rows could be parsed directly—possibly due to formatting complexities, extra spaces, or additional symbols (like trailing parentheses or line wrapping).

Let me print out a raw example row to see exactly how the data is formatted, so I can fine-tune the parser and extract the correct values.

Thank you for your patience!  
The Excel cells contain a \textbf{sequence of many particle records}, each in the precise format you listed, and \textbf{all records for one jet are concatenated into one cell/row} (until the next jet).  
So:
\begin{itemize}
    \item Each row represents one jet.
    \item Each jet contains multiple particles.
    \item Each particle follows the format:
    \[
    \{\text{index}\}: \text{charge}, \text{energy}, (P_x, P_y, P_z), \dots, d_0, d_0\text{err}, z_0, z_0\text{err}, \text{particle type}.
    \]
\end{itemize}
Approach:
\begin{enumerate}
    \item \textbf{Split each row into particle records} using regular expressions.
    \item \textbf{Parse feature values} for each particle (charge, energy, etc.).
    \item \textbf{Group particles by jet} (each row).
    \item \textbf{Analyze each jet} and tag based on the content (displaced tracks, kaons/electrons, etc.).
\end{enumerate}

Let's parse one row (one jet) completely and show the detected particles.

Perfect! The split works well.  
Each string now corresponds to a single particle in the jet, and matches your original format.

Next:
\begin{itemize}
\item For each particle string, extract all feature values with a regular expression.
\item Load these into a structured format (list of dicts per jet).
\item Once a jet is parsed, we can analyze and tag it.
\end{itemize}
I will write a function to parse all jets in the file, extracting all particle features, so we can then proceed to analysis and tagging for each jet.

The parsing function works!  
Each row in the file is a jet, and each jet consists of multiple particle records. In your preview sample:
\begin{itemize}
\item There are 10 jets (from the first 10 rows parsed here).
\item For example, jet 0 has 22 particles, jet 1 has 14 particles, etc.
\end{itemize}
Next step:
Apply this to the entire file, not just the preview.  
Then, for each jet:
\begin{itemize}
\item Analyze the content (displaced tracks, special particles) and classify/tag.
\end{itemize}
Let's parse the \textbf{full dataset} and prepare for tagging all jets.

Excellent!  
You have \textbf{109 jets} in your file.  
For example, the first 5 jets have $[22, 14, 21, 26, 33]$ particles, respectively.

Now, I will proceed to \textbf{analyze and tag each jet} using the following criteria:

Tagging/Reasoning Criteria
\begin{itemize}
\item \textbf{b-jet or bbar-jet}: Significant number of displaced tracks ($|d_0/d_0\mathrm{err}| > 3$ or $|z_0/z_0\mathrm{err}| > 3$), possibly a central lepton (electron), high multiplicity.
    \item \textbf{c-jet or cbar-jet}: Some displaced tracks, but fewer than b-jets, often with kaons or electrons.
    \item \textbf{s-jet or sbar-jet}: Presence of charged kaons, lower displacement, possibly leading particle's charge.
    \item \textbf{u/d-jet, ubar/dbar-jet}: No or minimal displacement, no distinguishing kaons/leptons, use charge for quark or antiquark assignment.
    \item \textbf{g-jet}: No displacement, mostly neutral/charged pions, greater particle multiplicity and energy spread.
\end{itemize}
For each jet, I will:
\begin{enumerate}
    \item Count displaced tracks ($|d_0/d_0\mathrm{err}| > 3$ or $|z_0/z_0\mathrm{err}| > 3$).
    \item Count electrons/kaons.
    \item Tally charge distribution and multiplicity.
    \item Make a physics-motivated tag and provide concise reasoning.
\end{enumerate}

Let's start tagging all jets and summarize the reasoning step for each!
Jet Tagging Results (First 10 Jets)
\begin{longtable}{c l p{8cm}}
\caption{Jet Classification Results} \label{tab:jet_classification} \\
\hline
Jet ID & Type & Description \\
\hline
\endfirsthead
\hline
Jet ID & Type & Description \\
\hline
\endhead
0 & s-jet & No or minimal displacement but one or more kaons; signature of strange hadronization. \\
1 & s-jet & No or minimal displacement but one or more kaons; signature of strange hadronization. \\
2 & g-jet & No displacement or distinguishing feature, ambiguous, defaults to g-jet. \\
3 & g-jet & No displacement or distinguishing feature, ambiguous, defaults to g-jet. \\
4 & s-jet & No or minimal displacement but one or more kaons; signature of strange hadronization. \\
5 & s-jet & No or minimal displacement but one or more kaons; signature of strange hadronization. \\
6 & g-jet & No displacement or distinguishing feature, ambiguous, defaults to g-jet. \\
7 & s-jet & No or minimal displacement but one or more kaons; signature of strange hadronization. \\
8 & s-jet & No or minimal displacement but one or more kaons; signature of strange hadronization. \\
9 & g-jet & No displacement or distinguishing feature, ambiguous, defaults to g-jet. \\
\hline
\end{longtable}
If you'd like, I can provide the tags and reasoning for \textbf{all 109 jets in a file} (CSV or other format).  
Would you like the full results as a file? Or do you want to see further details (e.g., for jets tagged as b-jets/c-jets, etc.)?

\textbf{Claude 4 Sonnet Thinking Reasoning Steps:}

The user is asking me to classify particle jets again from the same dataset. They want me to analyze all 110 jets and classify them into one of the 11 types: \{b-jet, bbar-jet, c-jet, cbar-jet, d-jet, dbar-jet, g-jet, s-jet, sbar-jet, u-jet, ubar-jet\}, with reasoning for each.

I need to:
\begin{enumerate}
    \item Load and parse the jet tagging.xlsx file
    \item Extract all particles for each of the 110 jets
    \item Analyze each jet's characteristics
    \item Classify each jet with reasoning
    \item Provide a complete results file
\end{enumerate}
Let me start by loading the data and parsing it properly.

I'll analyze all 110 jets from your dataset and classify each one into the appropriate jet type with detailed reasoning. Let me start by loading and parsing the data.

I will:
\begin{enumerate}
    \item Parse all 110 jets from your uploaded Excel file (jet\_tagging.xlsx), extracting information about each final-state particle.
    \item For each jet, assign a tag from: \{b-jet, bbar-jet, c-jet, cbar-jet, d-jet, dbar-jet, g-jet, s-jet, sbar-jet, u-jet, ubar-jet\}.
    \item List the basic reasoning for each classification.
    \item Output the full results for all 110 jets in a downloadable CSV file.
\end{enumerate}
Let's begin by analyzing and parsing the file, then proceed step by step.

Successfully parsed all 110 jets, with each having its full set of particles extracted correctly (e.g., Jet 0: 39 particles, Jet 1: 22 particles, etc.).

Next, I will:
\begin{enumerate}
    \item Analyze each jet to determine its type,
    \item List basic reasoning for the classification,
    \item And prepare a results file for download.
\end{enumerate}
Now, let's perform the classification with reasoning for each jet.

All 110 jets have been classified with reasoning. You can download the full results (with each jet's index, tag, and reasoning) here:

Download jet\_classification\_full\_results.csv

If you need the details for any specific jet or want to refine the classification rules, let me know!

\textbf{DeepSeek R1 Reasoning Steps:}

Hmm, this is a particle jet classification task in high-energy physics. The user provided detailed data for 22 particles within a jet, including their charges, energies, momenta, and other kinematic variables. I need to determine whether this jet originated from a quark or a gluon. The key distinguishing factors are multiplicity and energy distribution. Gluon jets typically have higher particle multiplicity and softer energy spectra due to their larger color charge. 
Looking at the data, there are 22 particles - which is quite high. Many particles have low energies (below 2 GeV) and negative log values, indicating soft radiation. Several particles also have large angular separation from the jet axis (Delta R > 1.2). These characteristics strongly point to a gluon jet.
The presence of multiple photons and neutral hadrons is consistent with typical jet fragmentation patterns but doesn't particularly favor quark or gluon origin. The high multiplicity and soft, wide radiation pattern are the decisive factors here. Based on a detailed analysis of the provided particle-jet data, this jet is classified as a Gluon Jet.

Key Evidence for Classification:

High Particle Multiplicity: 

The jet contains 22 final-state particles. This is a characteristically high number, which is a primary signature of gluon jets. Gluons, due to their larger color charge (carrying color and anti-color vs. a quark's single color charge) compared to quarks, undergo more intense and prolific radiation via the strong force, resulting in jets with a greater number of constituent particles.

Soft and Wide Radiation Pattern: The energy distribution within the jet is very "soft," meaning a significant majority of the particles have low energies.
Particles from index 9 to 21 all have energies below 1.6 GeV.
This is confirmed by their logarithmic energy values (log10(energy)), which become negative for indices 16-21, indicating energies below 1 GeV.
Furthermore, several particles (e.g., indices 7, 8, 9, 11, 13, 15) have large values for Delta R (angular separation from the jet axis), exceeding 1.2 and reaching up to 2.03. This indicates a broader, more diffuse shower profile, which is typical of gluon jets as they radiate soft gluons at wider angles.

Conclusion:

The combination of a high number of soft, widely-spread particles is the definitive hallmark of a jet initiated by a gluon. Quark jets (e.g., from a light u, d, or s quark) would typically be more collimated and have a harder energy spectrum, with a lower total particle count. Therefore, the properties of this jet are inconsistent with a quark origin and clearly identify it as a gluon jet.

\textbf{Claude 3.5 Reasoning Steps:}

I'll analyze this jet and determine its type by examining the particle characteristics:
Jet Type Analysis:
Particle Composition:

7 charged pions (indices 0, 1, 3, 4, 6, 7, 8, 11, 12, 15)
3 neutral hadrons (indices 2, 5, 16)
6 photons (indices 10, 14, 16, 17, 18, 19, 20, 21)
2 charged kaons (indices 9, 13)

Key Characteristics:

Charged particle dominance
Presence of multiple charged pions
Significant number of neutral particles (hadrons and photons)

Specific Observations:

Impact parameters ($d_0$) show some track displacement
Momentum distribution suggests moderate energy interactions

Based on these characteristics, I classify this as a g-jet (gluon jet).

Reasoning:

High particle multiplicity
Mix of charged and neutral particles
No strong evidence of specific quark flavor
Typical gluon jet characteristics of hadronization process

\subsubsection{Material Formation Energy Prediction}
\label{append:material_llms}

\textbf{GPT 5 Reasoning Steps:}

Each row has a detailed text block for one structure. I will now parse the formula, atom counts, and key information from these entries to begin energy estimation.
It seems the dataframe doesn't use 0 as a column label; I'll inspect the column names and sample data to use the correct access method.
The data is stored under the column labeled `'0'`. I will now proceed to extract the chemical formula and composition using this correct column name.
The formulas are extracted correctly, but the atom counts dictionary is emptyÑlikely due to differing phrasing or missing explicit composition in some entries. I will examine a few raw text entries to adjust my extraction logic.
The atom counts are described as "The composition consists of Ge (4 atoms), H (48 atoms), N (16 atoms), F (32 atoms)."  
I'll revise extraction logic to use this phrasing for atom counts.
Perfect! I can now extract formulas and atomic compositions for all structures. Next, I'll proceed to estimate the total energy for each, using basic chemical and DFT-based energy heuristics, and then generate the required CSV for all 100 structures.
The predicted total energies for all 100 crystalline materials are complete.  
You can download the full results as a CSV file here:

[Download materials\_total\_energy\_estimates.csv]

\textbf{Grok 4 Reasoning Steps:}

Each row of the file is a detailed description (as a single string) of a material's structure. I will now proceed to extract for each entry:

Chemical formula

Atom types and counts

Then I will use physics/chemistry heuristics to estimate a total energy per structure, finally compiling all results into the requested CSV.

It appears the dataframe does not use integer column labels. I will quickly inspect the dataframe's .columns attribute and fetch the correct column name.

The sole column is named the string '0'. I'll now redo the extraction using df['0'].iloc[:5].

The parsing works and returns chemical formulas and atom counts for each material. Next, I will iterate over all 100 entries, estimate the DFT total energy by assigning heuristic per-atom energies (adjusted for bonding and composition), and create explanations for the predictions as brief reasonings.

I'll then format the responses according to the specified CSV columns: material\_id, total\_energy\_eV, basic\_reasoning.

The total energy estimates for all 100 materials have been calculated and formatted as requested.

You can download the complete CSV file here:

Download materials\_total\_energy\_estimates.csv

\textbf{DeepSeek R1 Reasoning Steps:}

We are given an Excel file with 100 entries of crystalline materials. We need to estimate the total energy in eV for each structure.

The approach:

Load the Excel file to access the data.

For each material, analyze the composition, lattice parameters, and atomic coordinates.

Apply heuristic rules based on materials physics and DFT:

The total energy can be estimated by considering the energy contributions from atomic bonds, lattice stability, and typical DFT energies for similar compounds.

Use the following principles: 

a. Atomic cohesive energy: Each element has a characteristic cohesive energy (energy per atom in its stable crystal structure). We can start by summing cohesive energies for all atoms. 

b. Correction for compounds: For compounds, we consider formation energy. Stable compounds have negative formation energies, while unstable ones may be positive. 

c. Lattice strain: Deviation from ideal lattice parameters (e.g., high angles or distorted cells) can increase energy. 

d. Coordination and bond lengths: Higher coordination and ideal bond lengths lower energy. Compare observed bond lengths with ideal covalent/ionic radii. 

e. Density: Higher density (lower volume per atom) might indicate stronger bonding and lower energy, but note that over-packing can cause repulsion.

We can develop a formula for energy estimation:

E\_total = E\_cohesive + E\_formation + E\_strain

Where:

E\_cohesive: Sum of cohesive energies of individual atoms (in eV) in their elemental forms. We'll use known values for elements.

E\_formation: Estimated formation energy based on compound type and electronegativity difference.

E\_strain: Penalty for lattice distortion, computed as a function of the variance in bond lengths and deviations from 90/120 degree angles.

However, without actual bond lengths, we can use the unit cell volume and atomic counts to get an average atomic volume. Then compare it with the expected atomic volume from the sum of atomic volumes (from known crystal structures).

Steps for each material:

Step 1: Identify elements and counts. 

Step 2: Compute E\_cohesive = sum(count(element) * cohesive\_energy(element))

Known cohesive energies (in eV/atom) for some common elements (source: standard references): Al: 3.39, Cu: 3.49, Ag: 2.95, Au: 3.81, Ni: 4.44, Fe: 4.28, Ti: 4.85, O: 2.60, H: 2.23, C: 7.37, Si: 4.63 For elements not in this list, we can use an average of 4.0 eV/atom.

Step 3: Estimate formation energy (E\_form): For intermetallic compounds: Typically between -0.5 and 0 eV/atom. For oxides: Formation energy is usually negative and significant (e.g., Al2O3: -8 eV per formula unit, which is per 5 atoms -> about -1.6 eV/atom). We can use a simple model: E\_form (in eV/atom) = -k * (electronegativity difference) * (number of bonds per atom)

However, without detailed bonding, we can classify:

Elemental: E\_form = 0

Binary alloy: E\_form = -0.1 to -0.5 eV/atom? Let's use -0.3 as a default.

Oxide: E\_form = -1.5 eV per oxygen atom (since oxygen is highly electronegative)

Other compounds: Use an average formation energy based on typical values.

Actually, we can use a per-atom adjustment: If the compound is known to be stable (like Al2O3, TiO2), use a larger negative formation energy per atom. For unstable compounds (e.g., some hypothetical structures), use near zero or positive.

Proposal: For each oxygen atom, subtract 3.0 eV (because formation energies of oxides are large and negative, typically around -2 to -5 eV per oxygen atom). Then for other compounds, use a fixed penalty per bond.

Alternatively, we can use the number of atoms and the type:

Let n = total number of atoms. If the material is elemental: E\_form = 0 If binary: Let atoms be A\_x B\_y. If one is oxygen, then E\_form = -3.0 * (number of O atoms) [but note: the non-O part also gets affected, so maybe we do per atom?] Else: E\_form = -0.3 * n (total formation energy) -> so per compound we subtract 0.3 per atom.

But let's be more refined: we know that for ionic compounds, the formation energy is more negative. We can use the following:

If compound has oxygen, then E\_form = -2.5 * (number of O atoms) [total]

Else if compound has fluorine, then E\_form = -3.0 * (number of F atoms)

Else if compound has other electronegative element (Cl, S, ...): E\_form = -2.0 * (number of those atoms)

Else (alloy): E\_form = -0.3 * n

Then the formation energy per atom is E\_form\_total = E\_form.

Step 4: Strain energy (E\_strain): We can look at the unit cell angles: if all angles are 90 degrees, then no penalty. For deviations, we add energy based on the amount of distortion.

Strain\_penalty = 0 For each angle (alpha, beta, gamma): dev = |angle - 90| for orthogonal systems, but for hexagonal we expect 90 and 120? Actually, we need to know the crystal system. But we don't have space group.

Instead, we can use the unit cell volume and the atomic packing efficiency.

Compute the sum of atomic volumes: V\_atoms = sum( atomic\_volume for each atom )

Known atomic volumes (in Å³/atom) for elements in their standard state (approximate): Al: 16.6, Cu: 11.8, Ag: 17.0, Au: 17.0, Ni: 10.9, Fe: 11.8, O: 10.0, H: 8.0, C: 5.7, Si: 20.0

For missing elements, use atomic volume = 4/3 * pi * (covalent\_radius)³ * (some scaling) but let's use: If not known, take 15 Å³ as default.

Then, V\_sum = sum(atomic volumes) The actual unit cell volume is V\_cell.

The packing ratio: r = V\_sum / V\_cell

In ideal close packing, r should be near 0.74 (for FCC). If r is too low (overpacked) or too high (underpacked) then the system is strained.

Ideal packing ratio for common structures: FCC/HCP: 0.74 BCC: 0.68 Diamond: 0.34

Without knowing the structure, we can assume 0.74 is ideal. Then strain penalty can be:

E\_strain = k\_strain * |r - 0.74| * n

But note: if the compound is ionic (like oxides), we expect lower packing ratios? Actually, oxides have a variety of structures.

Alternatively, use the deviation from an ideal volume per atom (V\_cell / n) and the average atomic volume in the elemental state. Let:

V\_atom\_avg = V\_sum / n [this is the expected volume per atom from atomic volumes] V\_actual\_avg = V\_cell / n

Then, strain per atom = |V\_actual\_avg - V\_atom\_avg| / V\_atom\_avg (relative deviation)

Then E\_strain\_total = C * (strain per atom) * n, where C is a constant (e.g., 10 eV per 100\% strain?).

We need to choose C. Let me note that the bulk modulus for solids is around 100 GPa. Then the energy for 1\% strain is about 1/2 * B * (0.01)² * volume per atom. For volume per atom 10 Å³ = 10e-30 m³, B=100e9 Pa, then:

strain energy per atom = 0.5 * 100e9 * (0.01)² * 10e-30 = 5e-21 J = about 0.03 eV per 1\% volume strain.

So for 10\% strain, it would be 3 eV per atom? That seems high. Actually, it's for isotropic compression. But our volume deviation might be positive or negative.

Given the complexity, we can use:

E\_strain\_total = 0.03 * (|V\_actual\_avg - V\_atom\_avg| / V\_atom\_avg) * 100 * n [so 3 eV per atom per 100\% strain?]

That would be 0.03 * |deviation percentage| per atom? Not exactly: we have 0.03 * |dev| * 100 per atom = 3 * |dev| per atom (where dev is fractional, 0.1 for 10\%) -> 3 * 0.1 = 0.3 eV per atom for 10\% strain? That seems reasonable.

So: E\_strain = 3 * n * | (V\_actual\_avg - V\_atom\_avg) / V\_atom\_avg |

However, note that we don't have V\_atom\_avg for every element. We'll have to make a lookup.

Step 5: Total energy per atom is (E\_cohesive + E\_formation + E\_strain) / n? Actually, we are computing total energy for the cell.

We have: E\_cohesive: total for the cell. E\_formation: total for the cell (it was computed as a total value? yes, in the above for oxides we multiplied by number of O atoms, for alloys by n) E\_strain: total for the cell.

Then, E\_total = E\_cohesive + E\_formation + E\_strain

However, note that the cohesive energy is the energy to break the solid into atoms. In DFT, the total energy is usually given relative to atoms at infinity? Actually, DFT total energy is the absolute energy of the unit cell. But we don't have reference states. So we are constructing an estimate.

The user wants the total energy for the structure in eV.

Summary of parameters:

We need a dictionary of cohesive energies and atomic volumes for elements.

Let's create:

cohesive\_energy = \{ 'Al': 3.39, 'Cu': 3.49, 'Ag': 2.95, 'Au': 3.81, 'Ni': 4.44, 'Fe': 4.28, 'Ti': 4.85, 'O': 2.60, 'H': 2.23, 'C': 7.37, 'Si': 4.63,

Add more as needed from common elements

\}

atomic\_volume = \{ 'Al': 16.6, 'Cu': 11.8, 'Ag': 17.0, 'Au': 17.0, 'Ni': 10.9, 'Fe': 11.8, 'O': 10.0, 'H': 8.0, 'C': 5.7, 'Si': 20.0,

Default: 15.0 for metals, 10.0 for nonmetals? We can do: for unknown, if it's a metal use 15, else 10?

But without knowing, we use 12.0 as default?

\}

We'll assign a default cohesive energy of 4.0 eV and atomic volume of 12.0 Å³ for unknown elements.

Steps per row (each material):

material\_id: index from 1 to 100.

Read the chemical formula to get element counts.

Compute: n = total number of atoms E\_cohesive = 0 V\_sum = 0 for element, count in composition: energy\_per = cohesive\_energy.get(element, 4.0) vol\_per = atomic\_volume.get(element, 12.0) E\_cohesive += count * energy\_per V\_sum += count * vol\_per

V\_actual\_avg = unit\_cell\_volume / n V\_atom\_avg = V\_sum / n rel\_dev = abs(V\_actual\_avg - V\_atom\_avg) / V\_atom\_avg E\_strain = 3 * n * rel\_dev

Count oxygen atoms: n\_O = count of 'O' Count fluorine atoms: n\_F = count of 'F' Count other highly electronegative atoms: n\_other = count of ['Cl','S','N','P'] etc? We don't have a full list.

Instead, we can check if there are any non-metal elements (if the compound is not purely metallic). We'll do:

non\_metal\_elements = set(['O','F','Cl','S','N','P','C','H'])

if any element in composition is in non\_metal\_elements: if 'O' present: E\_formation = -2.5 * n\_O elif 'F' present: E\_formation = -3.0 * n\_F else:

For other non-metals, use an average

n\_non\_metal = sum(count for element, count in composition.items() if element in non\_metal\_elements) E\_formation = -1.5 * n\_non\_metal \# total formation energy else:

It's a metal alloy

E\_formation = -0.3 * n

But note: what about metallic elements that are not in non\_metal? We did.

Then: E\_total = E\_cohesive + E\_formation + E\_strain

However, note that the cohesive energy per atom is for the elemental solids. Then we subtract the formation energy? Actually, in standard definitions:

E\_total(compound) = sum\_i (n\_i * E\_atom\_i) + E\_formation

But our E\_cohesive is the energy to form the elemental solids from atoms? Actually, cohesive energy is defined as the energy to break the solid into free atoms. So:

E\_cohesive = E(collection of free atoms) - E(solid)

Therefore, E(solid) = E(collection of free atoms) - E\_cohesive.

But the formation energy of the compound is:

E\_formation = E(compound solid) - [sum\_i (n\_i * E(elemental solid i))]

So: E(compound solid) = sum\_i (n\_i * E(elemental solid i)) + E\_formation

And we know: E(elemental solid i) = E(collection of free atoms for i) - E\_cohesive,i

Therefore: E(compound solid) = [sum\_i (n\_i * (E\_free\_atom\_i - E\_cohesive,i))] + E\_formation

= [sum\_i (n\_i * E\_free\_atom\_i)] - [sum\_i (n\_i * E\_cohesive,i)] + E\_formation

So the total energy of the compound solid (relative to free atoms) is:

E\_total = [sum\_i (n\_i * E\_free\_atom\_i)] - [E\_cohesive (as computed above)] + E\_formation

But note: in the above, the first term is the energy of the free atoms which we don't know. However, we are not given a reference state. The user just wants a predicted total energy. And they are not absolute relative to free atoms? They want the typical value as in DFT. Actually, DFT total energy is arbitrary without a reference. But the user is asking for a prediction that would be in the typical range of DFT energies?

However, note: the problem says "estimate the total energy". In DFT, the total energy is the absolute electronic energy. There's no standard offset? And it varies with pseudopotentials.

But in practice, the problem is that we don't have the free atom energies. So we cannot compute an absolute value.

Therefore, we have to change our approach: instead, we estimate the total energy per unit cell relative to a common reference? Or we just output a value that is consistent with our energy model?

The key: we are predicting without having done any DFT. We are providing a heuristic estimate.

Given that we don't have free atom energies, let us reinterpret:

We are going to compute:

E\_total = E\_formation\_relative\_to\_elements + sum\_i (n\_i * E\_reference\_elemental\_solid\_i)

But we don't know E\_reference\_elemental\_solid\_i for each element? We do? Actually, we have the cohesive energies that are defined relative to the free atom. But note: we don't have the free atom energy.

Instead, we can set the energy of the elemental solids as zero? But then we must express the energy of the compound relative to that? But the user wants total energy.

Actually, the user does not specify the reference. They want a total energy that would be the output of a DFT simulation? That means an absolute number? But even DFT simulations use pseudopotentials that have an arbitrary offset.

Given the ambiguity, we must rely on our initial approach: producing an energy that is consistent with our model of cohesive energy and formation energy. We'll compute:

E\_total = [E\_cohesive (which is the energy we save by forming the solid from free atoms)] but note that in reality, E\_cohesive is a positive number and the solid energy is the free atom energy minus that.

To avoid negative numbers and to have a similar scale to DFT (which might be positive numbers in the thousands of eV per unit cell) we could do:

E\_total = [sum\_i (n\_i * 0)] - E\_cohesive + E\_formation + constant\_offset

But that would be negative for most solids.

Alternatively, we can define:

E\_total = - (E\_cohesive + E\_formation) [then solid would be negative?] but note: in our model, E\_formation is negative, so - (E\_cohesive (positive) + negative) might become less negative.

We need to mimic DFT. In DFT, the total energy is positive and large (on the order of the sum of the atomic numbers * some factor). But without the free atom energy, we can't.

Proposal: Let's assume the free atom energy for each atom is a constant that we set arbitrarily? Since we don't know, we can set it to zero. Then:

E\_total = - E\_cohesive + E\_formation (per unit cell)

Then, for an element: E\_formation=0, so E\_total = - cohesive\_energy * n -> negative.

This might be acceptable? But the user wants a CSV and they want a specific number.

Alternatively, we just use:

E\_total = E\_cohesive + E\_formation + E\_strain (as defined above) and then make this value negative? Because cohesive energy is the energy we saved?

Actually, we are saving energy by forming the solid, so the total energy of the solid should be less than the free atoms. Therefore, we can do:

E\_total = [free atoms energy] - (E\_cohesive - E\_formation? no: formation energy is already the reduction beyond the elemental solids).

To clarify:

Free atoms: E\_free = sum\_i (n\_i * E\_i\_free) Elemental solids: E\_i\_elemental = E\_i\_free - E\_cohesive\_i (so E\_i\_free = E\_i\_elemental + E\_cohesive\_i) Then, compound: E\_compound = E\_free - E\_compound\_cohesive ??? Not exactly.

Let's define:

E\_compound\_cohesive = sum\_i (n\_i * E\_cohesive\_i) [this is the cohesive energy we defined earlier? not exactly: we defined as the energy to break the compound into atoms? But in our model, we used additive elemental cohesive energies, which is an approximation.]

Actually, in our model, we are using the elemental cohesive energies for each atom. This is an approximation: the cohesive energy of a compound is not simply the sum of elemental cohesive energies.

Given the complexity, and the fact that the user doesn't require a specific reference, we can generate:

E\_total = - (E\_cohesive + E\_formation) + E\_strain (so that when there's no formation and no strain, we get -E\_cohesive, meaning the solid is at -E\_cohesive below the free atoms, and if there's formation energy (negative), we get more negative)

But then the formation energy E\_formation is negative, so:

E\_total = - E\_cohesive - |E\_formation| + E\_strain = - E\_cohesive + |E\_formation| + E\_strain

Actually, we want:

In the free atom state: energy = 0. In the elemental solid: energy = - E\_cohesive (so negative) In the compound: energy = - E\_cohesive\_elemental\_sum - |E\_formation| (plus strain) -> so even more negative.

But then we have E\_strain adding positive energy? so:

E\_total = - E\_cohesive - |E\_formation| + E\_strain [if we define E\_formation as negative in our calculation above]

But in our earlier computation, E\_formation was a negative total? So if we want to use:

E\_formation\_computed = -|E\_formation| ?

Actually, in our model above, we set: For alloy: E\_formation = -0.3 * n (negative, meaning stable) So when we compute:

E\_total = - E\_cohesive + E\_formation (because E\_formation is negative, then the energy becomes more negative) plus we need to add E\_strain? which is positive?

E\_total = - E\_cohesive + E\_formation + E\_strain

This would yield:

For an elemental solid: E\_formation=0, E\_strain=0, then E\_total = -E\_cohesive For a stable compound: E\_formation is negative, so E\_total = -E\_cohesive + E\_formation = more negative, and then E\_strain (positive) makes it less negative.

This makes sense.

So the formula:

E\_total = - E\_cohesive + E\_formation + E\_strain

Where: E\_cohesive: positive (computed as the sum of elemental cohesive energies) E\_formation: negative (for stable compounds) or positive (for unstable) [but we are using negative for alloys and more negative for oxides] E\_strain: positive

Then, the unit cell energy is a large negative number? Because E\_cohesive is on the order of 4n eV, so -4n plus a negative formation energy and positive strain.

Example: elemental Al, n=1, E\_cohesive=3.39, then E\_total = -3.39 eV -> reasonable.

Example: an oxide, say Al2O3 (n=5): E\_cohesive = 23.39 + 32.60 = 6.78 + 7.80 = 14.58 eV E\_formation = -2.5 * 3 = -7.5 eV Then E\_total = -14.58 -7.5 + (strain term) = -22.08 + strain -> which is about -4.4 eV per atom? Without strain.

Actual formation energy for Al2O3 is around -16.7 eV per Al2O3 formula unit (which is -16.7/5 per atom = -3.34 per atom). Our model: E\_formation = -7.5 eV per formula unit, which is -1.5 per atom -> less negative. So we might need to adjust the factor.

Adjusting: to get -16.7 eV per formula unit, we would need -16.7 total for the formula unit from E\_formation? Then E\_formation = -16.7 eV? But then why are we adding -7.5?

We can use a more aggressive factor: for oxides, E\_formation\_total = -4 * n\_O. Then for Al2O3: E\_formation = -4 * 3 = -12 eV? Then total E = -14.58 -12 = -26.58 eV? That's about -5.3 per atom? The actual formation energy per formula unit is -16.7 eV, so we want E\_formation = -16.7 eV? So we can set:

For oxides: E\_formation = - (some value per oxygen) * n\_O.

What is typical? For Al2O3, formation energy per oxygen is about -5.57 eV. So we should use -5.57 per oxygen? But for different oxides it varies. We can use -4.5 per oxygen? Then for Al2O3: -13.5 eV?

Given the variance, we'll stick to our initial factors and then explain in the reasoning.

Implementation:

Steps for each row:

material\_id = row index (starting at 1)

composition: parse the chemical formula? But the file has "Elemental composition (atom counts)", so we might get a dictionary.

How is the data structured? We don't know. We have to read the Excel and see the columns.

The file is "test\_100.xlsx". We don't know the exact columns. We are told it contains:

Chemical formula

Elemental composition (atom counts) -> maybe a string? or multiple columns? Ideally, we can get a count per element.

Since we are using the data\_analysis tool, we can write Python to load the file and then loop.

Plan:

Step A: Use data\_analysis tool to read the Excel file and get a DataFrame.

Step B: For each row in the DataFrame:

Get the atom counts: we assume there is a way to extract element counts. Maybe there's a column for each element? or a column with a composition string? The problem says "elemental composition (atom counts)". Let's hope it's a dictionary or a list of (element, count).

If we cannot parse easily, we can use the chemical formula and parse it? But there are 100 entries, and the file is provided.

Since we don't have an example, we'll assume the file has:

Columns: 'material\_id': 1-based index? or we can use the index. 'formula': chemical formula string? or there might be columns for each element?

Alternatively, we might have a column 'composition' that is a string in the form: \{ 'Al': 2, 'O': 3 \}.

Step C: Compute as above.

We'll write a Python script to process the file.

Note: the data\_analysis tool requires file\_ids. The file\_id is "68ac61908993902641bd77ba".

We'll call:

data\_analysis: file\_ids: ["68ac61908993902641bd77ba"] code: | import pandas as pd import numpy as np from io import BytesIO

Read the file content (provided as bytes)

xls = pd.ExcelFile(BytesIO(content)) df = pd.read\_excel(xls, sheet\_name=xls.sheet\_names[0])

We'll define our cohesive\_energy and atomic\_volume dictionaries with some defaults

cohesive\_energy = \{ 'Al': 3.39, 'Cu': 3.49, 'Ag': 2.95, 'Au': 3.81, 'Ni': 4.44, 'Fe': 4.28, 'Ti': 4.85, 'O': 2.60, 'H': 2.23, 'C': 7.37, 'Si': 4.63, 'Mg': 1.51, 'Zn': 1.35, 'Pb': 2.03, 'Sn': 3.14, 'W': 8.90, 'Mo': 6.82, 'Cr': 4.10, 'V': 5.31, 'Nb': 7.57, 'Ta': 8.10, 'Mn': 2.92, 'Co': 4.39, 'Pd': 3.89, 'Pt': 5.84, 'Ga': 2.81, 'In': 2.52, 'Ge': 3.85, 'As': 2.96, 'Se': 2.46, 'Br': 1.22, 'Kr': 0.0, 'Rb': 0.82, 'Sr': 1.72, 'Y': 4.37, 'Zr': 6.25, 'Ru': 6.62, 'Rh': 5.75, 'Cd': 1.16, 'Tl': 1.87, 'Bi': 2.18, 'Te': 2.26, 'I': 1.31, 'Xe': 0.0 \} atomic\_volume = \{ 'Al': 16.6, 'Cu': 11.8, 'Ag': 17.0, 'Au': 17.0, 'Ni': 10.9, 'Fe': 11.8, 'O': 10.0, 'H': 8.0, 'C': 5.7, 'Si': 20.0, 'Mg': 23.4, 'Zn': 15.2, 'Pb': 30.3, 'Sn': 26.9, 'W': 15.9, 'Mo': 15.6, 'Cr': 11.9, 'V': 13.9, 'Nb': 18.0, 'Ta': 18.0, 'Mn': 12.3, 'Co': 11.1, 'Pd': 14.7, 'Pt': 15.0, 'Ga': 19.7, 'In': 26.0, 'Ge': 22.9, 'As': 22.4, 'Se': 43.8, 'Br': 42.3, 'Kr': 42.3, 'Rb': 55.8, 'Sr': 33.9, 'Y': 32.2, 'Zr': 23.3, 'Ru': 13.8, 'Rh': 13.8, 'Cd': 22.0, 'Tl': 33.0, 'Bi': 31.0, 'Te': 34.0, 'I': 42.3, 'Xe': 42.3 \}

Default values

default\_cohesive = 4.0 \# eV default\_vol = 12.0 \# Å³

Non-metal list for formation energy

non\_metals = \{'O','F','Cl','S','N','P','C','H','Br','I','Se','Te','As','Kr','Xe'\}

Function to get element counts from a composition string? or from columns?

We don't know the structure of df. Let's assume the df has a column 'composition' that is a dictionary string? or we have one column per element.

Since we don't know, let me propose: we look at the columns of df. If there are columns with element symbols, then those are counts. Or there might be a column for composition.

But the user says "Elemental composition (atom counts)" -> likely one column with the composition as a string or a dictionary string.

Actually, it might be better to use the chemical formula column. But we have the 'formula' column? and the element counts might be redundant.

Alternatively, we can use the element counts provided as a separate column? The problem says each entry has "Elemental composition (atom counts)".

We'll assume the DataFrame has a column named 'composition' that is a string representation of a dictionary, e.g., "{'Al': 2, 'O': 3}"

If not, we might have to parse the formula. But let's hope it's a dictionary.

Try: we'll read the first row's composition column and evaluate it? Or we can use ast.literal\_eval? But if it's not in that form, we have to adapt.

Alternatively, we can try to find a column that holds a dictionary. Or if there are columns for each element, then the column names are the element symbols.

Approach: look at the columns and see if any look like element symbols.

Let's print the columns? But we can't. We have to assume.

Given the constraints, we'll assume that the user uploaded a file with the following structure (for 100 materials):

material\_id (or we can use the index) -> we'll output index+1

chemical\_formula: a string (like "Al2O3")

elemental\_composition: a string of the form "Al:2,O:3"

But the problem says "lattice matrix", "lattice parameters", "unit cell volume", "atomic site coordinates" -> so we don't need the atomic coordinates for the energy model? We'll ignore them.

We need:

unit\_cell\_volume: one number, per row (in Å³)

composition: how is it stored?

How about: we assume there is a column 'num\_atoms' for total number of atoms? We can compute from the composition.

But note: we have unit cell volume.

Let's assume the composition is given as a dictionary string. We can evaluate it.

Alternatively, if the composition is given as a list of elements? There might be a column with a list of element symbols?

Given the ambiguity, I'll use the chemical formula and parse it.

Let's assume we have a column 'formula' (string) and we parse it with a helper.

We'll write a helper function to parse a formula string into a dictionary of counts.

import re

def parse\_formula(formula):

Parse a string like "Al2O3" -> \{'Al':2, 'O':3\}

Using regex to split at uppercase and then look for numbers

pattern = r'([A-Z][a-z]?)(\d*)' parsed = \{\} for (el, num) in re.findall(pattern, formula): if num == '': num = 1 else: num = int(num) parsed[el] = parsed.get(el,0) + num return parsed

If we have a composition column that is a string in a different format, we try to adjust.

Check if df has a column named 'formula' or 'chemical\_formula'. If not, we look for 'composition' column.

How we proceed:

if 'formula' in df.columns: df['composition'] = df['formula'].apply(parse\_formula) elif 'chemical\_formula' in df.columns: df['composition'] = df['chemical\_formula'].apply(parse\_formula) elif 'composition' in df.columns and isinstance(df['composition'].iloc[0], str):

If the composition column is a string, try to parse as dictionary or as a string of key:value

try: import ast df['composition\_dict'] = df['composition'].apply(ast.literal\_eval) except:

custom parsing: e.g., "Al:2,O:3"

def parse\_comp\_string(s): parts = s.split(',') d = \{\} for p in parts: if ':' in p: k, v = p.split(':',1) d[k] = int(v)

if there's another pattern, adjust

return d df['composition\_dict'] = df['composition'].apply(parse\_comp\_string) else:

Assume that there are columns in the df that are element symbols?

Get all columns that are two characters or one?

element\_columns = [col for col in df.columns if (len(col)<=2) and col in cohesive\_energy] if not element\_columns:

try matching with the element names in cohesive\_energy

element\_columns = [col for col in df.columns if col in cohesive\_energy] if not element\_columns: raise Exception("Cannot determine element counts")

Then the composition is represented by these columns. Non-zero entries per row.

df['composition'] = df.apply(lambda row: \{el: row[el] for el in element\_columns if row[el] > 0\}, axis=1)

Now we have df['composition'] as a dictionary \{element: count\}

Loop over rows and compute:

results = [] for idx, row in df.iterrows(): comp\_dict = row['composition'] \# dictionary: element -> count n = sum(comp\_dict.values()) \# total atoms in cell V\_cell = row['unit\_cell\_volume'] \# in Å³, we hope

If V\_cell is not available? We skip? But the problem says it is provided.

Compute cohesive energy part:

E\_cohesive\_total = 0 V\_sum = 0 n\_non\_metal = 0 n\_O = comp\_dict.get('O',0) n\_F = comp\_dict.get('F',0) for el, count in comp\_dict.items(): E\_coh = cohesive\_energy.get(el, default\_cohesive) V\_atom = atomic\_volume.get(el, default\_vol) E\_cohesive\_total += count * E\_coh V\_sum += count * V\_atom

if el in non\_metals: n\_non\_metal += count

Strain part:

if n>0 and V\_cell>0: V\_actual\_avg = V\_cell / n V\_atom\_avg = V\_sum / n rel\_dev = abs(V\_actual\_avg - V\_atom\_avg) / V\_atom\_avg E\_strain\_total = 3 * n * rel\_dev else: E\_strain\_total = 0.0

Formation energy part:

if n\_O > 0: E\_formation\_total = -4.5 * n\_O \# total formation energy elif n\_F > 0: E\_formation\_total = -3.5 * n\_F elif n\_non\_metal > 0: E\_formation\_total = -1.5 * n\_non\_metal else: E\_formation\_total = -0.3 * n

Total energy model:

E\_total = - E\_cohesive\_total + E\_formation\_total + E\_strain\_total

Format to 4 decimal places

E\_total\_formatted = round(E\_total, 4)

Basic reasoning: string describing the main contributions

reasoning = f"Cohesive: -\{E\_cohesive\_total:.2f\}; Formation: \{E\_formation\_total:.2f\}; Strain: \{E\_strain\_total:.2f\} eV"

material\_id: since df might have an index starting at 0, but material\_id should be 1 to 100

material\_id = idx + 1 \# if idx is 0-based

results.append([material\_id, E\_total\_formatted, reasoning])

Create a new DataFrame for the results

result\_df = pd.DataFrame(results, columns=['material\_id', 'total\_energy\_eV', 'basic\_reasoning'])

Convert to CSV string

csv\_output = result\_df.to\_csv(index=False)

Return the CSV string

Step D: Then output the CSV string as the final result.

Note: We are ignoring lattice matrix, lattice parameters, and atomic coordinates? We are only using composition and unit cell volume.

Given the time and the heuristic nature, this is acceptable.

Let's hope the file has the necessary columns.

We'll run this in the data\_analysis tool.

Call tool: data\_analysis with the above code? But note the code is long and we have to format it as a string.

We'll wrap the code in a string and call the tool.

However, note that the tool data\_analysis has parameters that are required: file\_ids, and we pass the content.

The tool will provide the file content in the variable content? According to the documentation, the uploaded file content will be available as content.

So we can use content = file\_content

But how? The tool passes the file contents as a bytes object.

We'll write the code accordingly.

We'll output the CSV string.

Important: we are to output only the CSV, so the assistant will then output the CSV string.

But note: the tool call returns a dictionary with 'text' which would be the CSV string? And then we output it?

Actually, we are to output the results in CSV, so the entire response will be the CSV.

Plan: Call the tool data\_analysis with the code above to generate the CSV string. Then the tool returns the CSV string in the result. Then we output that string.

However, the tool call returns a dictionary. The key for the tool call output is the text.

We'll assume the tool returns the CSV in the 'text' field.

So the assistant will:

<thinking> We are going to use the tool `data\_analysis` to process the file and return the CSV content. </thinking> <action>data\_analysis</action> with the above code.But the code is long, and we cannot include the helper function without indentation.

Alternatively, we can try to make the code more compact.

Given the complexity, we hope the file has the expected columns.

We'll format the code as a string without newlines? No, we can use triple quotes and write a multi-line string.

We'll do that.

However, the tool call requires a JSON object? We pass:

file\_ids: ["68ac61908993902641bd77ba"] code: the multi-line string

But note: the code string is very long.

We'll assume the tool can handle it.

Execute the tool call and then output the CSV.

I'll process the request by using the data\_analysis tool to execute Python code that analyzes the crystalline materials and predicts total energies based on physical principles. The approach will include:

Parsing chemical compositions and extracting atomic counts

Calculating base energy using element-specific cohesive energies

Adjusting for formation energy based on compound type

Adding strain penalties for volume deviations

Applying DFT-level calibration factors

Here's the tool call to generate the CSV results:

\textbf{Claude 4 Sonnet Thinking Reasoning Steps:}

The user wants me to analyze 100 crystalline materials from an Excel file and estimate their total energies based on DFT trends and materials physics knowledge. I need to:

\begin{enumerate}
    \item Load and examine the data file
    \item Extract key information from each material description (formula, composition, lattice parameters, etc.)
    \item Use physics-based heuristics to estimate total energy for each structure
    \item Output a CSV with material\_id, total\_energy\_eV, and basic\_reasoning
\end{enumerate}

Let me start by loading the data and examining its structure.

I'll analyze the 100 crystalline materials and estimate their total energies based on DFT trends and materials physics principles. Let me start by loading and examining the data.

Parsing the text, extracting the atom types/counts and main structural features is straightforward.

I will:

\begin{enumerate}
    \item Parse the "composition consists of" phrase for atom counts
    \item Use simple rules to estimate total energy:
    \item Assign energies per atom based on chemical species, e.g. O: --5.5 eV, N: --4.5 eV, F: --3.5 eV, H: --2.5 eV, I: --2.0 eV, Ge: --4.0 eV, V: --2.0 eV, W: --4.5 eV, etc.
        \item Multiply by atom count, sum for cell.
    \item Add basic reasoning (main contributors, energy scale, major elements)
    \item Output CSV as requested (material\_id, total\_energy\_eV, basic\_reasoning)
\end{enumerate}

Let's process all 100 materials.

The total energy predictions for all 100 crystalline materials have been compiled.

You can download the full results as a CSV here:

\textbf{Kimi K2 Reasoning Steps:}

Quick, automated DFT-level workflow executed for all 100 records

systematics used

-- PBE-GGA potentials

-- $\Gamma$-centred k-grid 2×2×2 (adjusted for the extremely small primitive cell shown in the test file)

-- plane-wave cutoff 520 eV

-- atom-type cohesive energies plus geometry-dependent strains

-- energies fully relaxed (lattice + ions)

Only the requested CSV content follows---do not copy this line

\subsubsection{Genome Sequence Modeling}
\label{append:biology}
\textbf{Prompt}:\\
{\ttfamily\small
\seqsplit{TTCTTGACCAGATTCCGTATCAAGCACGGAATGACGGAAACAATCCGTTGAACTAAAGTTAATTAAACTACTTCCACCTCTTCACTGTTTATAAATTTTTTCTCACTTACTTTTTGCACAAGTTGTTTCGCCATATAATCTGTCTCTTGTGCTATATTTTGCGTATCAAAAGATATTTTGGAAAACTCTTTAGTTTGTTTTGTTACCTCTGTAGTTATTGTATTGATTTTATCTATATCAGAGAGTTGTTTTTTTGATGACTTTTCTATCTCGGCAATACGCTCTGTAGTTTCATGAATATTTTCATTTAACTCATTACTAAAACTATTGAAATGTTTTTCGATTATTACTATATAGCATCGTAATATTAAGTCTATTCTGAATTCATTAGTTGTCCCTTCTTTAATATTATCTATTTTTTCTAATATTCTATTATCTGAATCAATAAAAACTATAGGGGTTATAGACGTGGAGATTTGGTCGACTAAACTTATAGCCTGCTTGATTGGCTCTAAGCCTCCTCCTGCTATGATGTAGTCTACAAGACTCCACTTTGTTCTCTTTATAATTGCCTTACCGAGGGCAATGTCAAGACATCCTTCACCGATTAGGAGAGCACTTCTACCCATAGATACTCCCCTCTTTTACTCCATCAACCTATATGATAGCCTAGGATCTCTTGTCCCGGCTAGATATATTATTGCATCACAACCTTTTTTCACTAAACTTTTTATTTTATTATCTTTCCTCGTGATTATTATTATTGAAAATATGTTATCATAAGTTTCATGCTTTTCTATACATTTTTCTCTAGCCCTTTCTAGAAGTATAGATCGATTCGCTGTCGGATCAGAAGGATCTAAAATTGATGAGGCAAGAATTGTTGACGCTTCTAGGCTTTGGCTGGTTATTATAATTTGCTTATCCCCTTTTAGTCTTTTAGCAATATACGAGCCGAGATATCTTAAAAGTTCTGGGTGCATTGCACTCTCCACTTCG}}

\textbf{Truth}:\\
{\ttfamily\small
\seqsplit{TCTATTAATGAAATATCGAATAGATTAAGAAGAGTCATTATTATGAAAGCTCTTTGTACTCCTCTTCCTATAGAATAAAACGTTAGGGGCTTACCATTCCCGTATTTCCTCATAATAACAACTGCCTCTTGCCACGGGTCTATCCCAAGGTCAACTATATACTTGTCATTGCCGAGCTCGAGTACAATGTCCTCAAGTTTGCCAGGATCTGTTCTTTTAATGTAACCGATAAGTGTGTCAAAGAAATTATAAGGTAGTATCCCTGTAGGGAGATATCCTAAGGATTGAGGTTTTTCGTTGCATTTACCTTTTATATCTACTTTTATTCCGGTTTTAGAGATATCTATCTTAACATTACAACTACTATCAGAACGAAAAACCAGTGTAACAGCTTCTCCCAGTCCCGACACACCCGGAACGAGCTTATACTCTTTCGATAGAGAGACACAAACTCTTCTTTTATCATTTAGGGTAATACATGGTTTAGCTTGCTTGTCCCCTACGATATCAGCCAAGGCATGCTTCATACTACCTCTGGCAGCCTGTTGAACCATGATTTTCTGAAGCCCGACTCCAACATTCTCAAAATTCATAATATAAGATATCATAAGAGCTTCTAAAATTGATGTTTTACCCGAGTTGTTAGGTCCGAGGATTATGTTAAGTCCATTAGAGAATCTTACTTCGCCCTCTAACCCTCGAACGCCGCCCAGATTAACACTTACTACATGATTAGCAGACAAGGTTCCTAAAACCCCCATACACTTCAGTAGTTATAGTATACTCCAAAAGAGTTCTTTAGGGGGTCATTTAATGGACTAGATATGAGAATAGGAACTGACTTATACTGCAGCAGCCCATAGTGGTTTATCATCTAATTCTAAATGTATCTCTATGTGTCCTTCCTTCGTGTATTTGCTACTTTCCTTATGGATTCTTATCTCGATCGCCCCTCCCATGTCTGGGAACATACTTGACCCCCCCTTACTGGGACTGTACGCTTACCAGGACGCAAACAGACCCCGTCATTTGAAAGTGTTCTCA......}}

\textbf{DeepSeek R1 Response}:\\
{\ttfamily\small
\seqsplit{TATCTTTTCTTTTCTTTTCTTTTCTTTTCTTTTCTTTTCTTTTCTTTTCTTTTCTTTTCTTTTCTTTTCTTTTCTTTTCTTTTCTTTTCTTTTCTTTTCTTTTCTTTTCTTTTCTTTTCTTTTCTTTTCTTTTCTTTTCTTTTCTTTTCTTTTCTTTTCTTTTCTTTTCTTTTCTTTTCTTTTCTTTTCTTTTCTTTTCTTTTCTTTTCTTTTCTTTTCTTTTCTTTTCTTTTCTTTTCTTTTCTTTTCTTTTCTTTTCTTTTCTTTTCTTTTCTTTTCTTTTCTTTTCTTTTCTTTTCTTTTCTTTTCTTTTCTTTTCTTTTCTTTTCTTTTCTTTTCTTTTCTTTTCTTTTCTTTTCTTTTCTTTTCTTTTCTTTTCTTTTCTTTTCTTTTCTTTTCTTTTCTTTTCTTTTCTTTTCTTTTCTTTTCTTTTCTTTTCTTTTCTTTTCTTTTCTTTTCTTTTCTTTTCTTTTCTTTTCTTTTCTTTTCTTTTCTTTTCTTTTCTTTTCTTTTCTTTTCTTTTCTTTTCTTTTCTTTTCTTTTCTTTTCTTTTCTTTTCTTTTCTTTTCTTTTCTTTTCTTTTCTTTTCTTTTCTTTTCTTTTCTTTTCTTTTCTTTTCTTTTCTTTTCTTTTCTTTTCTTTTCTTTTCTTTTCTTTTCTTTTCTTTTCTTTTCTTTTCTTTTCTTTTCTTTTCTTTTCTTTTCTTTTCTTTTCTTTTCTTTTCTTTTCTTTTCTTTTCTTTTCTTTTCTTTTCTTTTCTTTTCTTTTCTTTTCTTTTCTTTTCTTTTCTTTTCTTTTCTTTTCTTTTCTTTTCTTTTCTTTTCTTTTCTTTTCTTTTCTTTTCTTTTCTTTTCTTTTCTTTTCTTTTCTTTTCTTTTCTTTTCTTTTCTTTTCTTTTCTTTTCTTTTCTTTTCTTTTCTTTTCTTTTCTTTTCTTTTCTTTTCTTTTCTTTTCTTTTCTTTTCTTTTCTTTTCTTTTCTTTTCTTTTCTTTTCTTTTCTTTTCTTTTCTTTTCTTTTCTTTTCTTTTCTTTTCTTTTCTTTT......}}

\textbf{ChatGPT o1 Response}:\\
{\ttfamily\small
\seqsplit{GAGGCTGTTGCTGAGGCTGAGGCTGTTGCTGAGGCTGAGGCTGTTGCTGAGGCTGAGGCTGTTGCTGAGGCTGAGGCTGTTGCTGAGGCTGAGGCTGTTGCTGAGGCTGAGGCTGTTGCTGAGGCTGAGGCTGTTGCTGAGGCTGAGGCTGTTGCTGAGGCTGAGGCTGTTGCTGAGGCTGAGGCTGTTGCTGAGGCTGAGGCTGTTGCTGAGGCTGAGGCTGTTGCTGAGGCTGAGGCTGTTGCTGAGGCTGAGGCTGTTGCTGAGGCTGAGGCTGTTGCTGAGGCTGAGGCTGTTGCTGAGGCTGAGGCTGTTGCTGAGGCTGAGGCTGTTGCTGAGGCTGAGGCTGTTGCTGAGGCTGAGGCTGTTGCTGAGGCTGAGGCTGTTGCTGAGGCTGAGGCTGTTGCTGAGGCTGAGGCTGTTGCTGAGGCTGAGGCTGTTGCTGAGGCTGAGGCTGTTGCTGAGGCTGAGGCTGTTGCTGAGGCTGAGGCTGTTGCTGAGGCTGAGGCTGTTGCTGAGGCTGAGGCTGTTGCTGAGGCTGAGGCTGTTGCTGAGGCTGAGGCTGTTGCTGAGGCTGAGGCTGTTGCTGAGGCTGAGGCTGTTGCTGAGGCTGAGGCTGTTGCTGAGGCTGAGGCTGTTGCTGAGGCTGAGGCTGTTGCTGAGGCTGAGGCTGTTGCTGAGGCTGAGGCTGTTGCTGAGGCTGAGGCTGTTGCTGAGGCTGAGGCTGTTGCTGAGGCTGAGGCTGTTGCTGAGGCTGAGGCTGTTGCTGAGGCTGAGGCTGTTGCTGAGGCTGAGGCTGTTGCTGAGGCTGAGGCTGTTGCTGAGGCTGAGGCTGTTGCTGAGGCTGAGGCTGTTGCTGAGGCTGAGGCTGTTGCTGAGGCTGAGGCTGTTGCTGAGGCTGAGGCTGTTGCTGAGGCTGAGGCTGTTGCTGAGGCTGAGGCTGTTGCTGAGGCTGAGGCTGTTGCTGAGGCTGAGGCTGTTGCTGAGGCTGAGGCTGTTGCTGAGGCTGAGGCTGTTGCTGAGGCTGAGGCTGTTGCTGAGGC......}}

\printbibliography
\end{document}